\documentclass[10pt,journal,compsoc,twoside]{IEEEtran}

\ifCLASSOPTIONcompsoc
  \usepackage[nocompress]{cite}
\else
  \usepackage{cite}
\fi
\usepackage{tikz}
\usepackage{amsmath}
\usepackage{tcolorbox}
\usepackage{tikz}
\usepackage{amsmath}
\usepackage{filecontents}
\usepackage{xcolor}
\usepackage{multirow}
\usepackage{multicol}
\usepackage{subcaption}
\usepackage{float}
\usepackage{soul}
\usepackage{enumitem}
\usepackage{rotating}
\usepackage{pifont}
\usepackage{hyperref}
\usepackage{graphicx}
\usepackage{listings}
\usepackage{rotating}
\usepackage{makecell}
\usepackage{booktabs}
\usepackage{arydshln}
\usepackage{longtable}
\usepackage[switch]{lineno} 
\usepackage{lineno}

\begin{document}
\newcommand{\luizao}[1]{
    \textcolor{blue}{[\textit{Luiz \textbf{Super Fanfarrão}}: #1]}
    }
    
\newcommand{\luizaoo}[1]{
    \textcolor{blue}{\textbf{#1}}
    }

\newcommand{\mirelaehmuitoincrivel}[1]{
    \textcolor{cyan}{[\textit{Mirela \textbf{Fanfarrona}}: #1]}
    }

\newcommand{\mirelafanfarrona}[1]{
    \textcolor{cyan}{[\textit{Mirela \textbf{Fanfarrona}}: #1]}
    }

\newcommand{\fabriciothemage}[1]{
    \textcolor{orange}{[\textit{Fabrício \textbf{O Mais Fanfarrão de Todos}}: #1]}
    }

\newcommand{\sanjay}[1]{
    \textcolor{purple}{#1}
    }

\newcommand{\shine}[1]{
    \textcolor{magenta}{#1}
    }

\newcommand{\commentario}[1]{\textcolor{red}{\textbf{#1}}}

\newcommand{\change}[1]{\textcolor{black}{#1}}

\newcommand{\ToDo}[1]{
    \textcolor{red}{[\textit{ToDo}: \textbf{#1}]}
    }

\newcommand{\takeaway}[1]{\noindent
    \begin{tcolorbox}[]
    {#1}
    \end{tcolorbox}
}
\newenvironment{packed_enum}{
\begin{enumerate}
  \setlength{\itemsep}{1pt}
  \setlength{\parskip}{0pt}
  \setlength{\parsep}{0pt}
}{\end{enumerate}}

\newenvironment{packed_item}{
\begin{itemize}
  \setlength{\itemsep}{1pt}
  \setlength{\parskip}{0pt}
  \setlength{\parsep}{0pt}
}{\end{itemize}}

\newcommand{\cmark}{\ding{51}}%
\newcommand{\xmark}{\ding{55}}%
\newcommand{\totalfeat}{41}
\newcommand{\samplesize}{31}
\newcommand{\initialsamplesize}{$63$}
\newcommand{\XXX}{\textcolor{red}{XXX}\ }
\newcommand{\windowsize}{\commentario{10}\ }
\newcommand{\linkrepo}{\url{https://github.com/danielaoliveira/Naturalistic-Computer-Usage-Profiles}}
\newcommand{\studystart}{September 2020}
\newcommand{\studyend}{March 2021}

\newcommand{\eg}{\hbox{e.g.,\ }}
\newcommand{\ie}{\hbox{i.e.,\ }}
\newcommand{\etal}{\hbox{\em et al.}}
\newcommand{\etals}{\hbox{\em et al.'s}}

\newcommand{\changed}[1]{\textcolor{black}{#1}}
\newcommand{\savespace}[1]{\textcolor{black}{#1}}

\raggedbottom




\title{\changed{Online Binary Models are Promising for Distinguishing Temporally Consistent Computer Usage Profiles}}



%
%
%

\author{Luiz~Giovanini\textsuperscript{*}, 
Fabrício~Ceschin\textsuperscript{*}, 
Mirela~Silva,
Aokun~Chen,
Ramchandra~Kulkarni,
Sanjay~Banda,
Madison~Lysaght,
Heng~Qiao,
Nikolaos~Sapountzis,
Ruimin~Sun,
Brandon~Matthews,
Dapeng~Oliver~Wu,
André~Grégio,
and~Daniela~Oliveira

\IEEEcompsocitemizethanks{\IEEEcompsocthanksitem L. Giovanini (corresponding author), M. Silva, A. Chen, H. Qiao, N. Sapountzis, D. Oliver Wu, and D. Oliveira are with the Department of Electrical and Computer Engineering, University of Florida, Gainesville, FL 32611 USA (e-mail: lfrancogiovanini@ufl.edu).\protect

\IEEEcompsocthanksitem F. Ceschin and A. Grégio are with the Department of Informatics, Federal University of Parana, Curitiba, PR Brazil.\protect

\IEEEcompsocthanksitem S. Banda, M. Lysaght, and R. Kulkarni are with the Department of Computer and Information Science and Engineering, University of Florida, Gainesville, FL 32611 USA.\protect

\IEEEcompsocthanksitem R. Sun is with the Department of Computer Science, Northeastern University, Boston, MA 02115 USA.\protect

\IEEEcompsocthanksitem B. Matthews is with the Charles Stark Draper Laboratory, Cambridge, MA 02139 USA.\protect

}
\thanks{*The first two authors have equal contribution.}
\thanks{Manuscript received Month {dd, yyyy; revised Month dd, yyyy.}}}

\markboth{}
{Giovanini and Ceschin \MakeLowercase{\textit{et al.}}: Online Binary Models are Promising for Distinguishing Temporally Consistent Computer Usage Profiles}
%

\IEEEtitleabstractindextext{%
\begin{abstract}
    This paper investigates whether computer usage profiles comprised of process-, network-, mouse-, and keystroke-related events are unique and consistent over time in a naturalistic setting, discussing challenges and opportunities of using such profiles in applications of continuous authentication. We collected ecologically-valid computer usage profiles from $\samplesize$ MS Windows $10$ computer users over 8 weeks and submitted this data to comprehensive machine learning analysis involving a diverse set of online and offline classifiers. 
We found that:
(i) profiles were mostly \changed{consistent} over the 8-week data collection period, \changed{with most ($83.9\%$) repeating computer usage habits on a daily basis}; (ii) computer usage profiling has the potential to uniquely characterize computer users (with a maximum F-score of $99.90$\%); (iii) network-related events were the most relevant features to accurately recognize profiles (\changed{$95.69$\%} of the top features distinguishing users were network-related); and
(iv) \changed{binary models were the most well-suited for profile recognition, with better results achieved in the online setting compared to the offline setting (maximum F-score of $99.90$\% vs. $95.50$\%)}. 

\end{abstract}

}

\maketitle

\maketitle

\IEEEdisplaynontitleabstractindextext

%
\IEEEpeerreviewmaketitle

\IEEEraisesectionheading{\section{Introduction}\label{sec:introduction}}
%
%
%

\IEEEPARstart{C}{omputer} user profiling is the procedure of constructing a behavior-based digital identity of a person by leveraging their computer usage data, such as network traffic, process activity, mouse and keyboard dynamics~\cite{Kim2010MAT,Yang2015Character}. This procedure has the potential to uniquely characterize computer users by their usage patterns in terms of activity (\eg process and network events) and temporal consistency (\ie events repeating on a regular basis)~\cite{Tossell2012CWU,Fridman2017AAMDS,Payne2013SEMAA}. 
For example, a middle-aged CEO from a Fortune 500 company uses his computer very differently from a young software developer in a tech startup in California. While the former might uses email client, office software, web browser, and customized company management software, the latter will likely IDE, CAD, and software versioning tools, alongside a web browser, email client, and regularly accesses \texttt{.com.br} websites given their Brazilian descent.

Several factors make computer usage profiles (referred to as simply \textit{profiles} for the remainder of this paper) well-suited for applications in continuous authentication (CA). 
The automatic recording of profiles can be implemented in a transparent fashion without requiring user intervention, thus facilitating usability and acceptability~\cite{Chuang2013Ithinktherefore}.
Furthermore, especially with regards to corporate employees, although some might perceive the recording of computer usage as potentially privacy-invasive, most already have a very limited expectation of privacy~\cite{emami2021understanding}. 
In fact, some employee activities (\eg email and computer usage) are often recorded by their employers in both the public and private sectors~\cite{privacy}.

This paper does \textbf{not} propose a CA solution targeting personal or corporate computer users. 
Instead, based on naturalistic data collected from an ecologically-valid user study, we seek to provide initial evidence on whether profiles constitute a \emph{useful} and \emph{feasible} source of data to uniquely identify computer users \changed{while they use their devices}, discussing challenges and opportunities \changed{of leveraging profiles} for applications in CA. \changed{Towards this end, we also investigated} \emph{temporal changes} in profiles (\ie changes in computer usage habits taking place over time), which may impose additional challenges for CA \changed{and inform the design of more robust CA solutions.} \changed{Our study focused on answering} the following research questions:
 
\begin{itemize}

	\item \textbf{RQ1}: Are computer usage profiles consistent over time (\ie do computer usage habits repeat periodically)?

	\item \textbf{RQ2}: Do computer users have unique profiles? In other words, are profiles distinguishable from one another?
	
    \item \textbf{RQ3}: What features (\eg network events, processes) are most important for \changed{constructing} a unique profile?
	
\end{itemize}

To address these questions, we conducted an Institutional Review Board (IRB) approved user study to collect profiles from $\samplesize$ computer users (MS Windows 10) over an 8-week period \changed{(taking place asynchronously between \studystart\ and \studyend)} in an ecologically-valid setting, wherein users interacted with their computers naturally without any interference or probing from the research team. 
Importantly, our definition of profiles encompasses process-, mouse-, network-, and keystroke-related events on the users' computers. We then used \changed{time series analysis techniques} to investigate temporal changes in the profiles over the study period (\textbf{RQ1}). Our dataset was also submitted to a comprehensive machine learning analysis involving a diverse set of classifiers (one-class and binary models in both offline and online settings) aimed to \changed{assess profile uniqueness} (\textbf{RQ2}) and identify the most important features in distinguishing among profiles (\textbf{RQ3}). \changed{All artifacts created throughout the course of this study, including our module for extracting profiles and our dataset of ecologically-valid computer usage data itself, will be made publicly available for vetted research usage\footnote{\changed{We are discussing with our IRB and legal counsel about options of data release based on University-mediated agreements with vetted researchers to protect the de-identified dataset to ensure that the data is used for ethical and responsible research only.}}}.

The main takeaways of our analyses are: (1) \changed{profiles were mostly consistent over time and repeating on a daily basis, yet exhibited some level of irregularity related to factors such as days off, and change of habits due to the COVID-19 pandemic;} 
(2) the computer usage profiles have the potential to uniquely identify computer users; 
(3) network-related features were the most relevant to accurately distinguish users; and 
(4) \changed{online binary models are preferred over offline binary and one-class models due to better performance in distinguishing profiles and robustness to temporal changes in computer usage.}

The remainder of the paper is organized as follows. 
Section~\ref{sec:threat} discusses the threat model faced by CA approaches and the assumptions of our analysis. 
Section~\ref{sec:related} reviews related work. 
Section~\ref{sec:user_study} presents our user study methodology.
Section~\ref{sec:data_extrac} describes the design and implementation of the profile extractor used to gather users' computer usage data. 
Section~\ref{sec:ml} describes the methodology of our machine learning analyses.
\changed{Section~\ref{sec:results} presents our experimental results.} 
Section~\ref{sec:discussion} discusses study's findings, limitations, and suggestions for future work.
Section~\ref{sec:conclusions} concludes this paper.

\section{Threat Model and Assumptions}
\label{sec:threat}
This section discusses the main assumptions of this work and the threat model faced by CA approaches. 
First, we assume that CA methods are better suited to be employed in corporate environments, which are 
\changed{
disproportionately targeted by internal and external adversaries, and consists of large groups of employees performing their primary tasks with a institutional computer device.}
Second, although our discussion is focused on CA, this paper does not propose yet a new CA solution. 
Third, we consider that corporate employees 
\changed{
already face low expectations of privacy in their work environment~\cite{privacy}. For example, in many organizations, network traffic, files and emails are monitored, and the devices and applications employees can access are restricted.
The development of a CA solution leveraging computer usage profiles involves the recording of activity.  Despite that we do not advocate the development of privacy-invasive CA solutions without the users being aware of their low expectations of privacy. 
}

\smallskip
\noindent
{\bf Threat Model}. 
As discussed in Sec.~\ref{sec:introduction}, the main appeal of a CA system is not to replace traditional, point-of-entry authentication schemes (\eg passwords or security keys), but to complement them and address their limitations---mainly that, after a user is authenticated into the system, their identity is not subsequently verified. 
A CA approach can potentially flag events that do not fit a learned user profile, \eg connections to certain IP addresses or subnets, and patterns of exfiltration of information \changed{even after the user is authenticated, including insider attacks based on activity}. For example, it is plausible that employees working on the same project will have similarities in profiles (\eg same applications, files, schedules). Thus, in constructing a group profile, the malicious employee's outlier behavior (\ie an insider attack) could be flagged as unusual activity.
\changed{Given the challenges and privacy, security, and legal implications of conducting such study in a private corporation or government organization, we conducted our study with mainly university students and considering this large, diverse university environment as a proxy of an organization.}



\section{Related Work}
\label{sec:related}
There is a vast literature on leveraging user profiling for CA using a variety of data, such as network traffic (e.g., \cite{Yang2015Character, Shi2018-sm}), use of applications (e.g.,\cite{Fridman2017AAMDS, Mahbub2019-od}), and keystroke or mouse dynamics~\cite{Ahmed2014BRBFKD, KANG201572, sayed2013biometric, Sun2016-hl}. 
\changed{
There is, nonetheless, a scarcity of publicly available and diverse datasets containing users' application data~\cite{Mahbub2019-od, Murphy2017-ze}. Some large-scale micro-longitudinal datasets collected from several user study participants  are available (e.g., \cite{Mahbub2016-sq, Ferreira2015-qe}), but primarily focus on smartphone data instead of ecologically valid data from a personal laptop or desktop computer. This provides a possible explanation as to why most of related work pertains solely to the context of  smartphones. 
}

\changed{
A notable exception is the dataset by Murphy et al.~\cite{Murphy2017-ze}, wherein the keystroke data, mouse movements and clicks, and background processes were collected from 103 users over 2.5 years on each participant's private computer in an ecologically valid setting. The authors found that authentication performance degraded significantly when data was collected in less controlled environments. Though this dataset is larger than ours, it did not include network metadata, which in our analysis, was the most relevant type of data to distinguish user profiles; we additionally did not collect keystrokes from users. Moreover, Murphy et al. did not analyze the temporal consistency of the profiles.}

One line of research that closely resembles ours is that of Payne et al.~\cite{Payne2013SEMAA}, where the authors collected system events from seven volunteers for two hours in a controlled environment. However, unlike our user study, time information was not associated with user activities. Their experiment was also not ecologically valid because the users were not using their own devices nor their computers naturalistically. López \etal~\cite{URUENALOPEZ201938} used Self-Organizing Maps (SOM) to enhance insight and interpretability about user behavior data by untangling hidden relationships between variables. The authors generated the SOM visualizations of survey results that analyzed user behavior, security incidents, fraud, and data from a malware scanning tool to contrast Internet users' digital confidence with the level of malware infection. Their approach focused on drawing qualitative conclusions from their self-reported dataset. In contrast, we go beyond by exploring a naturalistic dataset quantitatively using multiple \changed{time series analysis methods and} machine learning models. 

In sum, our review of related work on user profiling for CA showed that none of the existing works evaluated the feasibility of \changed{desktop-based} based computer usage profiling in a naturalistic way, as proposed here. 
Instead, many data collection procedures restricted user behaviors to specific tasks~\cite{Ayotte2021-cm, Zhang2015TGBAUAUD} or devices~\cite{Ribeiro2015OSPCVUI}. Some CA proposals operate by building profiles based on short-term user activity records~\cite{Payne2013SEMAA}, while others did not explore changes in user behavior over time~\cite{Huang2017APEFKD, Zhang2015TGBAUAUD}. Thus, though these studies provide valuable insight into what technologies can be employed for computer usage profiling, this procedure's feasibility and temporal robustness remains understudied. Our findings shed light on such aspects, providing an actionable recommendation for future research and designing of effective, micro-longitudinal, behavior-based CA systems.

\section{User Study Methodology}
\label{sec:user_study}
This IRB-approved study requested that participants installed our extractor module on their personal computers and used their devices naturally for $8$ weeks.
The entire study \changed{took place asynchronously from} \studystart\ to \studyend, wherein we successfully captured $8$ weeks of computer usage data from $\samplesize$ of the total \changed{\initialsamplesize} participants enrolled. In this section, we detail our methodology.

\medskip
\noindent
{\bf Participants.} The study was originally comprised of \changed{\initialsamplesize} participants 
who were recruited via SONA\footnote{SONA is an online scheduling software used to recruit participants, manage studies, and provide a study database for university students to sign up in exchange for course credits.}, flyers, Internet advertising (\eg the UF Facebook page that advertises studies), and word-of-mouth. 
Interested individuals were guided towards an online survey to determine eligibility which, if cleared to participate, would ask the participant for informed consent along with demographic information. 
After study completion, the participants who were recruited via SONA were compensated with two course credits, while the remainder with a \$50 Amazon Gift Card. 
For inclusion in the data analysis, participants were required to complete the entire 8-week study period. 
This excluded $32$ of the original \changed{\initialsamplesize}\ enrolled participants (who experienced technical issues as discussed below), thus reducing our sample size to $\samplesize$ participants ranging from 18--53 years \changed{($M = 27.32$ years, $SD = 8.27$; 64.52\% female, 32.26\% male, 3.23\% gender non-conforming).}
Table~\ref{tab:user_demographics} \changed{in Appendix~\ref{appendix:demographics}} summarizes the demographics of our participants. On average, we recorded $272$ hours of computer usage data from each participant ($SD = 189.9$ hours, range: $31-859$), which corresponds to approximately 4.9 hours per study day ($SD = 3.4$ hr/day, range: $0.6-15.3$). 
We found some instances of high computer usage (\eg 15.3 hr/day) that 
we hypothesize 
may be due to the participant leaving their device turned on for extended periods of time.



\medskip
\noindent
{\bf Procedure.} 
At the beginning of the study, enrolled participants first were asked to complete an online survey to determine their eligibility: (i) above 18 years of age, (ii) own a personal computer (desktop or laptop) that is not shared and (iii) is used regularly,
(iv) have regular access to the Internet, (v) have Windows 10 Operating System installed, and (vi) reside in the United States {(as per UF IRB regulations for compensation)}.
Once the participants were deemed eligible, they received a consent form disclosing the study procedures, minimal study risks, and data protection measures. \changed{The study did not involve deception in that participants were informed of the actual research purpose: continuously record their computer activities for 8 weeks} to increase understanding of computer usage profiles.
Participants were informed that keyboard keys and \changed{data sent over the network such as file contents} 
\textbf{would not} be recorded, but that the following \textbf{would} be collected throughout the duration of the study: \changed{metadata related to software and network activity.}
Having read and electronically signed the informed consent form, participants were asked a series of demographic questions (\eg age, ethnicity) and to install our extractor on their personal computer. 
\changed{Of note, our informed consent was approved with fairly general language that made it clear that we would be installing monitoring software on the participants’ computers and collecting various kinds of computer usage information. This language  was deemed clear and sufficient by UF IRB for purposes of informed consent because it is common to use language that omits technical terms/jargon pertaining to the study focused research area. Certain jargon would not only be meaningless to most subjects, but also possibly give them a false sense of security towards the power of these procedures or engender a false sense of protection against potential risks the study entails.}
This extractor would record logs of system-level events and uploads them to our lab servers in a secure fashion. 
For each participant, the 8-week study period began on the day successful installation could be verified within our systems. 
Participants were instructed to use their computers naturally and were reminded not to share their personal computers with anyone while in the study. 
Only IRB-trained research assistants who were part of the project interacted with participants {to assist them with any questions or concerns they might have}. 
Upon completion of the study and after uninstalling the extractor, participants were asked to complete a final debriefing questionnaire comprised of seven questions pertaining to the consistency of their computer usage during the study period. 
After completing the debrief, participants were compensated. 



\medskip
\noindent
{\bf Data Attrition.} 
A few issues occurred during the initial run of the study.
{First, because international participants were ineligible as per UF IRB, we discarded three participants that were located outside the U.S. We compensated them nonetheless (with UF IRB approval) and discarded their collected data.}
Second, our extractor only collected the destination IP address from $41$ participants, thus we were unable to resolve which domains they accessed. To remedy this situation, we began capturing DNS Queries to resolve the destination IP address to its respective domain.
All $41$ participants who encountered this issue were \changed{communicated about an issue with data collected and} invited to restart the study, receiving extra compensation for their extended time. 
The $19$ participants who accepted were then asked to sign an addendum and promptly restarted the study once the new software update was available; the remaining $22$ who declined were compensated upon study completion and had their data discarded from the study.
All of these issues were properly reported to the UF IRB and approval was received to restart the study.
Lastly, after study completion, we noticed that seven participants had technical issues with the extractor; 
we therefore discarded them from our sample.
These issues, coupled with participant attrition, decreased our sample from \initialsamplesize\ to \samplesize\ participants.

\section{Computer Usage Profile Extraction}
\label{sec:data_extrac}
In this section, we describe the design and implementation of the MS Windows 10 profile extractor 
our team developed to collect computer usage data from the study participants. We first considered using existing user-level and system event extractor tools (\eg \cite{tanium, procmon, msetw, sysmon}) \changed{in a standalone fashion}, but decided against these options either because
(1) the tool was intended for diagnostic tasks, thus incurring prohibitive performance overheads for continuous usage, such as required for our user study~\cite{procmon}; 
(2) the tool required nodes where data was collected to be on the same network, which would make our user study infeasible~\cite{tanium}; or 
(3) the tool did not provide fine-grained process activities, such creation or termination of process/thread, the suspension/resumption of processes/threads, process memory access~\cite{msetw, sysmon}, which we deemed crucial in our design for the construction of computer usage-based profiles.

\changed{In our extractor, there are two main components installed on the users’ systems: one for collecting data and one for uploading the data. The data collection component was based on a Sysmon-based logger to collect system level events, and a user-level application to collect keyboard and mouse related events. To collect system-level events (process creation and termination, network connection), we developed a logger based on Sysmon~\cite{sysmon}, a suite of tools to debug, manage, diagnose Windows systems. To collect mouse and keyboard events, we developed a Windows application to collect three types of events– Mouse clicks, Keyboard clicks, and Process Information, completing information collected by the Sysmon-based logger. Whenever a mouse click is made on any application, the timestamp is recorded and associated with the application where the click occurred. If the user keeps clicking on the same application, no further entries are made for the next five minutes. If the user switches to another application and makes a mouse click, that click timestamp is recorded (see log entry examples below). Timestamps for keyboard events are recorded using the same method.}

\begin{itemize}

    \item \changed{\texttt{1288$|$C:/Program Files/Google/Chrome/ Application/chrome.exe$|$132162206271021146$|$}}
    
    \item \changed{\texttt{8056$|$C:/Program Files/AVAST/Application/ AvastBrowser.exe$|$132162207043754581$|$}}
    
\end{itemize}

\changed{This logging application also records complementary process information. It records the amount of time the process stays in user vs. kernel mode in one-minute granularities.} 
\changed{The second component installed in the user’s computers is an uploader, which is a python script that runs every 5 minutes. All logs captured by the extractor module are securely transferred to our lab server.}

\section{\changed{Data Analysis}}
\label{sec:ml}
This section goes over our data pre-processing steps and \changed{analyzes, which involved a diverse set of time series analysis methods and machine learning (ML) models}.


\subsection{Data Pre-processing}

We pre-processed the raw data of each participant to generate an $N \times 6$ matrix where each line represented one minute of computer activity on a given study day (the number of lines therefore varied among users according to their computer usage). The columns contained:
(i) a timestamp,  
(ii) a list of all active processes, 
(iii) a list of all domains accessed, 
(iv) the number of clicks associated with the timestamp, 
(v) the number of keystrokes associated with the timestamp, and
(vi) an indicator (binary) of background processes activity that was detected via the occurrence of network traffic coinciding with lack of keyboard/mouse events, indicating a network activity that was generated by the process independently and without the interaction of the user (\eg automatic updates running without the user's knowledge, or the user listening to music on YouTube without interacting with the browser for a long time period).

\subsection{\changed{Profile Temporal Consistency Analysis (RQ1)}}

\changed{We aimed to investigate whether the profiles were \textit{consistent over time} or \textit{periodic} (RQ1), that is, whether study participants repeated computer usage habits periodically. Towards this goal, from the pre-processed data of each user, we extracted a time series of the number of active minutes ($y$-axis) per hour within the 8-week study period ($x$-axis) in two conditions: (i) considering the entire activity exhibited by the user, which included background process activity, and (ii) discarding background process activity. We included the second condition because the portion of background process activity generated without user knowledge (e.g., automatic software updates) may add a periodic component to the profile that is not related with the user behavior. In other words, each of the $\samplesize$ users was represented by two time series of $1,344$ data points each (56 study days $\times$ 24 hours/day) ranging from $0$ to $60$.}

\begin{figure}[t]
    \centering
    \includegraphics[width=\columnwidth]{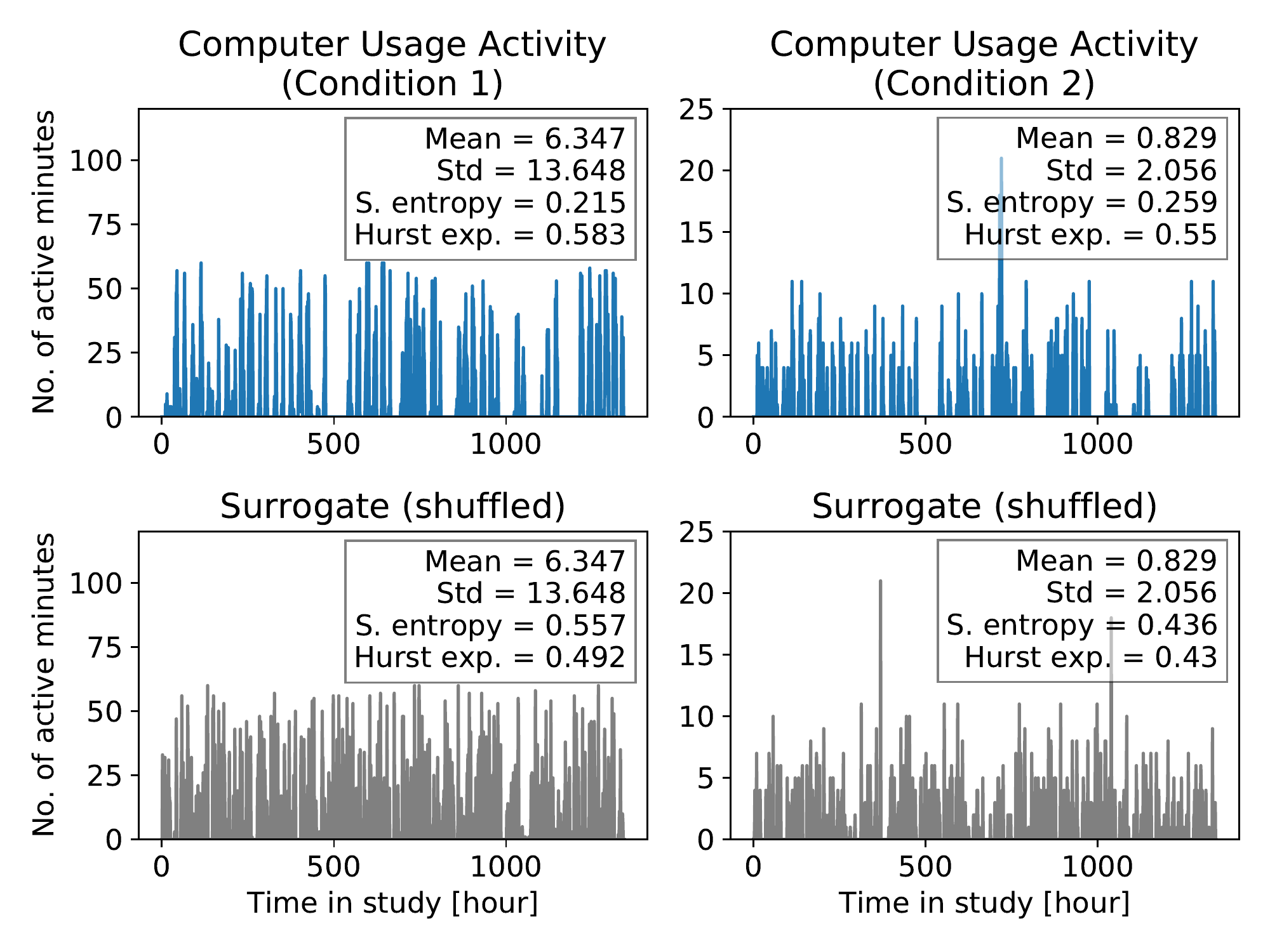}
    \caption{\changed{Time series of computer usage data and its respective surrogate counterpart for User $1$ in Conditions $1$ (left) and $2$ (right).}}
    \label{fig:time_series_surrogates}
\end{figure}


\changed{In a nutshell, our analysis was comprised of two steps. First, we used quantitative methods to investigate whether the time series describing the profiles were periodic (i.e., structured, correlated in time) or random/stochastic (i.e., unstructured, uncorrelated in time). Next, for all time series differing from unstructured data, we used standard techniques to check for periodicity and identify the period.}

\changed{In the first step of our analysis, we used the surrogate data testing method~\cite{theiler1992testing} to check for stochastic behavior in our time series. This method consists of comparing a particular nonlinear measure obtained from the time series with the distribution of the same measure obtained from a set of constructed time series---also known as \textit{surrogates}---with the same statistical properties (e.g., mean, variance) but completely uncorrelated in time. 
If the result from the time series deviates significantly from the surrogate distribution, one can conclude that the time series is unlikely to be purely stochastic and therefore possesses some level of correlation in time. 
We chose two well-known nonlinear measures able to capture temporal correlations in time series data:}

\begin{itemize}

    \item \changed{\textit{Sample entropy} corresponds to the negative natural logarithm of the conditional probability wherein sequences similar for $m$ points remain similar after adding one more data point ($m+1$) within a tolerance threshold $r$~\cite{richman2000physiological}. The more unstructured (random) the time series, the larger the sample entropy.}
    
    \item \changed{\textit{Hurst exponent} is a measure of long-term memory (or long-range correlations) of a time series, ranging between $0$ and $1$. A Hurst exponent of $0.5$ suggests a stochastic behavior. Results lower than $0.5$ suggest an anti-persistent behavior (a high value is followed by a low value and vice-versa) while results higher than $0.5$ suggest a trending behavior (a high value is followed by a higher one). The closer the Hurst exponent is to $0$ or $1$, the stronger the anti-persistent or trending behavior of the time series, respectively.}
    
\end{itemize}

\changed{We created $100$ unique surrogates for each time series by shuffling its data points, thus destroying any existing temporal correlation---which we measured via sample entropy and Hurst exponent---while preserving the statistical properties, as illustrated in Figure~\ref{fig:time_series_surrogates}. We computed the sample entropy and Hurst exponent using the \texttt{nolds} Python library with default parameters ($m=2$ and $r=0.2$ in the former). Before calculating sample entropy, we normalized the time series and surrogates to zero mean and unit variance to remove any non-stationary components that may confound results~\cite{costa2007noise}. We did not normalize the time series nor the surrogates to extract the Hurst exponent since we are interested in capturing any existing trends in the data. We computed both metrics $100$ times for each time series and then for its $100$ unique surrogates. Next, after checking the normality of the distribution of the results with the Shapiro-Wilk test ($\alpha=0.05$), we compared the distribution of each metric between time series vs. surrogates using either the paired t-test or the Wilcoxon signed-rank test ($\alpha=0.001$).
}

\begin{figure}[t]
    \centering
    \includegraphics[width=\columnwidth]{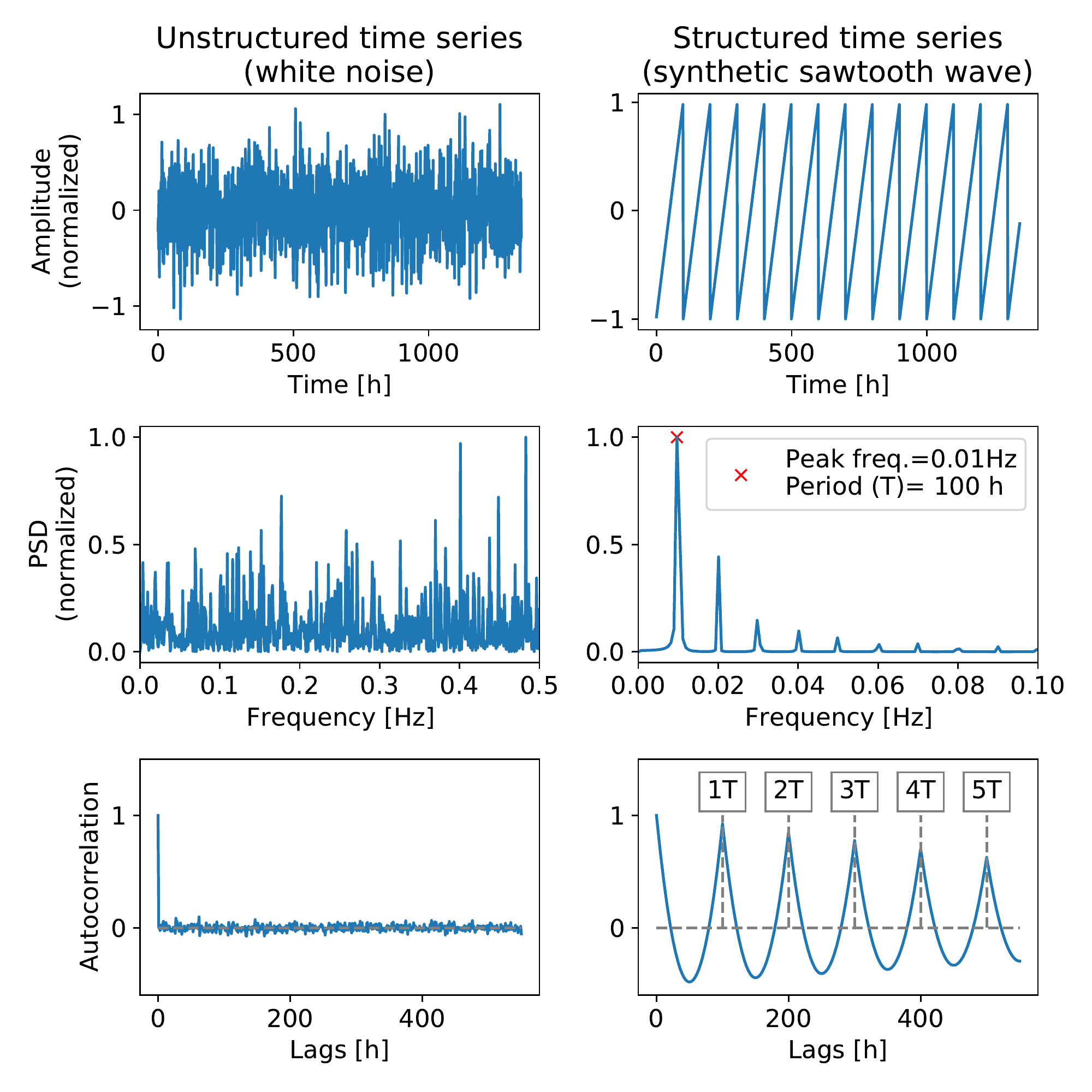}
    \caption{\changed{Periodogram and autocorrelation for an unstructured (left panel) and structured (right panel) time series.}}
    \label{fig:autocorr_periodogram}
\end{figure}

\changed{After verifying that our times series (profiles) differed from stochastic data, we attempted to identify their period using two popular signal processing techniques: the \textit{periodogram} and the \textit{autocorrelation function}~\cite{box2015time}. The former is obtained through the discrete Fourier transform, a method for expressing a time series as a sum of periodic components, and contains an approximation of the power spectral density (PSD, i.e., the power over different frequency values) of that time series. As illustrated in Fig.~\ref{fig:autocorr_periodogram}, the periodogram of a stochastic time series contains many components spread over different frequency values, thus not clearly exhibiting a main component.
Conversely, the periodogram of a structured time series exhibits clear peaks for the main frequency ($0.01$Hz in the given example) and its multiples (or harmonics). The frequency value containing the highest spectral power (aka peak frequency, $f_p$) is associated with the strongest periodic component of the signal through the relation $T=1/f_p$. We therefore estimated the period of each of our time series as the inverse of the peak frequency in its periodogram~\cite{box2015time}.}


\changed{The second technique we used to estimate the period of our time series and validate the results obtained with the periodogram was the autocorrelation. The autocorrelation measures the self-similarity of a given time series over different delay times or lags ($\tau$), ranging from $-1$ (perfectly negative autocorrelation) to $1$ (perfectly positive autocorrelation)~\cite{box2015time}.} 
\changed{As illustrated in Fig.~\ref{fig:autocorr_periodogram}, for a stochastic (unstructured) time series, the autocorrelation is nearly zero for all values of lags because the data points are uncorrelated in time. On the other hand, for a structured time series with period $T$ ($T=100$ hours in the given example), the autocorrelation exhibits peaks for $\tau=k.T$ where $k=\{1,2,3,...\}$, and its amplitude decreases as the lag increases. A common way to estimate the period of a given time series is through the examination of $N$ consecutive peaks in the autocorrelation function. In our analysis, we examined whether the five first peaks of each time series autocorrelation matched the period found through the periodogram. In other words, we verified whether the autocorrelation exhibited peaks for $\tau=k.T$ where $k=\{1,2,...,5\}$.}

\subsection{Machine Learning Analysis}

Our ML experiments aimed to \changed{assess profile uniqueness \textbf{(RQ2)}} and identify top features \changed{in distinguishing among profiles \textbf{(RQ3)}.}


\subsubsection{Feature Extraction}


An important aspect to be considered when \changed{analyzing} profiles \changed{with ML models} is the window of profile data needed for deciding whether the current behavior actually belongs to a certain user. 
We define this period as a \emph{sliding window of time}. 
Since our logs were collected on a minute-basis, our window size \changed{$t$} can assume any integer corresponding to the number of minutes \changed{(\ie $t=1,2,3,...$)}. 
\changed{In our strategy}, we summed the number of clicks, keystrokes, and background traffic activity \change{(all integer numbers)}, and concatenated the strings of the processes and domains used within consecutive $t$-minute windows, keeping the last timestamp of the window \changed{(Fig.~\ref{fig:sliding_window} in Appendix~\ref{appendix:feature_extraction} illustrates this process for a window size of $3$ minutes)}. \changed{Therefore,} every minute, a \changed{newly acquired} window size of $t$ minutes is analyzed by the ML classifier, which makes a prediction about whether \changed{such} window of activity belongs to a certain user. 
A disadvantage of this technique is that larger $t$ values are more susceptible to the cold-start problem {(\ie when there is a lack of initial information to start the recognition task, common in recommendation systems~\cite{jianbo2016})}. 
In our experiments, we opted to test several windows sizes to analyze the impact of small and large windows in predicting user identity based on the profiles. \changed{We tested $t=\{1,2,5,10,30,60\}$ minutes}.
    
Given that the numerical features (\# of clicks, keystrokes, and background traffic activity) only need to be normalized before being used by a classifier, our feature extraction process focused on the textual attributes of the matrix (list of processes and domains). For this purpose, we used TF-IDF, a statistical measure that evaluates how important a word (in our case, a process or \changed{domain name}) is to a text in relation to a collection of texts~\cite{10.5555/1394399}. Each text is represented by a sparse array that contains their TF-IDF values for each word in the vocabulary. 
\changed{Thus, each textual attribute from a window was transformed into its corresponding sparse array containing the TF-IDF values of each process and domain list (this process is also illustrated in Fig.~\ref{fig:sliding_window}, Appendix~\ref{appendix:feature_extraction})}. 
We envisioned that TF-IDF brings many advantages for applications in CA, given that (i) the more a word (\eg process) appears in an instance, the larger its feature weight is; 
(ii) the less a word appears in all files, the higher its importance is to distinguish instances; \changed{and (iii) it can be periodically updated to accommodate new processes and websites using evolving feature sets or retraining its vocabulary when, for instance, a concept drift is detected~\cite{gregioIEEEBrazil18, Xu2019}}.

By combining the sliding window technique and TF-IDF features, the classifier receives as input an array that represents the window, containing the numerical features (sum of number of clicks, keystrokes, and background traffic activity) and the TF-IDF values for the processes and domains in the window, all normalized using maximum absolute scaler~\cite{scikit-learn}. Thus, the textual attributes are used as a ``document'' that represents the user for a given moment.


\subsubsection{\changed{Profile Uniqueness Analysis (RQ2)}}

The goal of the first set of machine learning experiments was to examine whether the profiles are unique ({\bf RQ2}). 
For this purpose, we considered two types of ML models: 
(i) \textit{offline}, where the classifier is trained only once using an initial portion of the data and then performs predictions for the remaining data without retraining; and 
(ii) \textit{online}, where the classifier is periodically updated with more recent data.
Offline classifiers are more popular in the ML field, mainly in stationary distribution problems where data does not change over time (\eg object recognition). \changed{On the other hand, online models are conceptually more suitable to handle non-stationary distribution problems where data change over time}~\cite{MOA-Book-2018}. 
\changed{Since we do not know which of these scenarios best describe our profiles. For comprehensiveness, we considered both types of classifiers in our analyses.}


In both cases (offline and online learning), we conducted \changed{\textit{binary} and \textit{one-class classification} experiments. In the former, multi-class classifiers were used as binary discriminant functions, using the data from a given class (user) as positive, and the data from all the other classes (the remaining users) as negative~\cite{10.5555/1162264}. In the latter, one predictive model was created for \textit{each} user based on their data only (\ie regardless of other users' data)~\cite{han2011data}. Though popular in the ML field, multi-class classifiers were left out of our experiments due to their lack of feasibility for profile-based CA. This is because, in the multi-class setting, a single predictive model is created based on the data of \textit{all} users and then used to distinguish them among each other, thus requiring retraining every time a new user is added or removed from the group.}
Considering all variations of classifiers, parameters, number of runs, and window sizes, we trained and tested \changed{$20,460$} models in our experiments.

\medskip \noindent \textbf{Offline Classification}. 
For the offline experiments, we chose four multi-class \changed{classifiers used in the binary classification} (Random Forest, Stochastic Gradient Descent (SGD), Multi-Layer Perceptron (MLP) and LinearSVC~\cite{scikit-learn}) and two one-class (Isolation Forest, and One-Class SVM with both RBF and linear kernel). 
For the \changed{binary classifiers}, we used the default parameters in Scikit Learn~\cite{scikit-learn} for Random Forest ($100$ estimators), Stochastic Gradient Descent (hinge loss function), Multi-Layer Perceptron ($100$ hidden layers and relu activation function), and LinearSVC (Support Vector Machine linear kernel) for the SVM classifier.
To evaluate these models, we split the data (ordered by timestamps) in two sets: the first seven days of data (of each user) were used for training and the remaining for testing. 
\changed{For the binary models,} we created $\samplesize$ binary models (one for each user, where the task is to detect if the current behavior window belongs to a given user) with each multi-class classifier cited before. \changed{To evaluate the binary models we used different negative users in the training and test sets, representing a typical continuous authentication system.}
For the one-class models, we also adopted the default settings in Scikit Learn~\cite{scikit-learn} (Isolation Forest with $100$ estimators and two one-class SVM, the first with RBF kernel and the second with linear kernel). For each user, we created an outlier set containing the instances of all the other \changed{30} users, excluding the first seven days---that is, the same instances taken as test set in the \changed{binary} experiments, for a fairer comparison among one-class and \changed{binary} results.

\medskip \noindent \textbf{Online Classification.} 
For the online experiments, we selected four classifiers that support online learning: three multi-class \changed{in the binary classification} (Adaptive Random Forest; Stochastic Gradient Descent, SGD; and Perceptron~\cite{scikit-learn}) and one one-class (Half-Space Trees~\cite{10.5555/2283516.2283647}). We leveraged the default parameters in Scikit Multiflow~\cite{10.5555/3291125.3309634} and River~\cite{2020river} for both Adaptive Random Forest and Half-Space Trees~\cite{10.5555/2283516.2283647} (which was the only online one-class classifier available at the time of this writing), and the default parameters in Scikit Learn for SGD and Perceptron. The same training and test sets from the offline experiments were used. The training set was used to train the initial model and the test set to create a data stream, which was considered in a test-then-train evaluation, where each sample is tested by the model (generating a prediction used to compute the F-score) and then used to updated it~\cite{MOA-Book-2018}. 

\medskip \noindent \textbf{Evaluation.} 
\changed{As per the evaluation metrics, we included recall, precision, and F-score (the harmonic mean of precision and recall) in both offline and online classification experiments. These metrics provide a more realistic measure of a model's performance for unbalanced datasets than more popular metrics, such as accuracy~\cite{han2011data}. For each classifier, we obtained the recall, precision, and F-score values per user using the average of ten runs with different random states (for the classifiers and users in the training and test set in the binary classification experiments)~\cite{Breiman2001RF}.}
\changed{The only exception was the Adaptive Random Forest, in which we calculated the results for a single run due to time constraints (as it exhibited an excessively high training time in our experiments). We then computed the average recall, precision, and F-score results among the $\samplesize$ users for each classifier, along with the respective 95\% confidence intervals}.

\subsubsection{\changed{Top Features Analysis (RQ3)}}

To analyze the most important features \changed{in identifying each user}, we first divided their data in weeks to observe which are the most important over time. 
We trained a binary Random Forest classifier for each week of each user with a time window of $10$ minutes (the optimal value found in RQ1), resulting in $248$ models trained. 
\changed{As we found that most of the profiles repeat on a daily basis (see Sec.~\ref{sec:results}), we performed weekly training to consider several repetitions of computer usage habits to identify the most relevant features.}
\changed{Moreover, we chose the Random Forest for two reasons: it provides the importance of each feature in the classification process, and it outperformed the other tested binary models in both offline and online settings.} We then used the \changed{Gini importance property from scikit-learn's Random Forest implementation~\cite{scikit-learn}}
to explain our models' predictions for each instance of the corresponding week and user. 
Finally, we computed these values (the higher the value, the more important the feature is) for each feature and obtained their average, where we collected the top 10 for each user.





\section{\changed{Experimental Results}}
\label{sec:results}
\noindent
\textbf{\changed{Profile Temporal Consistency (RQ1)}}. \changed{In our surrogate data testing analysis, the sample entropy results obtained from the time series representing the computer usage profiles were statistically higher ($p<.001$) than the results obtained from the surrogates for all $\samplesize$ users in the two assessed conditions (with and without background activity). The difference in the sample entropy results between time series vs. surrogates was more noticeable in the first condition ($0.222\pm0$ vs. $0.714\pm0.025$) compared to the second ($0.277\pm0$ vs. $0.499\pm0.021$).}
\changed{Regarding the Hurst exponent results, statistically different ($p<.001$) values for time series vs. surrogates were observed for $29$ profiles (all users except User $5$ and User $22$) in the first condition (with background activity), and for all $\samplesize$ profiles in the second condition (without background activity).} 

\medskip
\takeaway{\changed{All $\samplesize$ profiles differed from unstructured data in terms of sample entropy in the two assessed conditions ($p<.001$). In terms of Hurst exponent results, the number of profiles that differed from unstructured data was $29$ and $31$ in the first and second conditions.}}

\begin{figure*}[htp]
\centering
\subfloat[User $9$, condition $1$]{%
  \includegraphics[clip,width=\columnwidth]{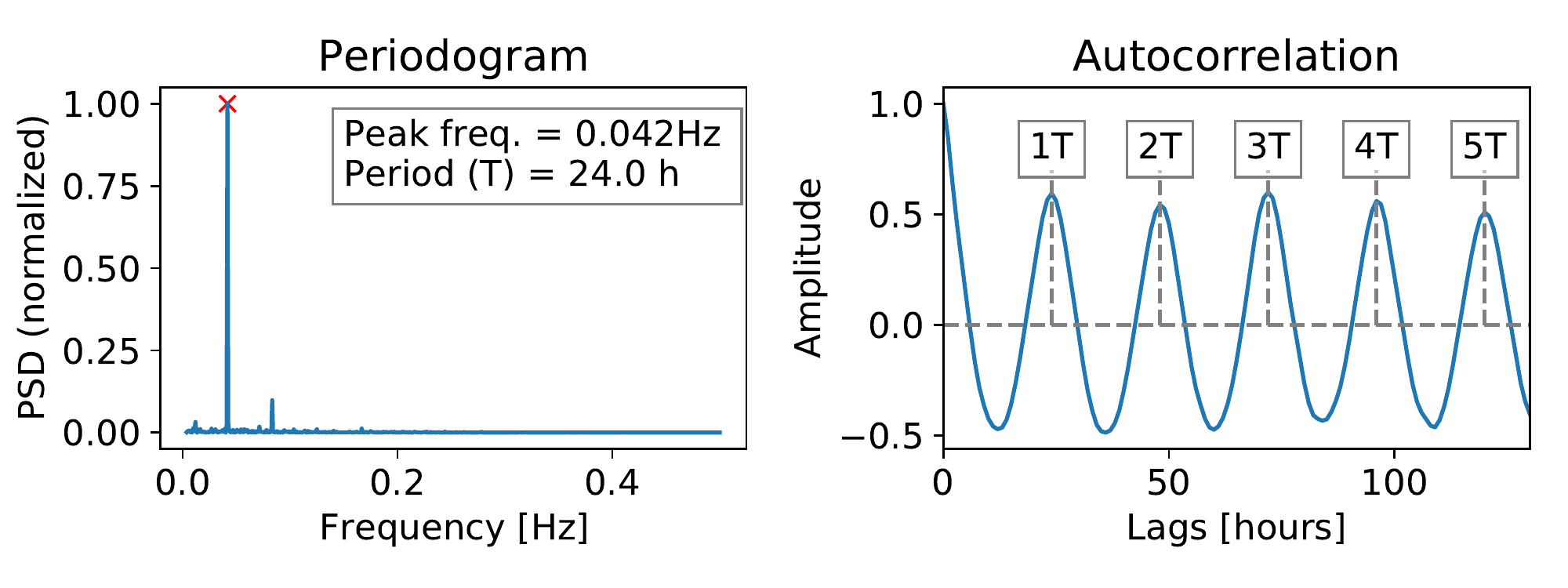}%
}
~
\subfloat[User $9$, condition $2$]{%
  \includegraphics[clip,width=\columnwidth]{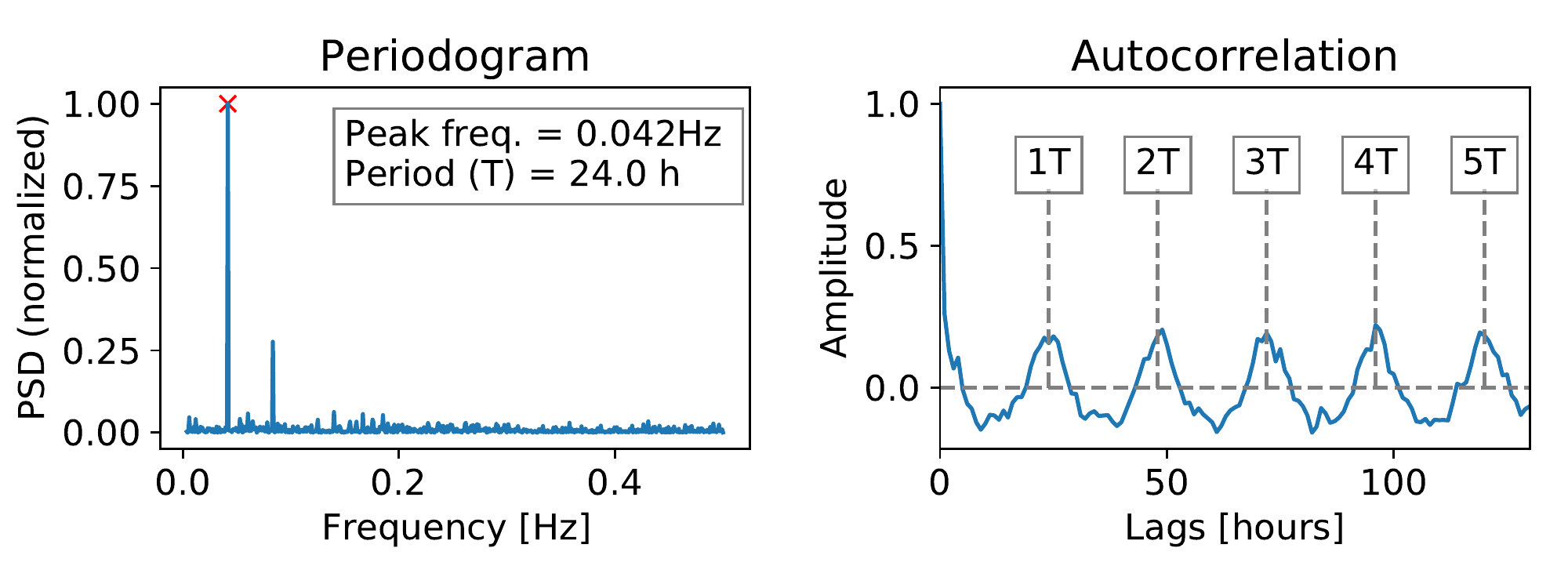}%
}

\subfloat[User $2$, condition $1$]{%
  \includegraphics[clip,width=\columnwidth]{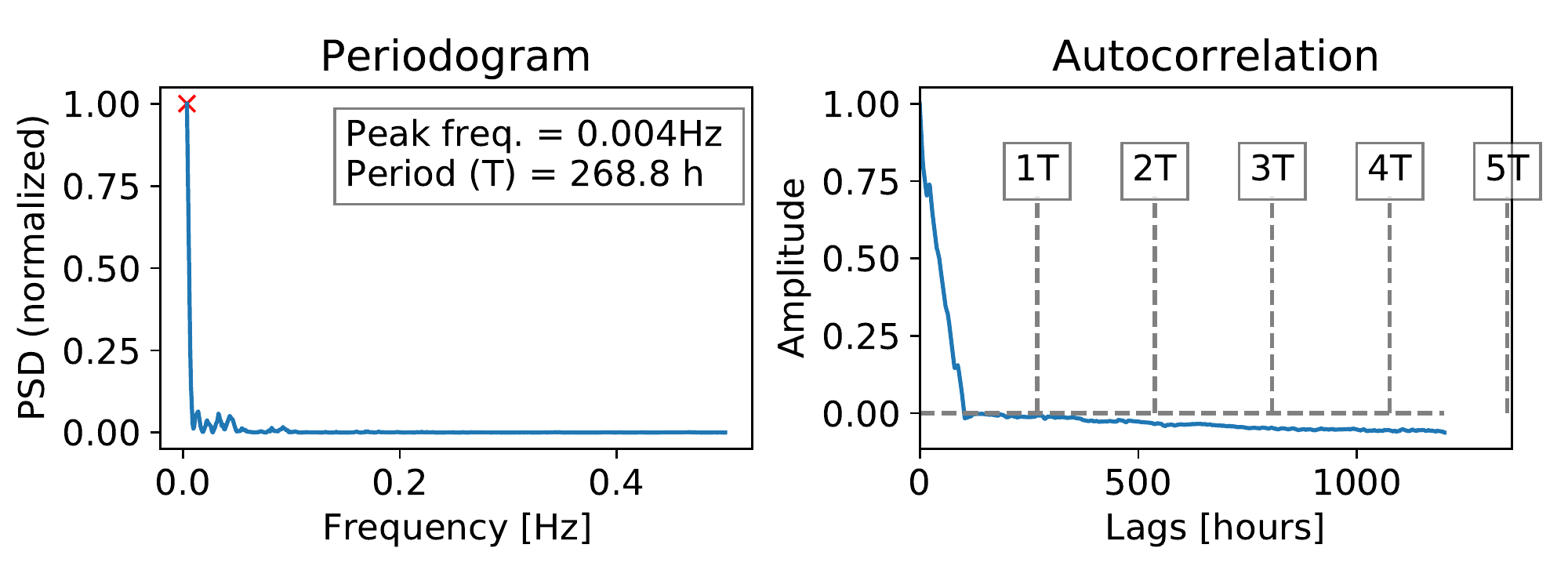}%
}
~
\subfloat[User $2$, condition $2$]{%
  \includegraphics[clip,width=\columnwidth]{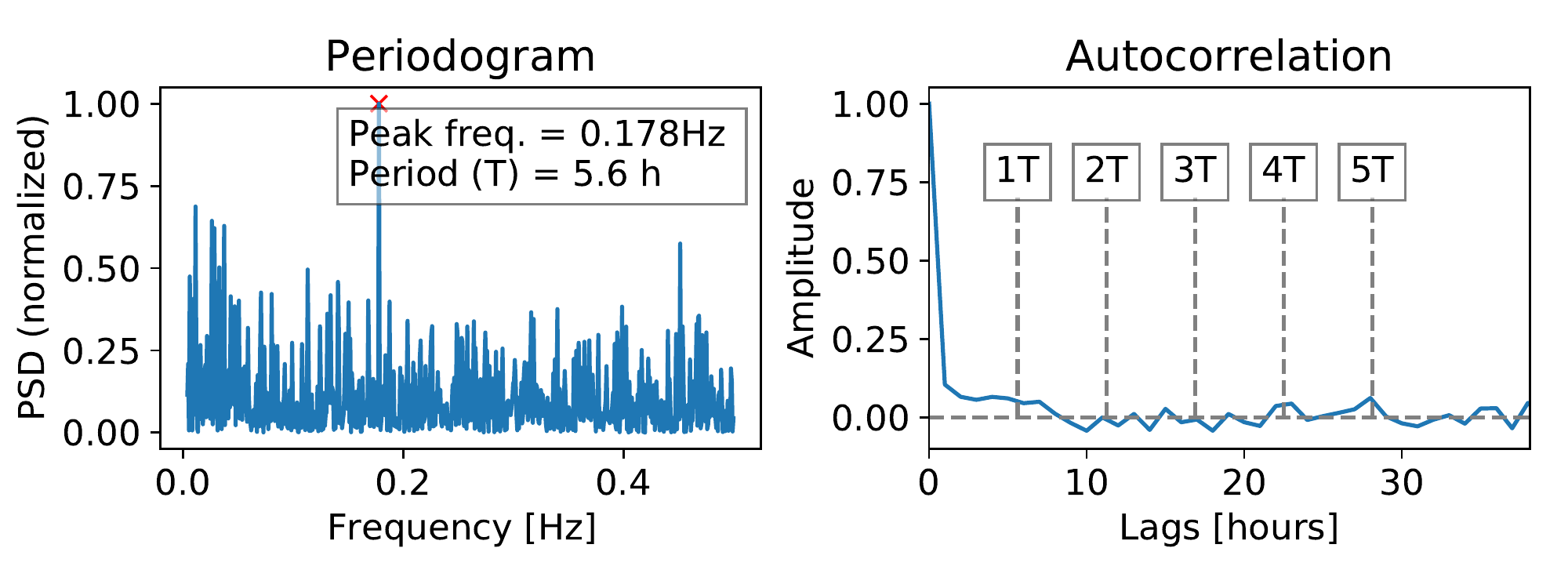}%
}

\subfloat[User $19$, condition $1$]{%
  \includegraphics[clip,width=\columnwidth]{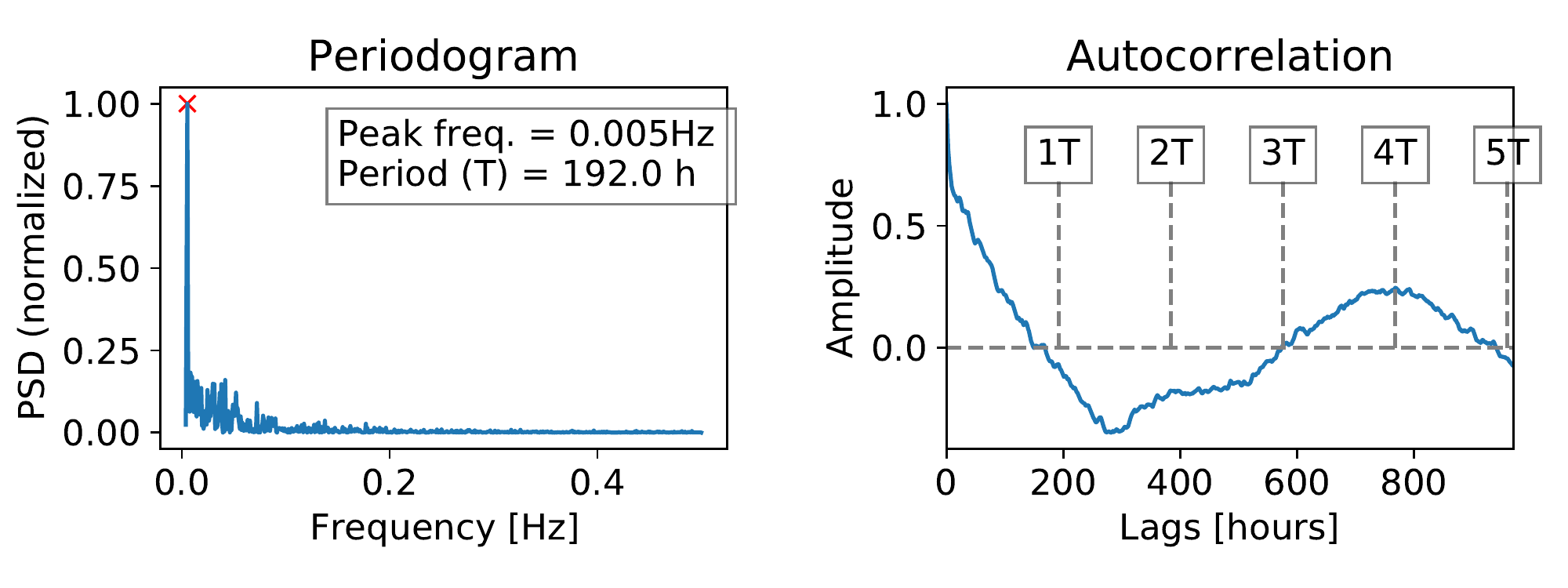}%
}
~
\subfloat[User $30$, condition $1$]{%
  \includegraphics[clip,width=\columnwidth]{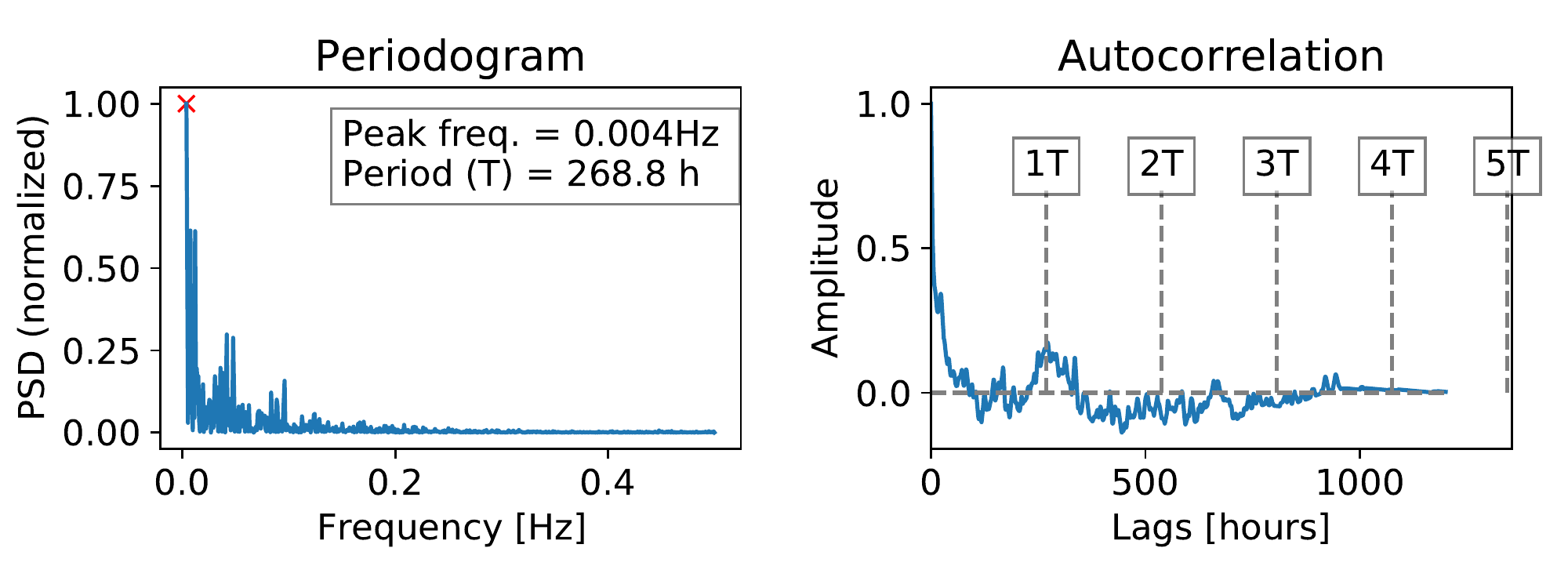}%
}

\subfloat[User $5$, condition $1$]{%
  \includegraphics[clip,width=\columnwidth]{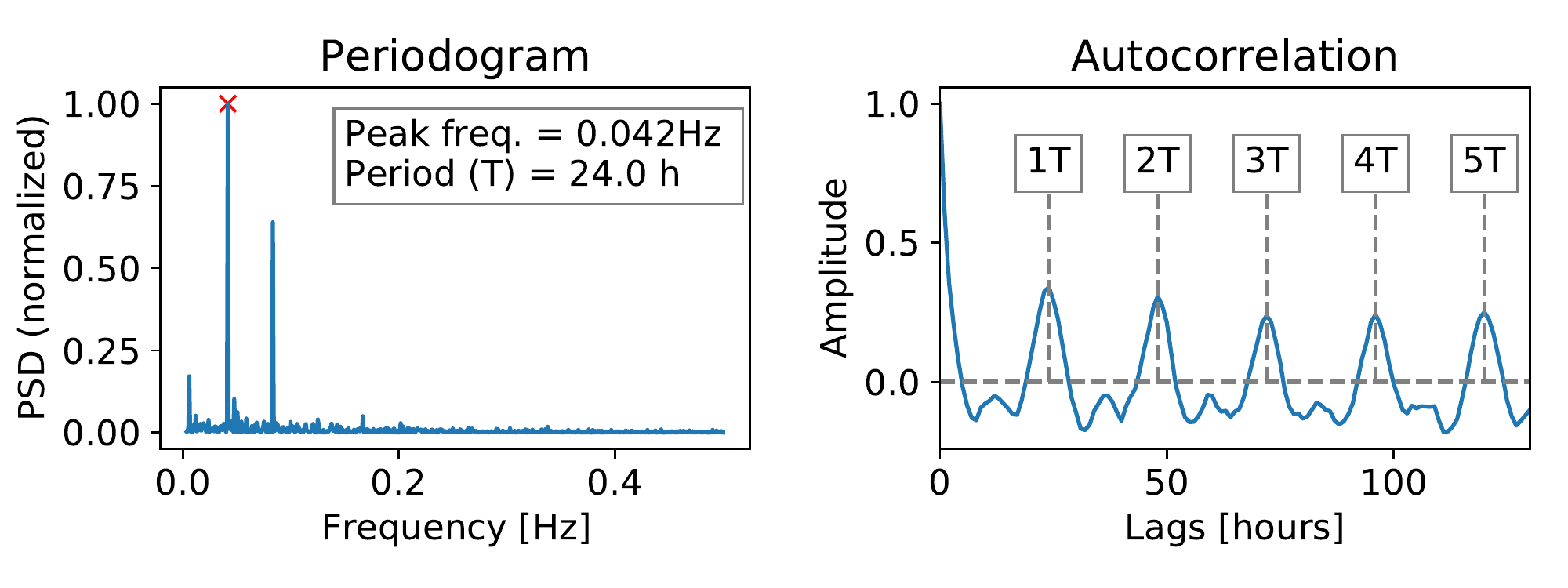}%
}
~
\subfloat[User $5$, condition $2$]{%
  \includegraphics[clip,width=\columnwidth]{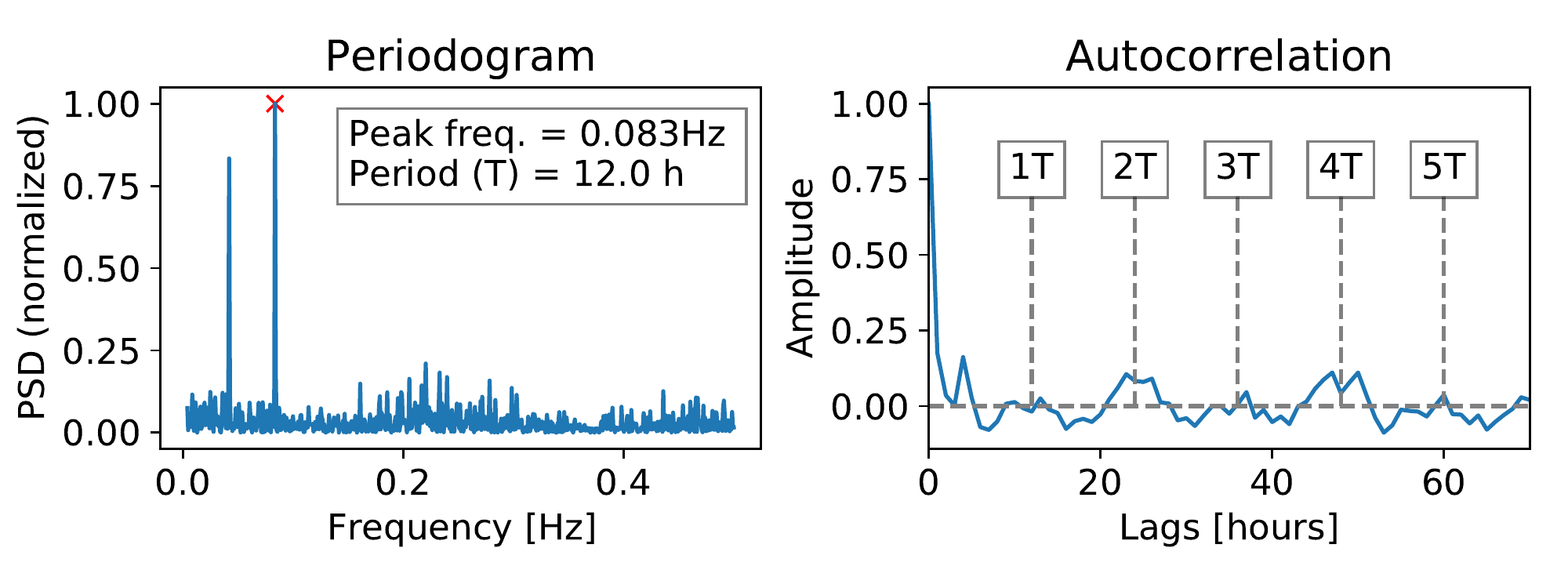}%
}


\caption{\changed{Periodogram and autocorrelation results for different users and conditions.}}
\label{fig:periodogram_autocorr_users}

\end{figure*}

\changed{We identified a period for $28$ profiles ($90.3\%$ of our sample) in the first condition (with background activity) and for $29$ profiles ($93.5\%$ of our sample) in the second condition (without background activity). In both conditions, $26$ profiles exhibited a period equal to $24$ hours, as exemplified in Fig.~\ref{fig:periodogram_autocorr_users} for User $9$. In other words, $83.9\%$ of the study participants repeated computer usage patterns daily. The periodogram and autocorrelation for all users and conditions are exhibited in Appendix~\ref{appendix:temporal_consistency} (Fig.~\ref{fig:periodogram_autocorr_all_users}).}

\changed{Figure~\ref{fig:periodogram_autocorr_users} contrasts results between the periodogram and autocorrelation for Users $2$, $19$, and $30$ in the first condition, and Users $2$ and $13$ in the second condition, preventing us from identifying a period in these cases. User $2$'s periodogram and autocorrelation were similar to the expected for unstructured data, especially in the second condition. For User $19$, the periodogram exhibited a period of $192$ hrs while the autocorrelation suggested $750$ hrs. In the remaining cases, the autocorrelation did not exhibit peaks consistent with the period found with the periodogram.}

\changed{Twenty-five profiles exhibited results consistent between the two conditions, with the period being equal to $24$ hours in all cases. In the cases where the results changed between conditions (Users $5$ and $8$), the period was found to be lower in the second condition (i.e., after removing background activity), as illustrated in Fig.~\ref{fig:periodogram_autocorr_users} for User $5$.}

\medskip
\takeaway{\changed{A period was found for $90\%$ of the profiles in Condition 1, and for $94\%$ in Condition 2. $84\%$ of the profiles exhibited a period of $24$ hours.}}

\begin{figure*}[t]
        \centering
        \begin{subfigure}[b]{.33\textwidth}
            \includegraphics[width=\textwidth]{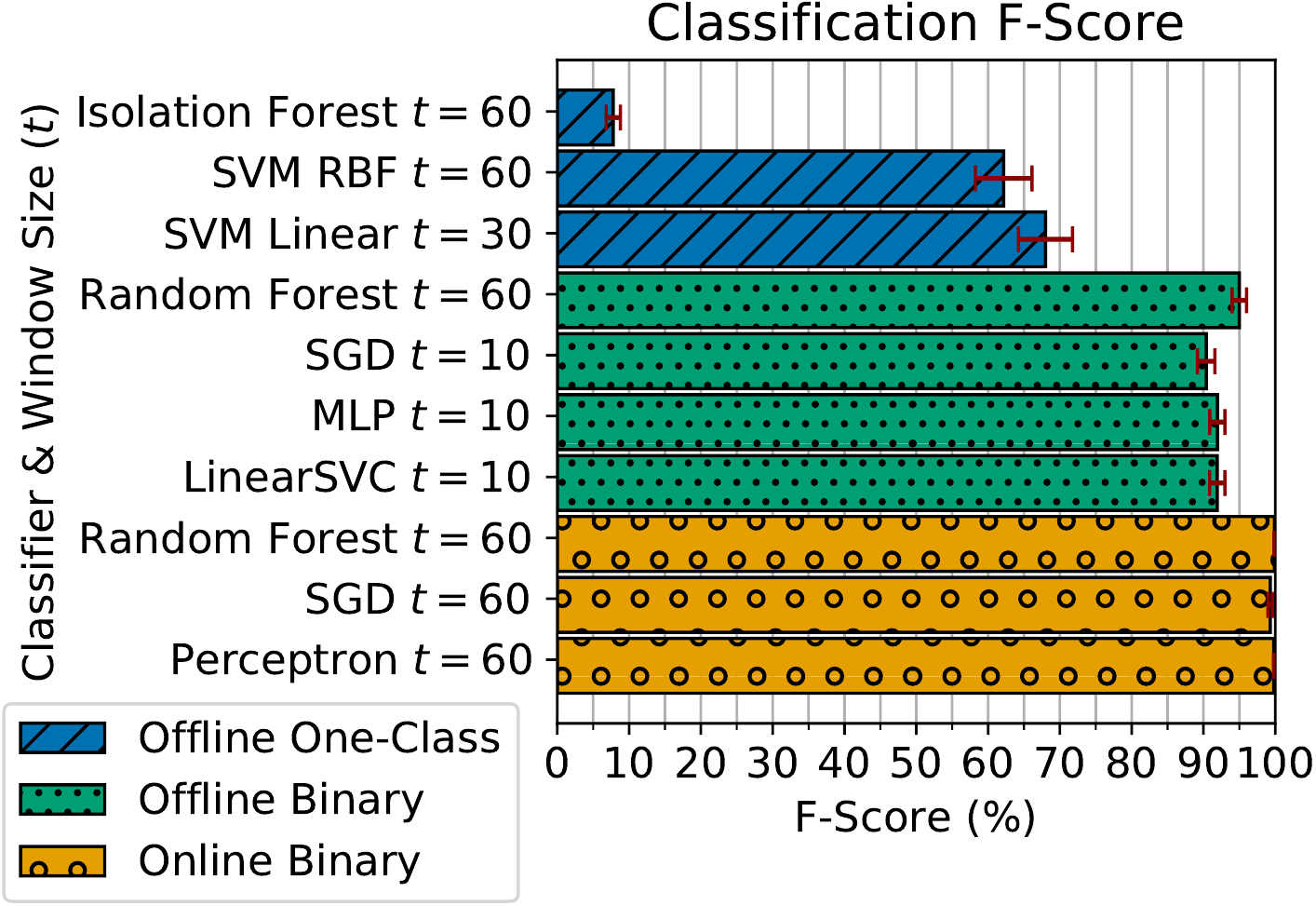}
            \caption{F-Score.}
            \label{fig:results_fscore}
        \end{subfigure}
        \begin{subfigure}[b]{.33\textwidth}
            \includegraphics[width=\textwidth]{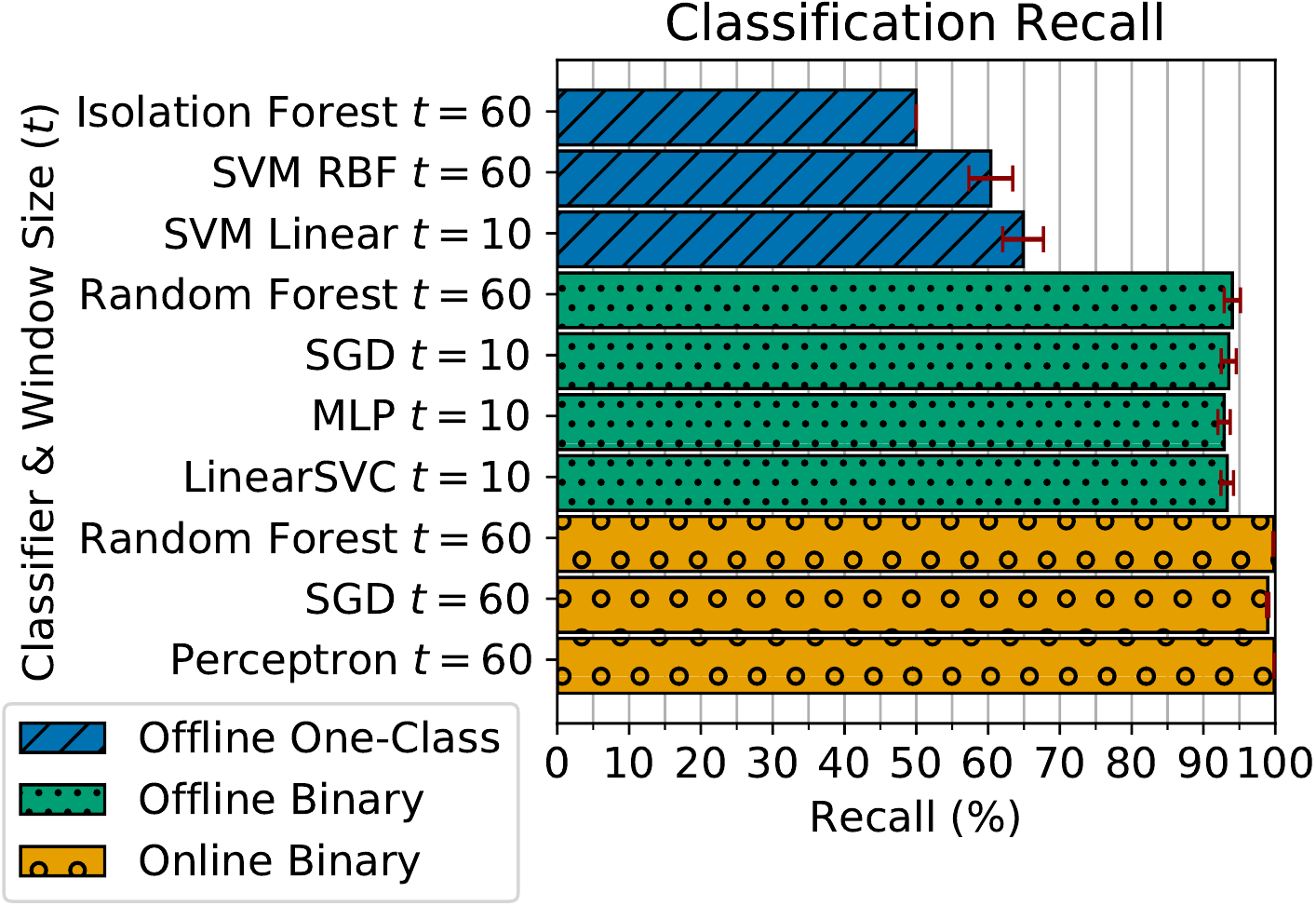}
            \caption{Recall.}
            \label{fig:results_recall}
        \end{subfigure}
        \begin{subfigure}[b]{.33\textwidth}
            \includegraphics[width=\textwidth]{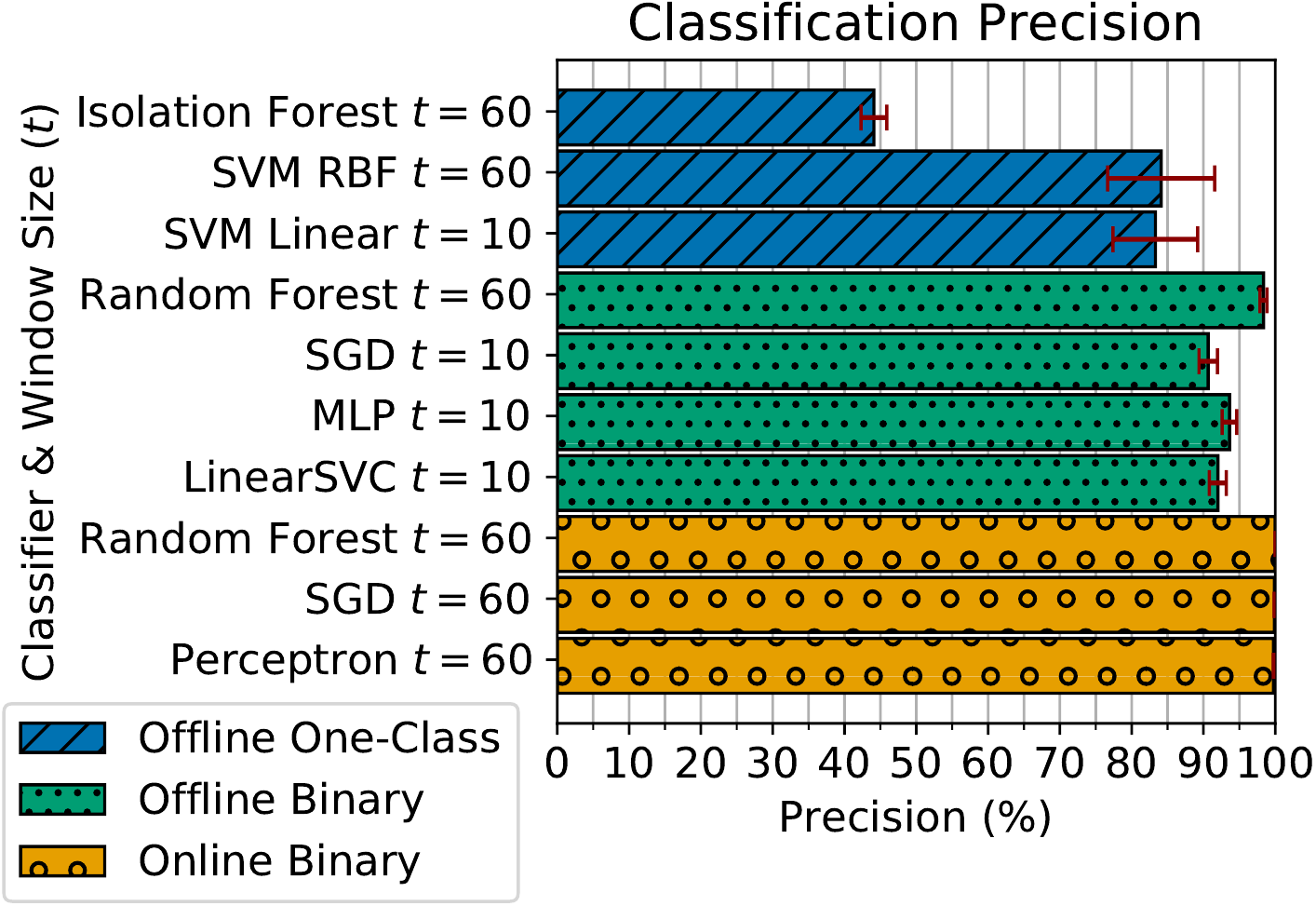}
            \caption{Precision.}
            \label{fig:results_precision}
        \end{subfigure}
    
        \caption{\changed{Top average F-score (a), recall (b), and precision (c) values in distinguishing among the computer usage profiles. The error bars denote the 95\% confidence intervals.}}
        \label{fig:results}
\end{figure*}

\noindent
\textbf{\changed{Profile Uniqueness (RQ2)}}. \changed{Figure~\ref{fig:results} shows the top results obtained to distinguish among the profiles. The complete results are exhibited in Appendix~\ref{appendix:classification_results} (Tables~\ref{tab:results_binary_offline},~\ref{tab:results_one_class_offline}, and~\ref{tab:results_online}). The online one-class classifier yielded the poorest results: $0\%$ precision, recall, and F-score; the Half-Space Trees model, the only model available, works better when anomalous data are rare, which is the opposite of our scenario~\cite{10.5555/2283516.2283647}.}

\changed{In summary, the best results were reached with the online binary models (top F-score of $99.90\%$), followed by the offline binary models (top F-score of $95.00\%$). We observed that evaluation metrics improved with the increase of the window size. Most of the binary classifiers (our top performers) achieved optimal results for $t=10$ min given that the increase of $t$ to $30$ min and $60$ min did not cause the evaluation metrics to improve substantially.}

\medskip
\takeaway{\changed{Computer usage profiles can be distinguished with high accuracy (F-score $>95\%$).}}

\noindent
\textbf{\changed{Profile Top Features (RQ3)}}. 
\changed{For $29$/$\samplesize$ users, the top 10 features were all web domains}. 
After unifying the top 10 features from each user and removing duplicates, we obtained a set of \changed{$186$} unique features in which \changed{$177$} were \changed{\emph{domains}} and only \changed{five were \emph{processes}}. The \changed{four} remaining features refer to \emph{background traffic} (which only appeared for two users), \emph{number of mouse clicks}, and \emph{number of keystrokes} (both appearing for a single user). From these \changed{$186$} top features, \changed{$45$} were observed for two or more users. 

\smallskip
\takeaway{The top features in distinguishing computer usage profiles were mostly comprised of \changed{domains.}}

\section{Discussion}
\label{sec:discussion}


\noindent
\textbf{Computer Usage Profiles Consistency.}
\changed{Our analysis of the temporal consistency of the profiles (\textbf{RQ1}) revealed that the majority of the users enrolled in our study did exhibit computer usage patterns consistent over time in the two assessed conditions (with and without background process activity). Interestingly, most of the users repeated computer usage on a daily basis over the 8-week study period. As the profiles also repeated for the integer multiples of their main period (as observed in the autocorrelation results), it is plausible to assume that such profiles also repeated every $j$ days where $j=\{1,2,...,56\}$, every $k$ weeks where $k=\{1,2,...,8\}$, and every $l$ months where $l=\{1,2\}$, within the study period. In other words, the profiles repeated not only on a daily basis, but also on a weekly and monthly basis, for example.}

\changed{We acknowledge that these results are likely to be sensitive to changes in the user's routine that alter the way they use the computer (e.g., change of role, vacation, and travel in corporate environments), which might be the reason for the lack of periodicity observed for User $2$. We therefore assume that most of our users did not have significant changes in their main occupation/tasks while in study.}

\begin{figure}[t]
    \centering
    \includegraphics[width=\columnwidth]{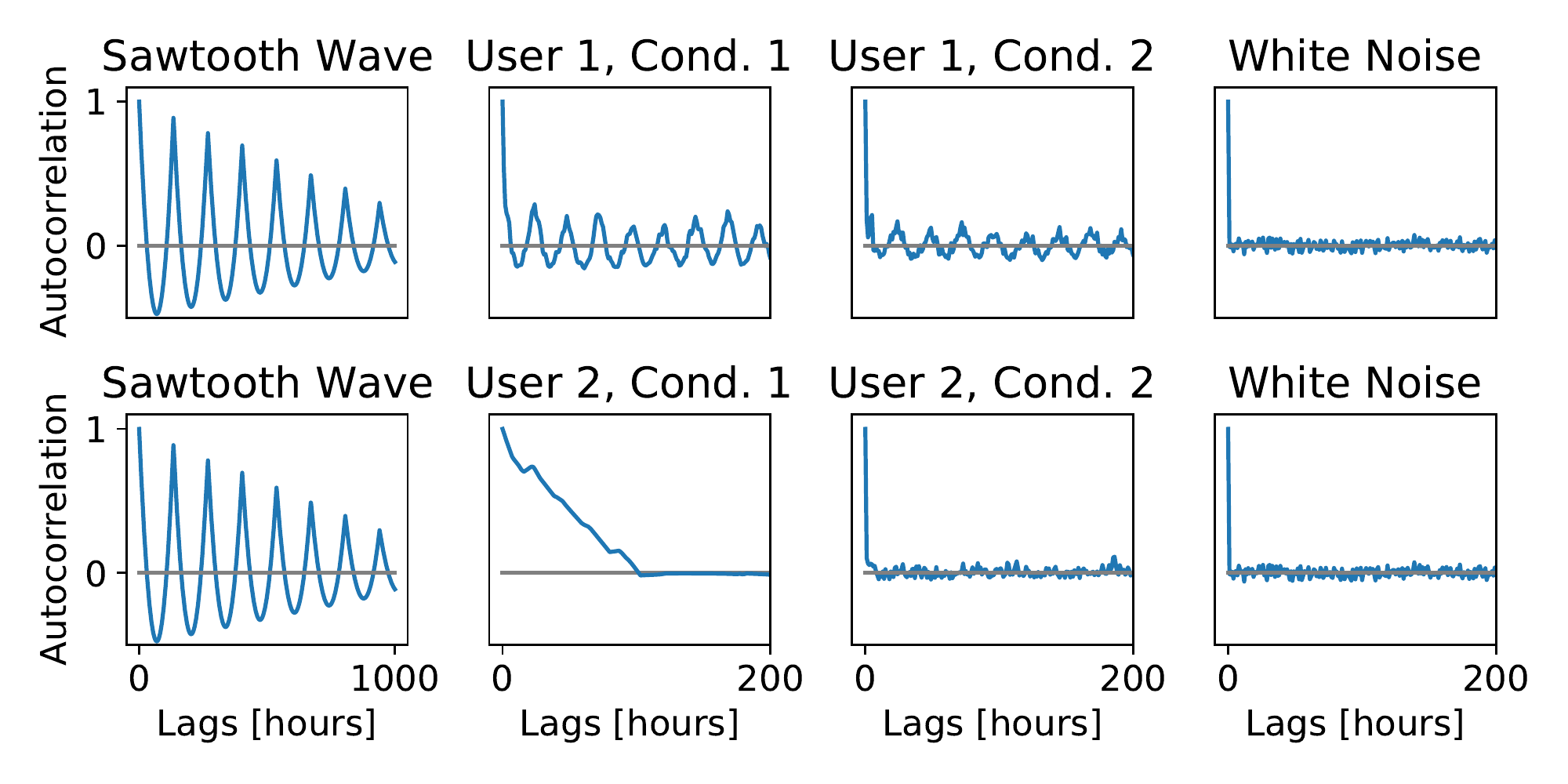}
    \caption{\changed{Comparison between the autocorrelation for purely structured data (sawtooth wave) and purely stochastic data (white noise); results from User $1$ (top) and User $2$ (bottom).}}
    \label{fig:autocorr_user1_user2}
\end{figure}

\changed{Our analysis involving the surrogate data revealed that all profiles differed from completely unstructured data in the two tested conditions, i.e., they are unlikely to be purely stochastic. 
However, our periodogram and autocorrelation results (Fig.~\ref{fig:periodogram_autocorr_all_users}) were not identical to those observed for purely structured data (Fig.~\ref{fig:autocorr_periodogram}). Our results look similar to the expected for structured time series contaminated with random noise. In both conditions, most profiles exhibited a periodogram and autocorrelation more similar to the expected for structured data (\eg User $1$), suggesting lower presence of noise. On the other hand, results more similar to the expected for unstructured data were also observed for some users (\eg User $2$), suggesting higher contamination by noise (Fig.~\ref{fig:autocorr_user1_user2}). In other words, users exhibited different levels of correlation in time in their use patterns.}

\changed{These findings suggest that profiles can be described as a combination of two components, one \emph{deterministic} (most likely nonlinear) and one \emph{stochastic}, with the latter being stronger for some users than others. In other words, it seems reasonable to infer that profiles are consistent over time, but yet subject to random factors that decrease their temporal correlations and make them less structured. 
}

\changed{Another factor that may have caused random drifts in the computer usage behaviors of our users is the current COVID-19 pandemic (our data collection happened during the months of the pandemic). The pandemic has greatly altered how users in general work, learn, socialize, and entertain themselves using their personal devices~\cite{Koeze2020-zm}, likely causing them to use the same computer for personal and work tasks. Several activities of daily life started to take place virtually due to safety concerns, such as grocery shopping, bank services, classes in general (\eg schools, universities, gyms), watching movies or concerts etc., and many of them may not happen on a regular basis (\eg someone might buy groceries twice a week for two weeks and then spend three weeks without buying anything). This has the potential to strengthen the stochastic component in the profiles.}

\changed{Importantly, when moving from the first condition (with background activity) to the second one (without background activity), we noticed a decrease in the peaks of the autocorrelation function (Fig.~\ref{fig:periodogram_autocorr_all_users}). We also observed an increase in the average sample entropy result (from $0.222\pm0.114$ to $0.277\pm0.124$), as well as a decrease in the average Hurst exponent result towards $0.5$ (from $0.608\pm0.104$ to $0.564\pm0.084$). All these results indicate that the removal of the background activity caused the profiles to be less correlated in time. This finding is aligned with our hypothesis that the background activity contains periodic events that can strengthen the deterministic component in the profiles. However, some of these events may be produced by the OS without the knowledge of the user and according to a predefined schedule (\eg automatic software update), which might make them easy to predict/mimic. Thus, it might be possible to increase the robustness of CA tools by leaving background process activity out of the computer usage profiles.}

\medskip
\takeaway{\changed{The profiles were mostly consistent, with users repeating habits on a daily basis, but subject to random factors decreasing their temporal consistency.}}

\smallskip
\noindent
\textbf{Profiles Uniqueness}. Our results from \changed{\textbf{RQ2}} partially supported the hypothesis that profiles comprised of process-, network-, mouse-, and keystrokes-related events can uniquely characterize computer users. While most of the tested \changed{binary} classifiers yielded promising results, 
the one-class models achieved poor results (in both offline and online settings). 
In other words, our profiles can be considered unique depending on the learning model used.
\changed{Importantly, we trained our classifiers using one week of data, which potentially contained seven repetitions of computer usage habits given that we found most of the profiles to repeat on a daily basis (\textbf{RQ1}). This may explain the promising results achieved with the binary classifiers. Using less training data can allow the CA solution to be implemented in a more timely manner, but may also affect its performance since less usage patterns would be available for the learning model.}

Our top feature analysis \textbf{(RQ2)} revealed that the set of domains accessed by the users was \changed{significantly more relevant to distinguish profiles, than other process- and keystrokes-related features.}
\changed{This finding suggests that collecting only network-related events can potentially lead to unique profiles. 
However, it is important to remark that our data was collected with users \changed{who experienced no restrictions on network activity, contrary to some corporate environment that restrict network activity, such as social media sites.} 
Restricting the list of domains that users can access will potentially decrease profile uniqueness and impose challenges to profile-based CA systems. In this case, enriching a profile with process-, mouse-, and/or keystroke-related events may help to increase profile uniqueness and improve user recognition with CA. }




\medskip
\takeaway{Computer usage profiles can uniquely characterize computer users when analyzed with \changed{binary classifiers}. Network-related events constituted the most important feature to construct unique profiles.}

\smallskip
\noindent
\textbf{Challenges and Opportunities of Using Computer Usage Profiling for Applications in CA}.
Regarding the feasibility of leveraging computer usage profiles for CA applications, our data suggests that \changed{binary models are better suited for profile-based CA tools than one-class models; while both allow the training on a per-user basis, which is advantageous for corporations due to employee turnover/hiring, the binary models were able to identify the profiles more accurately than the one-class ones. Moreover, our results suggest that online binary models should be preferred over offline ones due to better performance and robustness to temporal changes in profiles.}

\changed{One limitation of leveraging profiles for CA is that a certain amount of data is necessary to start recognizing users (a.k.a. cold-start problem)}. We found that a time window of $10$ minutes was the optimal trade-off to recognize our dataset of profiles \changed{with the binary models (our top performers), but this optimal value might change for other datasets}. This may be a challenge for profile-based CA tools given that the longer this optimal time window, the more susceptible the system is to a cold-start problem, causing the device to be initially vulnerable for a longer period of time. Moreover, the longer it takes for the system to recognize an anomaly, the less value the CA system might have for corporate security, because the anomaly could correspond to a malicious activity whose effects persist after detection.

\subsection{Limitations and Future Work}
\label{sec:limitations}


One limitation of our study lies in our sample, which is small \changed{($N = \samplesize$)} and comprised mainly of young adults from a large university (\ie not representative of all possible environments). Although a large university of nearly $50$K students \changed{may be somewhat similar to} an organization, corporate users likely present different computer usage profiles compared to university students. 
These factors thus limit the generalizability of our findings. 
Future studies with more diverse and representative samples are warranted. 

Importantly, we collected our data during the COVID-19 pandemic, which might have affected user behavior and, consequently, influenced our conclusions towards profiles' uniqueness and consistency. 
The yet unforeseeable changes to how our society may shift after the pandemic is unresolved and how this may impact the user profiling analyses presented in this paper will require investigations from future work (\eg will there be more sparse personal computer usage as users return to work in offices and classrooms?).

Another limitation of our study concerns the integrity of the systems in which computer usage data was recorded: we assumed that the users' systems were not compromised by malware during the study period. The presence of malware could have adversely affected the learning models, as malicious behaviors would be considered part of the usage profile. In addition, although we debriefed all participants upon study completion to confirm that they did not share their computers while in study, some self-reported responses might not be accurate.

\changed{Regarding our analysis of profiles' temporal consistency (RQ1), we focused on finding a period for each profile using well known techniques. Future research is needed to investigate whether profiles indeed posses a deterministic component combined with a stochastic one, as we concluded based on our results. Given the evidence of nonlinear dynamics in the profiles that we found via surrogate data testing, futures studies might benefit from using other non-linear metrics/techniques to better understand the behavior of computer usage patterns. For instance, one may try to reconstruct the phase space (or attractor) and investigate its properties through the correlation dimension (or fractal dimension) and Lyapunov exponents.}

\changed{In our classification experiments (RQ2 and RQ3)}, we employed a time window. \changed{This may constitute a limitation because}, in some cases, when the behavior is about to change, the classifier may have difficult in detecting the shift because the window will gradually change with time, especially when its size is large ($60$ minutes, for instance). Future work addressing this issue is needed to improve classification performance. Moreover, we only considered traditional metrics (precision, recall, and F-score) to evaluate the performance of the tested classifiers. \changed{Future studies focusing on CA systems are advised to consider other evaluation metrics that are better suited for this case}, such as frequency count of scores along with receiver operating characteristics (ROC) curve~\cite{sugrim2019robust}.

\section{Conclusions}
\label{sec:conclusions}

We conducted an ecologically-valid user study with $\samplesize$ computer users to systematically investigate whether computer usage profiles comprised of process-,  mouse-, keystrokes-, and network-related events are consistent over time and unique. Additionally, we investigated challenges and opportunities of using such profiles in applications of CA. 

\changed{Though we found evidence of temporal consistency for most of the profiles within the study period—with most of them reoccurring every 24 hours—our results suggest that these profiles experienced variations over time, due to factors such as days off.  This suggests that online ML models (which allow for periodical retraining) might be better suited for a profile- based CA tool than offline models (where training occurs only once).} 

In our comprehensive set of experiments, binary classifiers (online and offline) were able to accurately recognize the profiles, indicating that computer usage profiling can uniquely characterize computer users. \changed{Network domains} accessed by users were more relevant in recognizing them than the \changed{keyboard and mouse activity}.
Furthermore, online binary models exhibited better performance and robustness in handling temporal changes in profiles.

\ifCLASSOPTIONcompsoc
  \section*{Acknowledgments}
\else
  \section*{Acknowledgment}
\fi
\changed{This work was supported by NSF award 1815557. This material is based upon work supported by (while serving at) the National Science Foundation. We would like to acknowledge Arpitha Nagaraj Hegde for contributing to the development of our extractor module, Shlok Gilda for his advice on how to encode the computer usage data into time series, and Natalie Ebner for discussions related to an earlier version of our planned user study.}


\bibliographystyle{IEEEtran}
\bibliography{bibs/daniela.bib,bibs/eager.bib,bibs/profhyp.bib}


%

\vskip -2.8\baselineskip plus -0.7fil

\begin{IEEEbiographynophoto}{Luiz~Giovanini}
	is a Postdoc Researcher at the Department of Electrical and Computer Engineering (ECE) at the University of Florida (UF). 
	His current research interests include machine learning for applications in the cyber security and biomedical fields.
\end{IEEEbiographynophoto}

\vskip -2.8\baselineskip plus -0.7fil

\begin{IEEEbiographynophoto}{Fabrício~Ceschin}
    is a PhD candidate at the Department of Informatics at the Federal University of Parana, Brazil. His research interests include machine learning, adversarial machine learning, and data streams applied to cyber security. He has received support from the Google Research Awards for the Latin America program.
\end{IEEEbiographynophoto}

\vskip -2.8\baselineskip plus -0.7fil

\begin{IEEEbiographynophoto}{Mirela~Silva}
    is a PhD candidate at the Department of ECE at the UF. Her research interests include interdisciplinary computer privacy, focused heavily on the intersection of cyber deception and abuse through the lens of interindividual differences.
\end{IEEEbiographynophoto}

\vskip -2.8\baselineskip plus -0.7fil

\begin{IEEEbiographynophoto}{Aokun~Chen}
    is a Postdoctoral Researcher at the Department of Health Outcomes \& Biomedical Informatics at UF. He received his Ph.D. degree from the ECE Department at UF. His current research interests include machine learning for bioinformatics and cybersecurity
\end{IEEEbiographynophoto}

\vskip -2.8\baselineskip plus -0.7fil

\begin{IEEEbiographynophoto}{Sanjay~Banda}
    is an MSc. Student at the Department of Computer and Information Science and Engineering (CISE) at UF.  His current research interests Distributed Systems and Content Delivery Networks.
\end{IEEEbiographynophoto}

\vskip -2.8\baselineskip plus -0.7fil

\begin{IEEEbiographynophoto}{Madison~Lysaght}
    received her Bachelor’s degree from UF. Her research interests includes human computer interaction focused around design thinking for cyber deception.
\end{IEEEbiographynophoto}

\vskip -2.8\baselineskip plus -0.7fil

\begin{IEEEbiographynophoto}{Heng~Qiao}
    is a Ph.D. candidate at the Department of ECE at UF. His research interests include machine learning and anomaly detection.
\end{IEEEbiographynophoto}

\vskip -2.8\baselineskip plus -0.7fil

\begin{IEEEbiographynophoto}{Nikolaos~Sapountzis}
    is a Senior Software Engineer at Cisco. He received his PhD from the EURECOM with focus on 5G networks. His research interests include network optimization, wireless systems, as well as privacy and security.
\end{IEEEbiographynophoto}

\vskip -2.8\baselineskip plus -0.7fil

\begin{IEEEbiographynophoto}{Ramchandra~Kulkarni}
    is a Software Development Engineer with Query Processing team at Redshift, AWS. He received his MSc degree in Computer Science Department of CISE at UF. 
\end{IEEEbiographynophoto}

\vskip -2.8\baselineskip plus -0.7fil

\begin{IEEEbiographynophoto}{Ruimin~Sun}
    is a Postdoc Researcher in Northeastern University. She has Ph.D from the Department of ECE at UF. Her research interests include cyber physical system security, machine learning model privacy, and malware detection. 
\end{IEEEbiographynophoto}

\vskip -2.8\baselineskip plus -0.7fil

\begin{IEEEbiographynophoto}{Brandon~Matthews}
    is Senior Researcher for the Electro-Optical Systems Laboratory at GTRI. 
    He has more than 13 years of experience, largely focused on innovative modern electronic warfare applications. 
\end{IEEEbiographynophoto}

\vskip -2.8\baselineskip plus -0.7fil

\begin{IEEEbiographynophoto}{Dapeng~Oliver~Wu}
    is a professor at the Department of ECE at the UF. His research interests are in the areas of networking, communications, signal processing, computer vision, machine learning, smart grid, and information and network security. He has served as Editor in Chief of IEEE Transactions on Network Science and Engineering.
\end{IEEEbiographynophoto}

\vskip -2.8\baselineskip plus -0.7fil

\begin{IEEEbiographynophoto}{André~Grégio}
    is  an  Assistant  Professor  of  the  Department  of  Informatics at the UFPR, Brazil. He received his PhD in Computer Engineering from Unicamp, Brazil. His research interests include malware countermeasures and machine learning techniques applied to security data.
\end{IEEEbiographynophoto}

\vskip -2.8\baselineskip plus -0.7fil

\begin{IEEEbiographynophoto}{Daniela~Oliveira}
    is an Associate Professor in the Department of ECE at UF. Her current research interests are human factors in cyber security, phishing, and disinformation.
\end{IEEEbiographynophoto}





\clearpage
\label{sec:appendix}

%
%

\appendices

\section{Study Participants' Demographics}
\label{appendix:demographics}
\label{sec:appendix_demographics}

\changed{Table~\ref{tab:user_demographics} summarizes the demographics of our $\samplesize$ study participants.}


\begin{table}[ht]
\centering
\caption{Summary of study participants' demographics.}
\label{tab:user_demographics}
\resizebox{0.9\linewidth}{!}{%
\begin{tabular}{clc}
\toprule
\textbf{Category} &
  \textbf{Metric} &
  \textbf{\begin{tabular}[c]{@{}c@{}}Total \\ (N = 31)\end{tabular}} \\ \midrule
\multirow{2}{*}{\textbf{Gender}}    & Female                                                                   & 20 (64.52\%) \\
                                    & Male                                                                     & 10 (32.26\%) \\
                                    & \begin{tabular}[c]{@{}c@{}}Gender Variant/\\Non-Conforming\end{tabular} & 1 (3.23\%)   \\ 
                                    \hdashline
\multirow{4}{*}{\textbf{Age}}    
                                    & 18--25 years          & 16 (46.43\%) \\
                                    & 26--35 years          & 12 (39.28\%) \\
                                    & 36--45 years          & 2 (7.14\%) \\
                                    & $\geq$46 years        & 2 (7.14\%)\\ 
                                    \hdashline
\multirow{5}{*}{\textbf{\begin{tabular}[c]{@{}c@{}}Highest \\ Formal\\ Degree\end{tabular}}} &
  Associate &
  3 (9.68\%) \\
                                    & Bachelor's                                                                 & 11 (35.48\%) \\
                                    & Master's                                                                   & 0 (0.00\%)   \\
                                    & PhD/Doctorate                                                           & 5 (16.13\%)  \\
                                    & Other                                                                    & 12 (38.71\%) \\ \hdashline
\multirow{4}{*}{\textbf{\begin{tabular}[c]{@{}c@{}}Marital \\ Status\end{tabular}}} &
  Single &
  7 (22.58\%) \\
                                    & Married                                                                  & 11 (35.48\%) \\
                                    & Divorced                                                                 & 3 (9.68\%)  \\
                                    & In a relationship & 10 (32.26\%)  \\
                                    \hdashline
\multirow{5}{*}{\textbf{\begin{tabular}[c]{@{}c@{}}Living \\ Condition\end{tabular}}} &
  Alone &
  5 (16.13\%) \\
 & With spouse/S.O.* &
  13 (41.94\%) \\
                                    & With child(ren)                                                          & 3 (9.68\%)  \\
                                    & Assisted living                                                          & 0 (0.00\%)   \\
                                    & Other                                                                    & 10 (32.26\%)  \\ \hdashline
\multirow{3}{*}{\textbf{\begin{tabular}[c]{@{}c@{}}Employment \\ Status\end{tabular}}} &
  Employed &
  17 (54.84\%) \\
                                    & Unemployed                                                               & 14 (45.16\%) \\
                                    & Retired                                                                  & 0 (0.00\%)   \\ \hdashline
\multirow{3}{*}{\textbf{\begin{tabular}[c]{@{}c@{}}Yearly \\ Household \\ Income\end{tabular}}} &
  $<$\$10,000 &
  9 (29.03\%) \\
                                    & \$10,000 to $<$\$50,000     & 10 (32.26\%) \\
                                    & \$50,000 to $<$\$100,000    & 6 (19.35\%) \\
                                    & $\geq$\$100,000           & 6 (19.35\%) \\
                                    \hdashline
\multirow{2}{*}{\textbf{\begin{tabular}[c]{@{}c@{}}Latino \\ Ethnicity\end{tabular}}} &
  Not Hispanic/Latino &
  23 (74.19\%) \\
                                    & Hispanic/Latino                                                          & 8 (25.81\%)  \\ \hdashline
\multirow{4}{*}{\textbf{\begin{tabular}[c]{@{}c@{}}Primary \\ Language\end{tabular}}} &
  English &
  24 (77.42\%) \\
                                    & Portuguese                                                               & 4 (12.90\%)  \\
                                    & Sinhala                                                                  & 2 (6.45\%)   \\
                                    & Chinese                                                                  & 1 (3.23\%)   \\ \toprule
\multicolumn{3}{l}{\begin{tabular}[c]{@{}c@{}}$\ast$ S.O. = significant other\end{tabular}}
\end{tabular}%
}
\end{table}
\section{Feature Extraction Details}
\label{appendix:feature_extraction}
\label{sec:feature_extraction}

\begin{figure*}[ht]
    \centering
    \includegraphics[width=.9\textwidth]{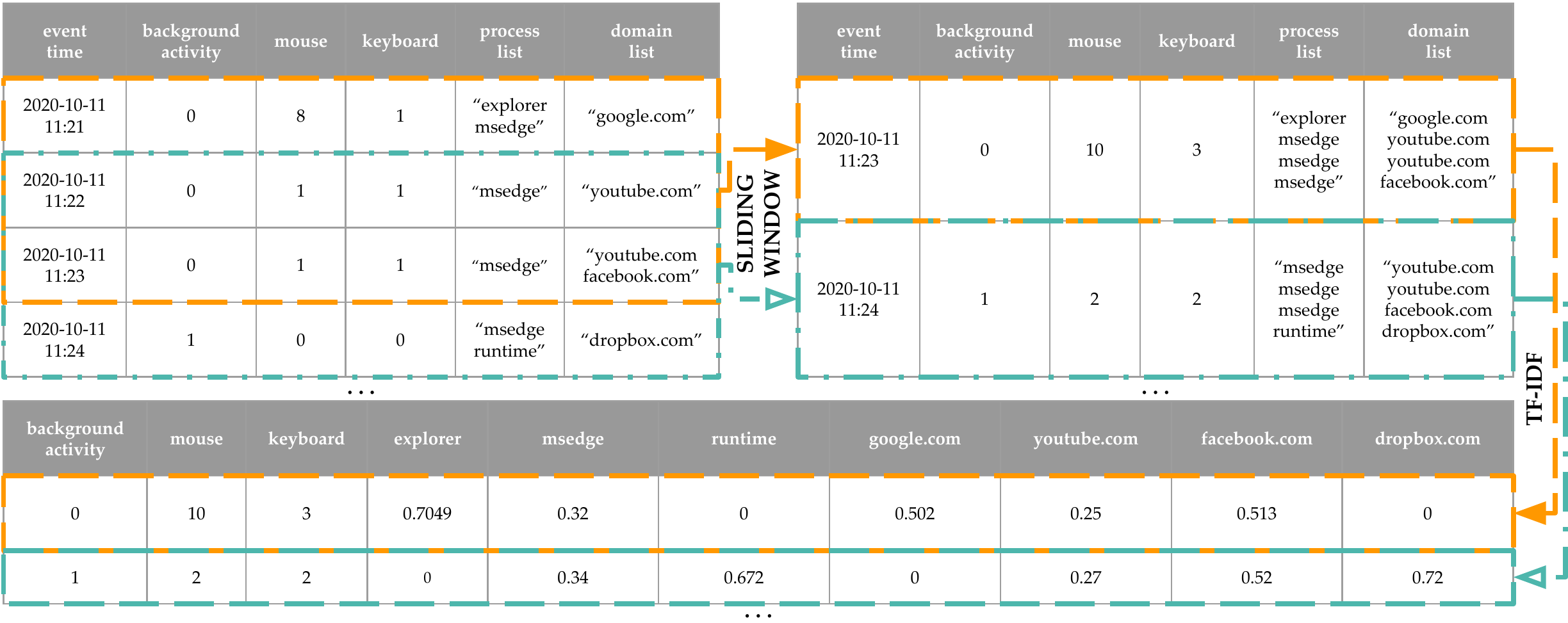}
    \caption{\changed{Example of sliding window feature extraction with window size $t=3$ minutes (illustrative TF-IDF values).}}
    \label{fig:sliding_window}
\end{figure*}

\changed{Figure~\ref{fig:sliding_window} illustrates the sliding window strategy used in our ML experiments for a window size of 3 minutes. As shown, we summed the number of clicks, keystrokes, and background traffic activity (all integer numbers), and concatenated the strings of the processes and domains used within consecutive $t$-minute windows ($t=3$ in the given example), keeping the last timestamp of the window. For an original matrix containing $N$ lines of active minutes, this process groups computer usage in $t$-minute windows.}

\section{Surrogate Data Testing Results}
\label{appendix:surrogate_testing_results}

\changed{Table~\ref{tab:entropy-hurst} exhibits the results of the statistical comparison for the sample entropy and Hurst exponent obtained from the time series of computer usage profiles vs. their surrogate counterparts in the two assessed conditions: with and without background process activity.}

\begin{table*}[]
\centering
\caption{\changed{Statistical comparison for the sample entropy and Hurst exponent, obtained by leveraging the time series of computer usage profiles against their surrogate counterparts. \textit{Condition 1} denotes that background process activities were assessed, and \textit{Condition 2} indicates that these background activities were \textit{not} assessed. All users have been de-identified.}}
\label{tab:entropy-hurst}
\resizebox{\textwidth}{!}{%
\begin{tabular}{@{}cccccccccccccccccccccc@{}}
\toprule
\textbf{} &
  \multicolumn{10}{c}{\textbf{Sample Entropy}} &
  \textbf{} &
  \multicolumn{10}{c}{\textbf{Hurst Exponent}} \\ \cmidrule(lr){2-11} \cmidrule(l){13-22} 
 &
  \multicolumn{5}{c}{\textbf{Condition 1}} &
  \multicolumn{5}{c}{\textbf{Condition 2}} &
   &
  \multicolumn{5}{c}{\textbf{Condition 1}} &
  \multicolumn{5}{c}{\textbf{Condition 2}} \\ \midrule
 &
  \multicolumn{2}{c}{\textbf{Time Series}} &
  \multicolumn{2}{c}{\textbf{Surrogates}} &
   &
  \multicolumn{2}{c}{\textbf{Time Series}} &
  \multicolumn{2}{c}{\textbf{Surrogates}} &
   &
  \textbf{} &
  \multicolumn{2}{c}{\textbf{Time Series}} &
  \multicolumn{2}{c}{\textbf{Surrogates}} &
   &
  \multicolumn{2}{c}{\textbf{Time Series}} &
  \multicolumn{2}{c}{\textbf{Surrogates}} &
   \\
\multirow{-2}{*}{\textbf{User}} &
  \textbf{$\mu$} &
  \textbf{$\sigma$} &
  \textbf{$\mu$} &
  \textbf{$\sigma$} &
  \multirow{-2}{*}{\textbf{p}} &
  \textbf{$\mu$} &
  \textbf{$\sigma$} &
  \textbf{$\mu$} &
  \textbf{$\sigma$} &
  \multirow{-2}{*}{\textbf{p}} &
  \textbf{} &
  \textbf{$\mu$} &
  \textbf{$\sigma$} &
  \textbf{$\mu$} &
  \textbf{$\sigma$} &
  \multirow{-2}{*}{\textbf{p}} &
  \textbf{$\mu$} &
  \textbf{$\sigma$} &
  \textbf{$\mu$} &
  \textbf{$\sigma$} &
  \multirow{-2}{*}{\textbf{p}} \\ \hdashline
1 &
  0.215 &
  0.000 &
  0.568 &
  0.026 &
  *** &
  0.259 &
  0.000 &
  0.467 &
  0.022 &
  *** &
   &
  0.583 &
  0.022 &
  0.500 &
  0.040 &
  *** &
  0.538 &
  0.020 &
  0.496 &
  0.035 &
  *** \\\hdashline
2 &
  0.146 &
  0.000 &
  0.295 &
  0.011 &
  *** &
  0.428 &
  0.000 &
  0.470 &
  0.019 &
  *** &
   &
  0.639 &
  0.008 &
  0.488 &
  0.039 &
  *** &
  0.546 &
  0.000 &
  0.489 &
  0.031 &
  *** \\ \hdashline
3 &
  0.263 &
  0.000 &
  0.951 &
  0.014 &
  *** &
  0.469 &
  0.000 &
  0.743 &
  0.029 &
  *** &
   &
  0.765 &
  0.013 &
  0.499 &
  0.035 &
  *** &
  0.644 &
  0.002 &
  0.498 &
  0.034 &
  *** \\ \hdashline
4 &
  0.214 &
  0.000 &
  0.901 &
  0.033 &
  *** &
  0.309 &
  0.000 &
  0.488 &
  0.021 &
  *** &
   &
  0.778 &
  0.000 &
  0.504 &
  0.038 &
  *** &
  0.642 &
  0.006 &
  0.487 &
  0.035 &
  *** \\ \hdashline
5 &
  0.539 &
  0.000 &
  1.237 &
  0.044 &
  *** &
  0.559 &
  0.000 &
  0.975 &
  0.040 &
  *** &
   &
  0.496 &
  0.016 &
  0.496 &
  0.037 &
 0.989 &
  0.517 &
  0.004 &
  0.493 &
  0.034 &
  *** \\ \hdashline
6 &
  0.126 &
  0.000 &
  0.370 &
  0.015 &
  *** &
  0.152 &
  0.000 &
  0.247 &
  0.010 &
  *** &
   &
  0.737 &
  0.000 &
  0.495 &
  0.039 &
  *** &
  0.630 &
  0.002 &
  0.487 &
  0.030 &
  *** \\ \hdashline
7 &
  0.228 &
  0.000 &
  0.934 &
  0.032 &
  *** &
  0.285 &
  0.000 &
  0.410 &
  0.016 &
  *** &
   &
  0.679 &
  0.011 &
  0.493 &
  0.041 &
  *** &
  0.509 &
  0.008 &
  0.486 &
  0.036 &
  *** \\ \hdashline
8 &
  0.092 &
  0.000 &
  0.159 &
  0.007 &
  *** &
  0.098 &
  0.000 &
  0.154 &
  0.005 &
  *** &
   &
  0.635 &
  0.018 &
  0.469 &
  0.023 &
  *** &
  0.649 &
  0.012 &
  0.477 &
  0.030 &
  *** \\ \hdashline
9 &
  0.258 &
  0.000 &
  0.899 &
  0.009 &
  *** &
  0.403 &
  0.000 &
  1.050 &
  0.043 &
  *** &
   &
  0.326 &
  0.045 &
  0.495 &
  0.040 &
  *** &
  0.348 &
  0.030 &
  0.502 &
  0.037 &
  *** \\ \hdashline
10 &
  0.234 &
  0.000 &
  0.648 &
  0.027 &
  *** &
  0.294 &
  0.000 &
  0.444 &
  0.020 &
  *** &
   &
  0.591 &
  0.033 &
  0.498 &
  0.042 &
  *** &
  0.383 &
  0.026 &
  0.496 &
  0.037 &
  *** \\ \hdashline
11 &
  0.181 &
  0.000 &
  0.646 &
  0.027 &
  *** &
  0.205 &
  0.000 &
  0.415 &
  0.017 &
  *** &
   &
  0.578 &
  0.019 &
  0.493 &
  0.034 &
  *** &
  0.424 &
  0.007 &
  0.485 &
  0.033 &
  *** \\ \hdashline
12 &
  0.087 &
  0.000 &
  0.432 &
  0.016 &
  *** &
  0.100 &
  0.000 &
  0.279 &
  0.011 &
  *** &
   &
  0.552 &
  0.014 &
  0.495 &
  0.036 &
  *** &
  0.628 &
  0.013 &
  0.482 &
  0.032 &
  *** \\ \hdashline
13 &
  0.378 &
  0.000 &
  0.991 &
  0.044 &
  *** &
  0.494 &
  0.000 &
  0.846 &
  0.038 &
  *** &
   &
  0.617 &
  0.049 &
  0.503 &
  0.037 &
  *** &
  0.546 &
  0.019 &
  0.488 &
  0.034 &
  *** \\ \hdashline
14 &
  0.226 &
  0.000 &
  0.668 &
  0.027 &
  *** &
  0.329 &
  0.000 &
  0.552 &
  0.025 &
  *** &
   &
  0.548 &
  0.039 &
  0.500 &
  0.036 &
  *** &
  0.543 &
  0.051 &
  0.489 &
  0.032 &
  *** \\ \hdashline
15 &
  0.173 &
  0.000 &
  0.552 &
  0.023 &
  *** &
  0.226 &
  0.000 &
  0.496 &
  0.019 &
  *** &
   &
  0.646 &
  0.023 &
  0.502 &
  0.037 &
  *** &
  0.585 &
  0.028 &
  0.492 &
  0.036 &
  *** \\ \hdashline
16 &
  0.147 &
  0.000 &
  0.784 &
  0.040 &
  *** &
  0.218 &
  0.000 &
  0.464 &
  0.016 &
  *** &
   &
  0.462 &
  0.003 &
  0.495 &
  0.037 &
  *** &
  0.628 &
  0.001 &
  0.483 &
  0.035 &
  *** \\ \hdashline
17 &
  0.065 &
  0.000 &
  0.281 &
  0.012 &
  *** &
  0.085 &
  0.000 &
  0.198 &
  0.007 &
  *** &
   &
  0.682 &
  0.002 &
  0.494 &
  0.036 &
  *** &
  0.675 &
  0.001 &
  0.484 &
  0.036 &
  *** \\ \hdashline
18 &
  0.143 &
  0.000 &
  0.216 &
  0.009 &
  *** &
  0.177 &
  0.000 &
  0.212 &
  0.009 &
  *** &
   &
  0.565 &
  0.052 &
  0.484 &
  0.029 &
  *** &
  0.437 &
  0.017 &
  0.476 &
  0.030 &
  *** \\ \hdashline
19 &
  0.162 &
  0.000 &
  0.951 &
  0.008 &
  *** &
  0.321 &
  0.000 &
  0.443 &
  0.019 &
  *** &
   &
  0.791 &
  0.000 &
  0.491 &
  0.040 &
  *** &
  0.626 &
  0.007 &
  0.484 &
  0.033 &
  *** \\ \hdashline
20 &
  0.230 &
  0.000 &
  0.807 &
  0.029 &
  *** &
  0.265 &
  0.000 &
  0.579 &
  0.026 &
  *** &
   &
  0.686 &
  0.012 &
  0.493 &
  0.038 &
  *** &
  0.535 &
  0.033 &
  0.488 &
  0.033 &
  *** \\ \hdashline
21 &
  0.123 &
  0.000 &
  0.283 &
  0.012 &
  *** &
  0.097 &
  0.000 &
  0.153 &
  0.006 &
  *** &
   &
  0.564 &
  0.003 &
  0.488 &
  0.034 &
  *** &
  0.566 &
  0.006 &
  0.477 &
  0.029 &
  *** \\ \hdashline
22 &
  0.113 &
  0.000 &
  0.610 &
  0.026 &
  *** &
  0.156 &
  0.000 &
  0.388 &
  0.017 &
  *** &
   &
  0.485 &
  0.004 &
  0.496 &
  0.034 &
  0.0033 &
  0.627 &
  0.005 &
  0.493 &
  0.031 &
  *** \\ \hdashline
23 &
  0.175 &
  0.000 &
  0.621 &
  0.026 &
  *** &
  0.256 &
  0.000 &
  0.495 &
  0.022 &
  *** &
   &
  0.538 &
  0.073 &
  0.494 &
  0.037 &
  *** &
  0.558 &
  0.010 &
  0.500 &
  0.040 &
  *** \\ \hdashline
24 &
  0.392 &
  0.000 &
  1.444 &
  0.026 &
  *** &
  0.329 &
  0.000 &
  0.424 &
  0.017 &
  *** &
   &
  0.657 &
  0.041 &
  0.487 &
  0.043 &
  *** &
  0.568 &
  0.011 &
  0.490 &
  0.035 &
  *** \\ \hdashline
25 &
  0.228 &
  0.000 &
  0.512 &
  0.023 &
  *** &
  0.223 &
  0.000 &
  0.415 &
  0.018 &
  *** &
   &
  0.669 &
  0.098 &
  0.497 &
  0.035 &
  *** &
  0.557 &
  0.031 &
  0.493 &
  0.035 &
  *** \\ \hdashline
26 &
  0.320 &
  0.000 &
  0.832 &
  0.033 &
  *** &
  0.337 &
  0.000 &
  0.671 &
  0.027 &
  *** &
   &
  0.635 &
  0.006 &
  0.488 &
  0.035 &
  *** &
  0.526 &
  0.011 &
  0.488 &
  0.035 &
  *** \\ \hdashline
27 &
  0.338 &
  0.000 &
  1.092 &
  0.040 &
  *** &
  0.417 &
  0.000 &
  0.674 &
  0.027 &
  *** &
   &
  0.439 &
  0.042 &
  0.495 &
  0.039 &
  *** &
  0.551 &
  0.049 &
  0.488 &
  0.034 &
  *** \\ \hdashline
28 &
  0.471 &
  0.000 &
  1.382 &
  0.041 &
  *** &
  0.372 &
  0.000 &
  0.869 &
  0.031 &
  *** &
   &
  0.568 &
  0.025 &
  0.496 &
  0.041 &
  *** &
  0.716 &
  0.000 &
  0.499 &
  0.035 &
  *** \\ \hdashline
29 &
  0.286 &
  0.000 &
  0.914 &
  0.039 &
  *** &
  0.311 &
  0.000 &
  0.611 &
  0.025 &
  *** &
   &
  0.593 &
  0.008 &
  0.493 &
  0.037 &
  *** &
  0.530 &
  0.026 &
  0.494 &
  0.034 &
  *** \\ \hdashline
30 &
  0.060 &
  0.000 &
  0.383 &
  0.017 &
  *** &
  0.108 &
  0.000 &
  0.192 &
  0.008 &
  *** &
   &
  0.733 &
  0.026 &
  0.503 &
  0.036 &
  *** &
  0.646 &
  0.005 &
  0.481 &
  0.036 &
  *** \\ \hdashline
31 &
  0.271 &
  0.000 &
  0.784 &
  0.034 &
  *** &
  0.304 &
  0.000 &
  0.654 &
  0.032 &
  *** &
   &
  0.625 &
  0.006 &
  0.493 &
  0.042 &
  *** &
  0.599 &
  0.009 &
  0.490 &
  0.039 &
  *** \\ \hdashline
\textbf{Average} &
  \textbf{0.222} &
  \textbf{0.000} &
  \textbf{0.714} &
  \textbf{0.025} &
  \textbf{} &
  \textbf{0.277} &
  \textbf{0.000} &
  \textbf{0.499} &
  \textbf{0.021} &
  \textbf{} &
  \textbf{} &
  \textbf{0.608} &
  \textbf{0.023} &
  \textbf{0.494} &
  \textbf{0.037} &
  \textbf{} &
  \textbf{0.564} &
  \textbf{0.014} &
  \textbf{0.489} &
  \textbf{0.034} &
  \textbf{} \\ \bottomrule
  \multicolumn{5}{l}{\textasteriskcentered{}\textasteriskcentered{}\textasteriskcentered{} = p-value $<.001$} &
  \multicolumn{1}{l}{} &
  \multicolumn{1}{l}{} &
  \multicolumn{1}{l}{} &
  \multicolumn{1}{l}{} &
  \multicolumn{1}{l}{} &
  \multicolumn{1}{l}{} &
  \multicolumn{1}{l}{} &
  \multicolumn{1}{l}{} &
  \multicolumn{1}{l}{} &
  \multicolumn{1}{l}{} &
  \multicolumn{1}{l}{} &
  \multicolumn{1}{l}{} &
  \multicolumn{1}{l}{} &
  \multicolumn{1}{l}{} &
  \multicolumn{1}{l}{} &
  \multicolumn{1}{l}{} &
  \multicolumn{1}{l}{}
\end{tabular}%
}
\end{table*}

\section{Profiles' Temporal Consistency Results}
\label{appendix:temporal_consistency}

\changed{Figure~\ref{fig:periodogram_autocorr_all_users} exhibits the periodogram and autocorrelation results for all $\samplesize$ study participants in the two assessed conditions: with and without background process activity.}

\renewcommand{\thesubfigure}{\roman{subfigure}}

\begin{figure*}[htp]
\centering

\subfloat[User $1$, condition $1$]{%
  \includegraphics[clip,width=\columnwidth]{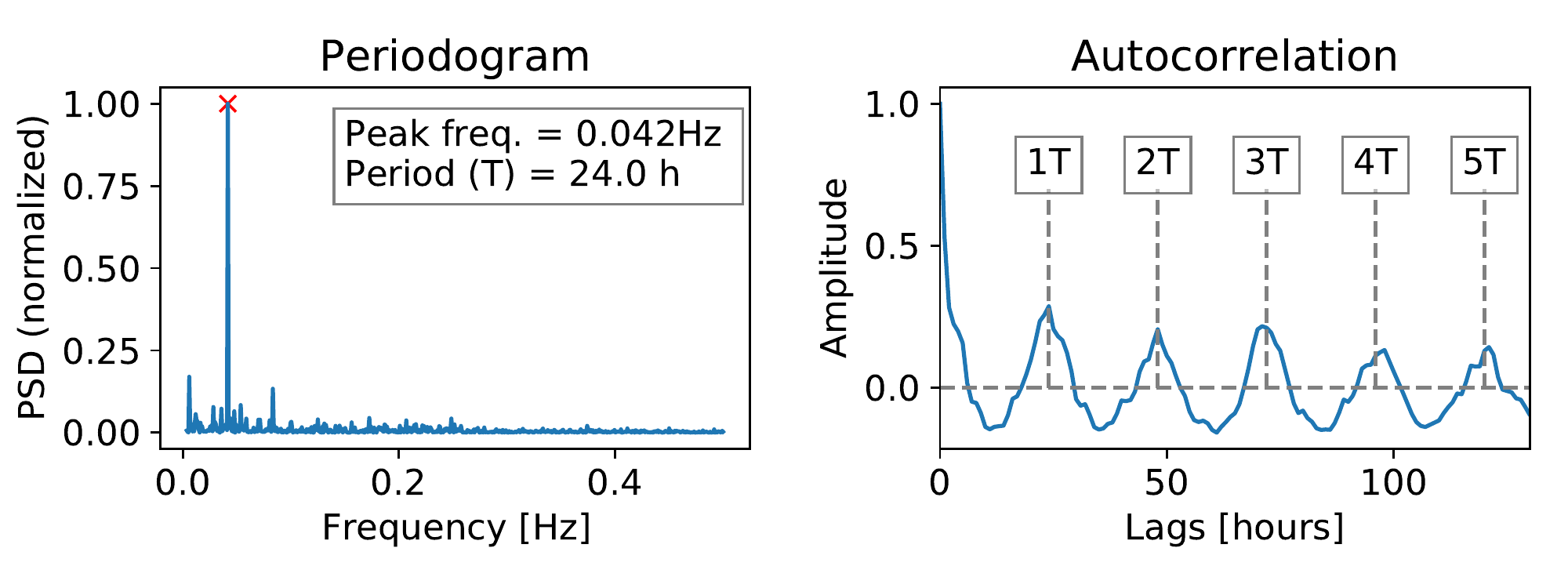}%
}
~
\subfloat[User $1$, condition $2$]{%
  \includegraphics[clip,width=\columnwidth]{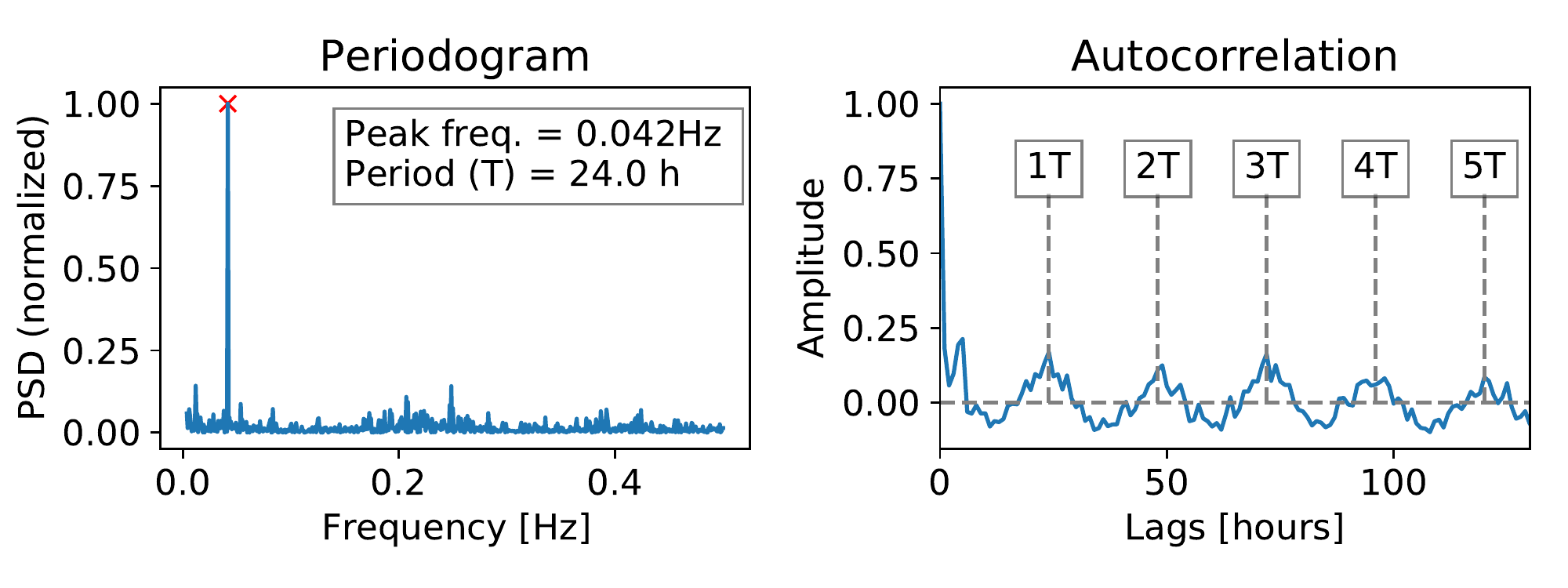}%
}

\subfloat[User $2$, condition $1$]{%
  \includegraphics[clip,width=\columnwidth]{figs/2021/Time series/periodogram_autocorr_user_2_condition_1.pdf}%
}
~
\subfloat[User $2$, condition $2$]{%
  \includegraphics[clip,width=\columnwidth]{figs/2021/Time series/periodogram_autocorr_user_2_condition_2.pdf}%
}

\subfloat[User $3$, condition $1$]{%
  \includegraphics[clip,width=\columnwidth]{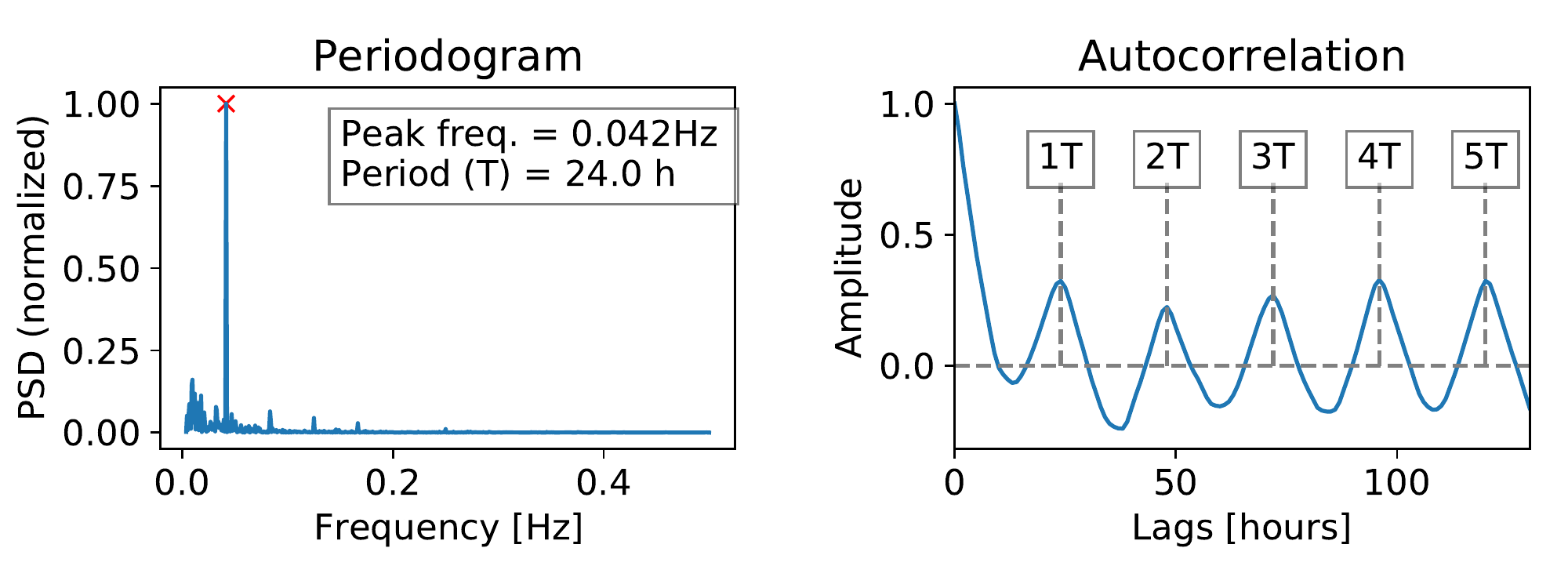}%
}
~
\subfloat[User $3$, condition $2$]{%
  \includegraphics[clip,width=\columnwidth]{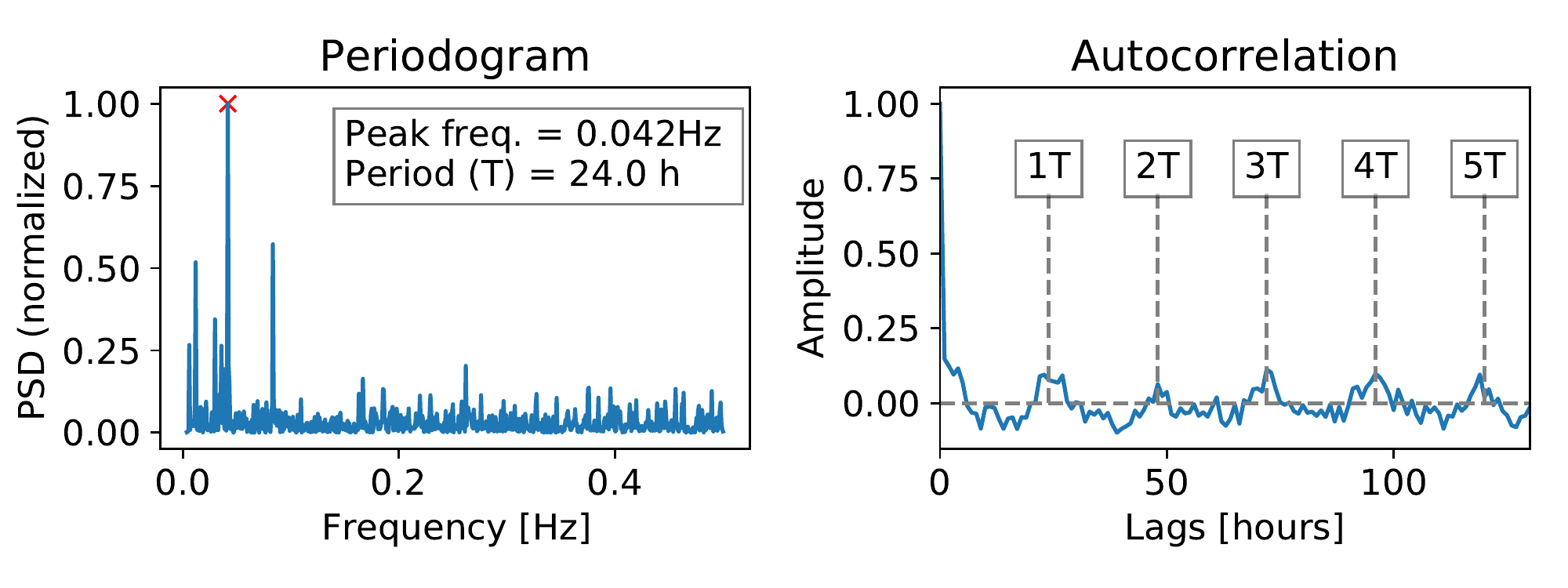}%
}

\subfloat[User $4$, condition $1$]{%
  \includegraphics[clip,width=\columnwidth]{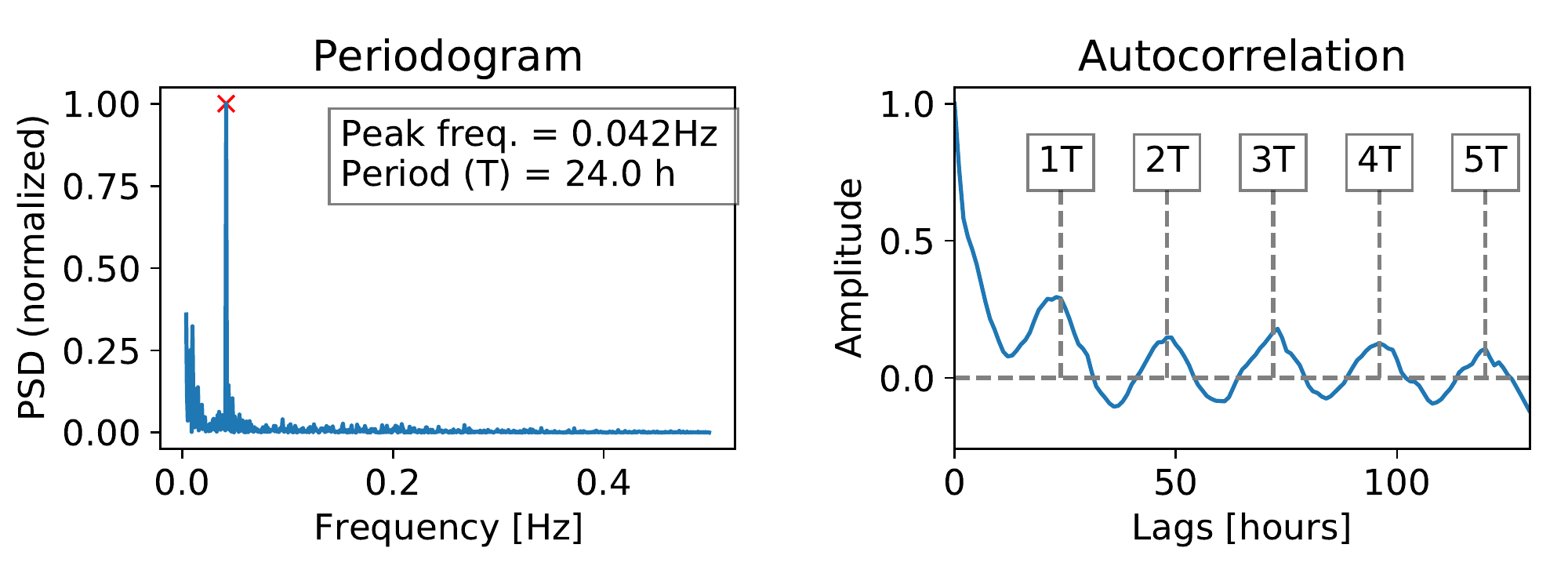}%
}
~
\subfloat[User $4$, condition $2$]{%
  \includegraphics[clip,width=\columnwidth]{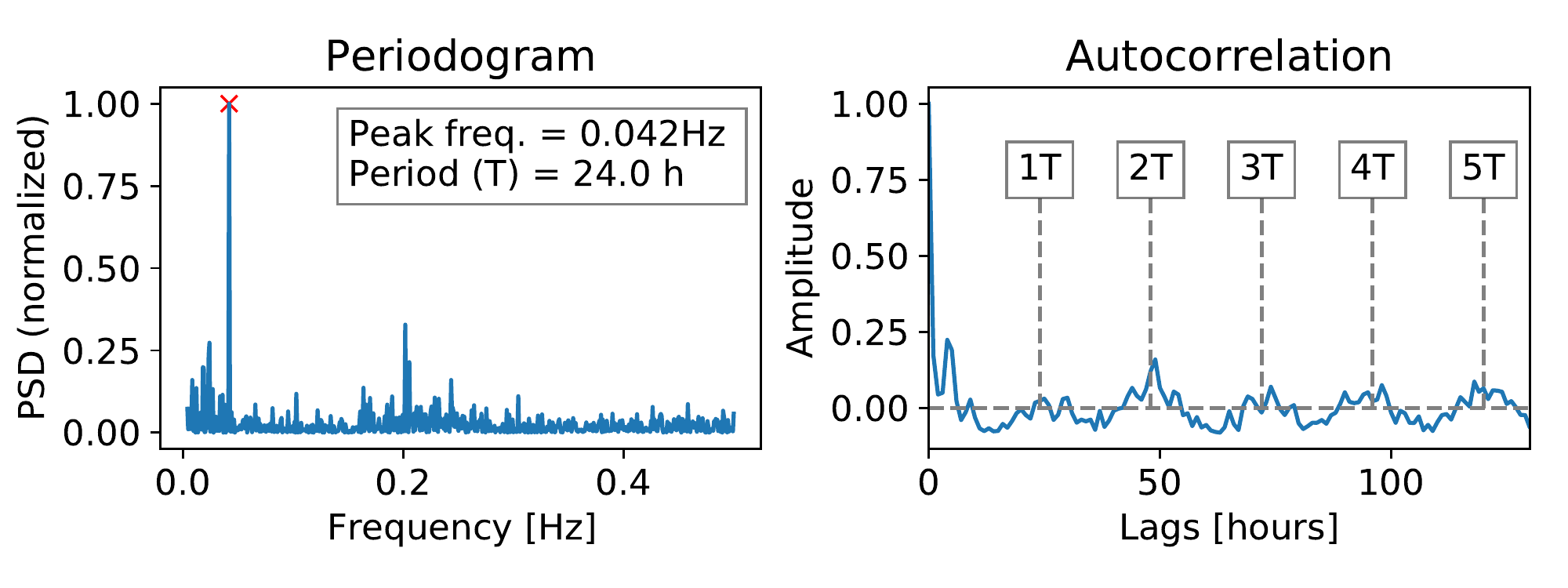}%
}

\subfloat[User $5$, condition $1$]{%
  \includegraphics[clip,width=\columnwidth]{figs/2021/Time series/periodogram_autocorr_user_5_condition_1.pdf}%
}
~
\subfloat[User $5$, condition $2$]{%
  \includegraphics[clip,width=\columnwidth]{figs/2021/Time series/periodogram_autocorr_user_5_condition_2.pdf}%
}

\subfloat[User $6$, condition $1$]{%
  \includegraphics[clip,width=\columnwidth]{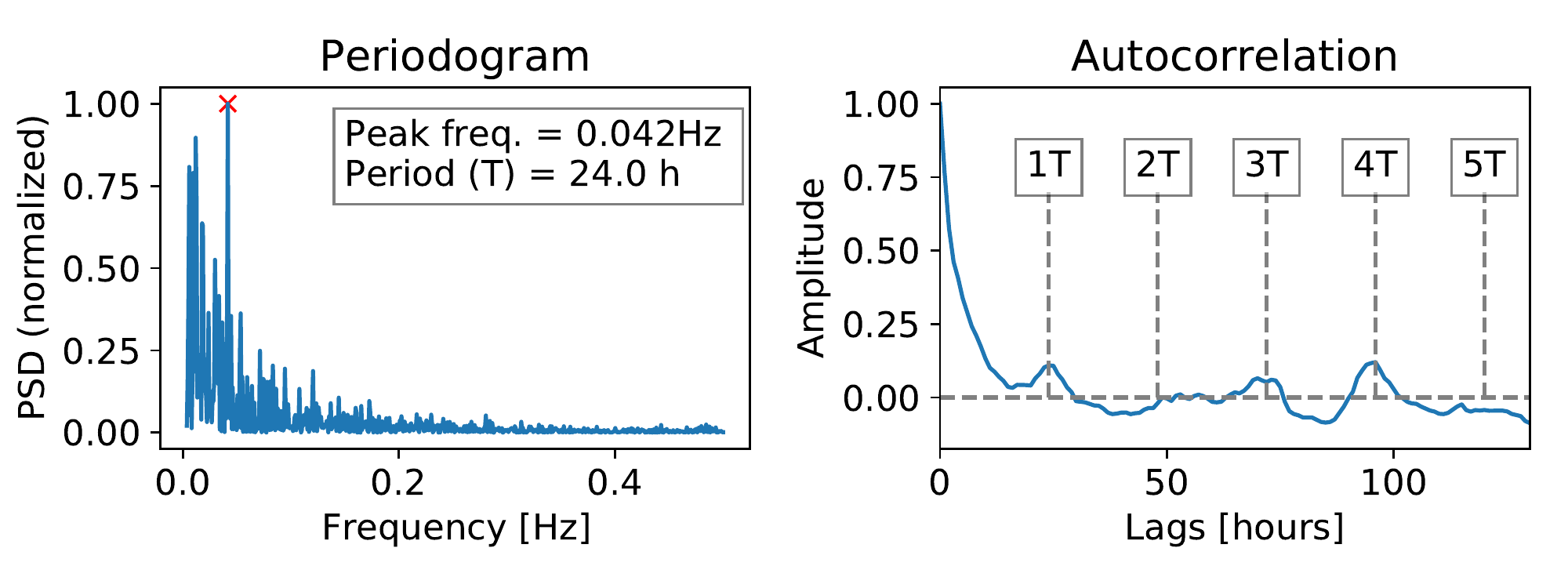}%
}
~
\subfloat[User $6$, condition $2$]{%
  \includegraphics[clip,width=\columnwidth]{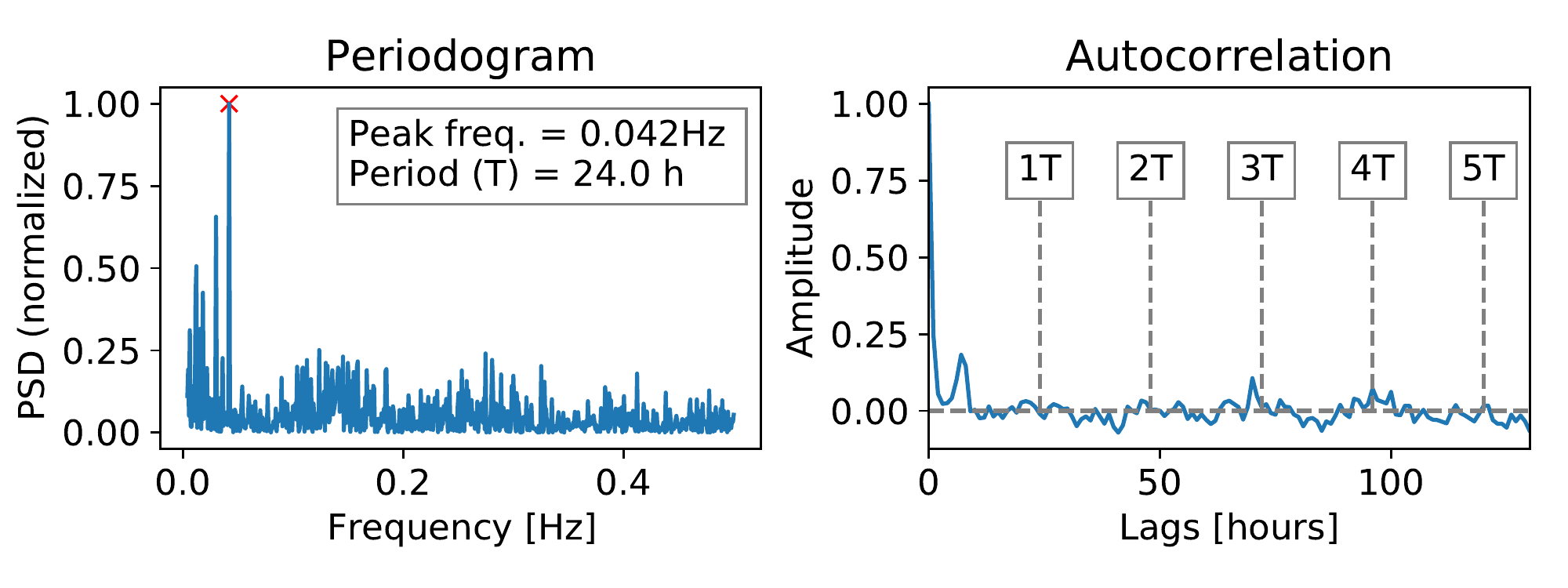}%
}

\end{figure*}

\begin{figure*}[htp]
\ContinuedFloat\centering

\subfloat[User $7$, condition $1$]{%
  \includegraphics[clip,width=\columnwidth]{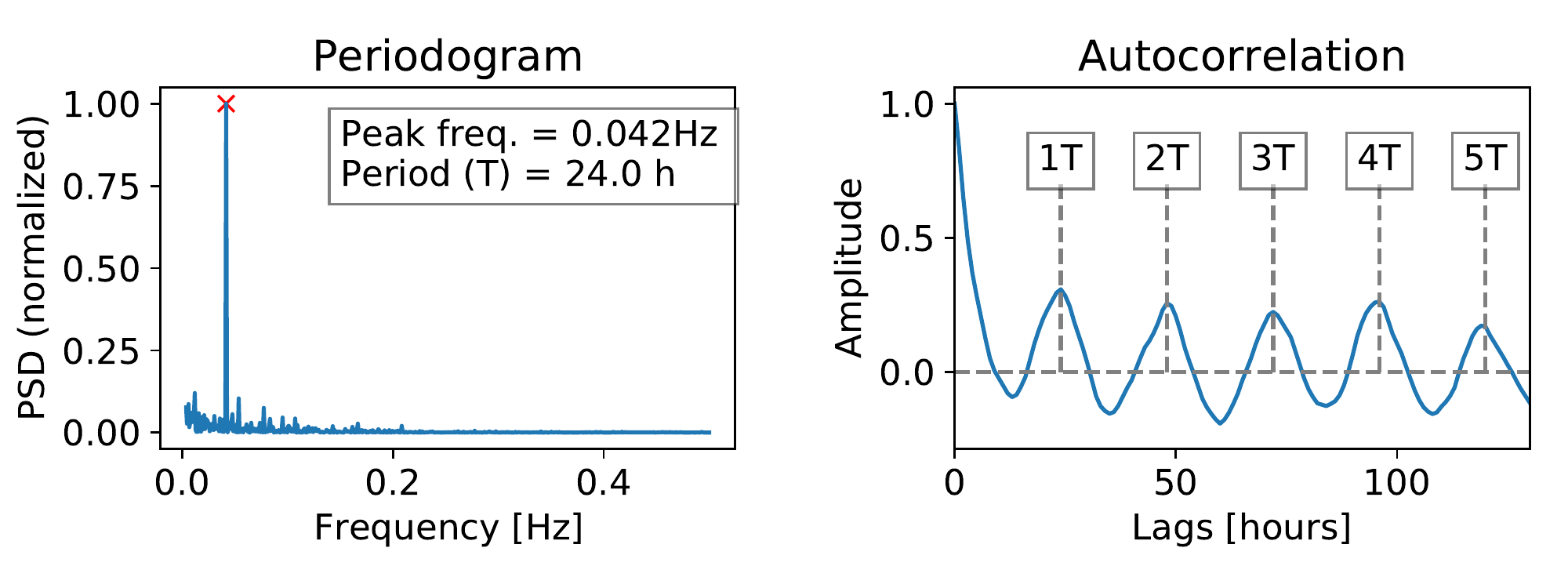}%
}
~
\subfloat[User $7$, condition $2$]{%
  \includegraphics[clip,width=\columnwidth]{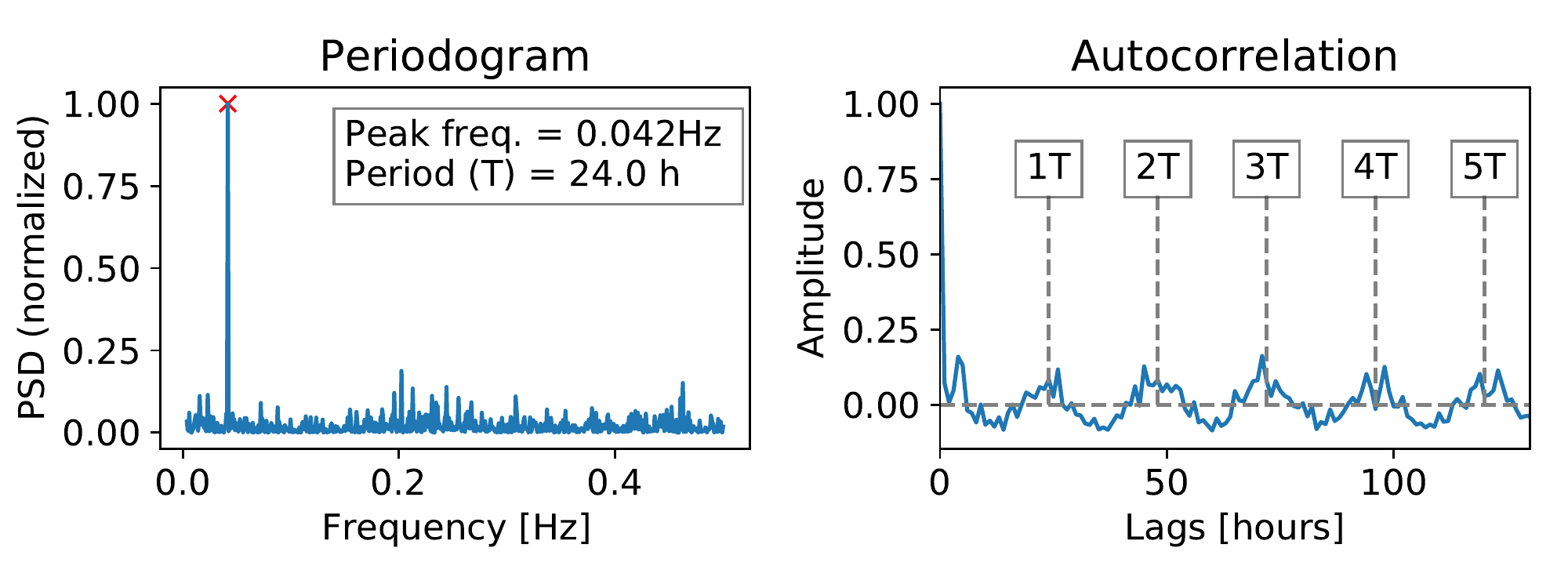}%
}

\subfloat[User $8$, condition $1$]{%
  \includegraphics[clip,width=\columnwidth]{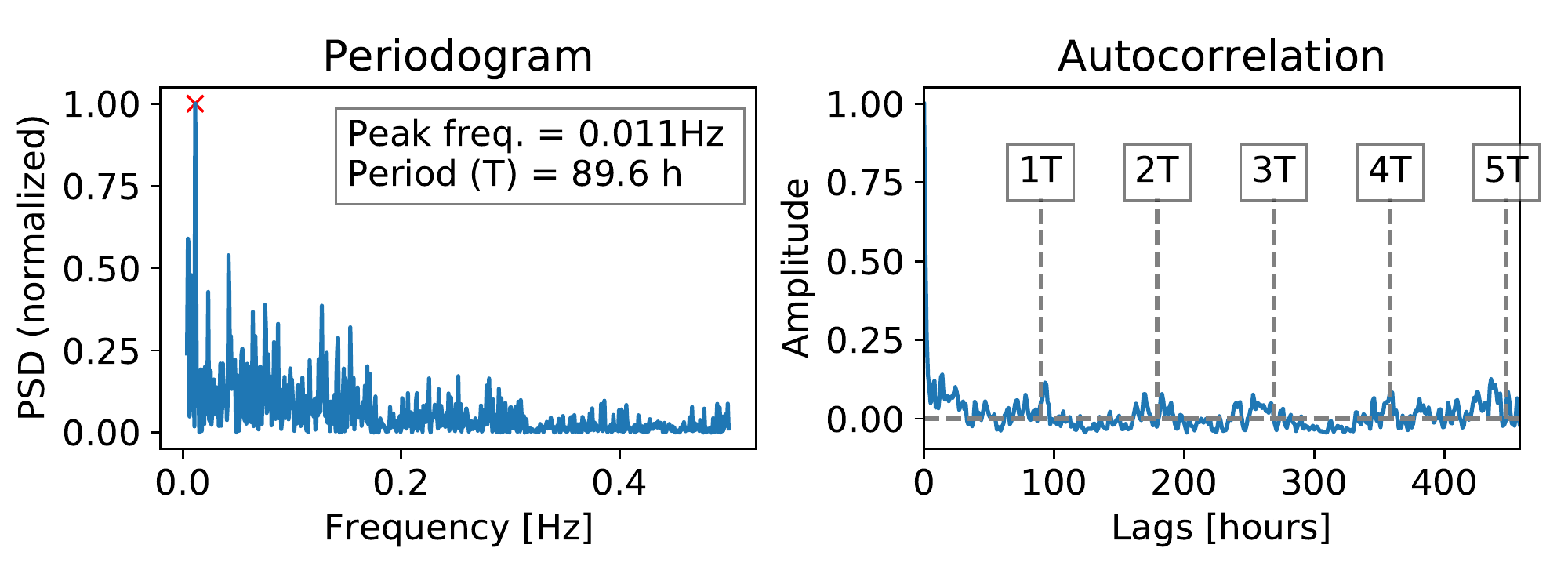}%
}
~
\subfloat[User $8$, condition $2$]{%
  \includegraphics[clip,width=\columnwidth]{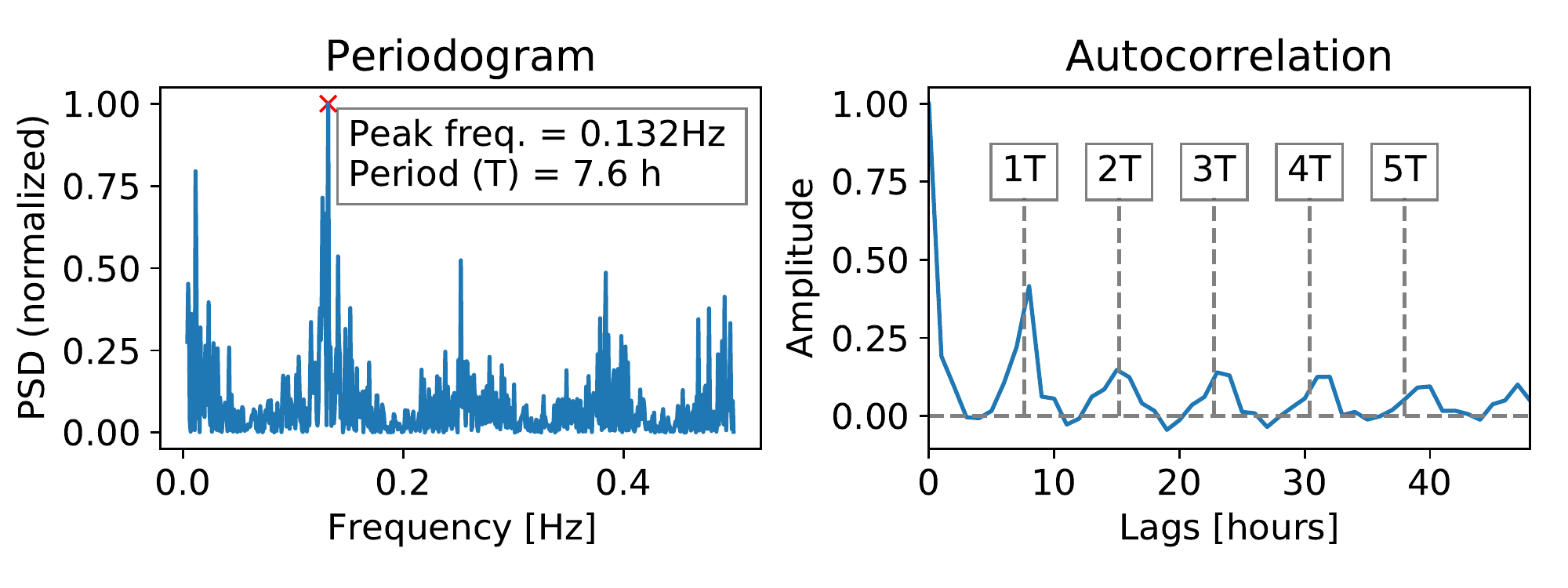}%
}

\subfloat[User $9$, condition $1$]{%
  \includegraphics[clip,width=\columnwidth]{figs/2021/Time series/periodogram_autocorr_user_9_condition_1.pdf}%
}
~
\subfloat[User $9$, condition $2$]{%
  \includegraphics[clip,width=\columnwidth]{figs/2021/Time series/periodogram_autocorr_user_9_condition_2.pdf}%
}

\subfloat[User $10$, condition $1$]{%
  \includegraphics[clip,width=\columnwidth]{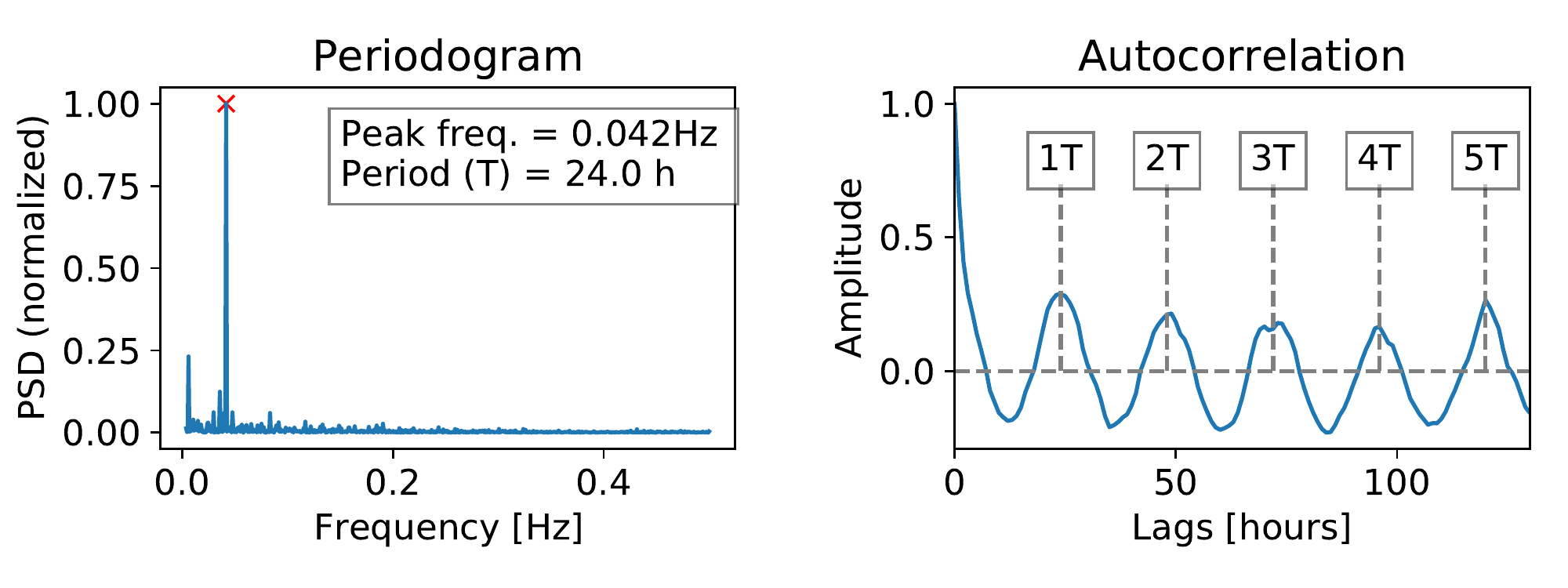}%
}
~
\subfloat[User $10$, condition $2$]{%
  \includegraphics[clip,width=\columnwidth]{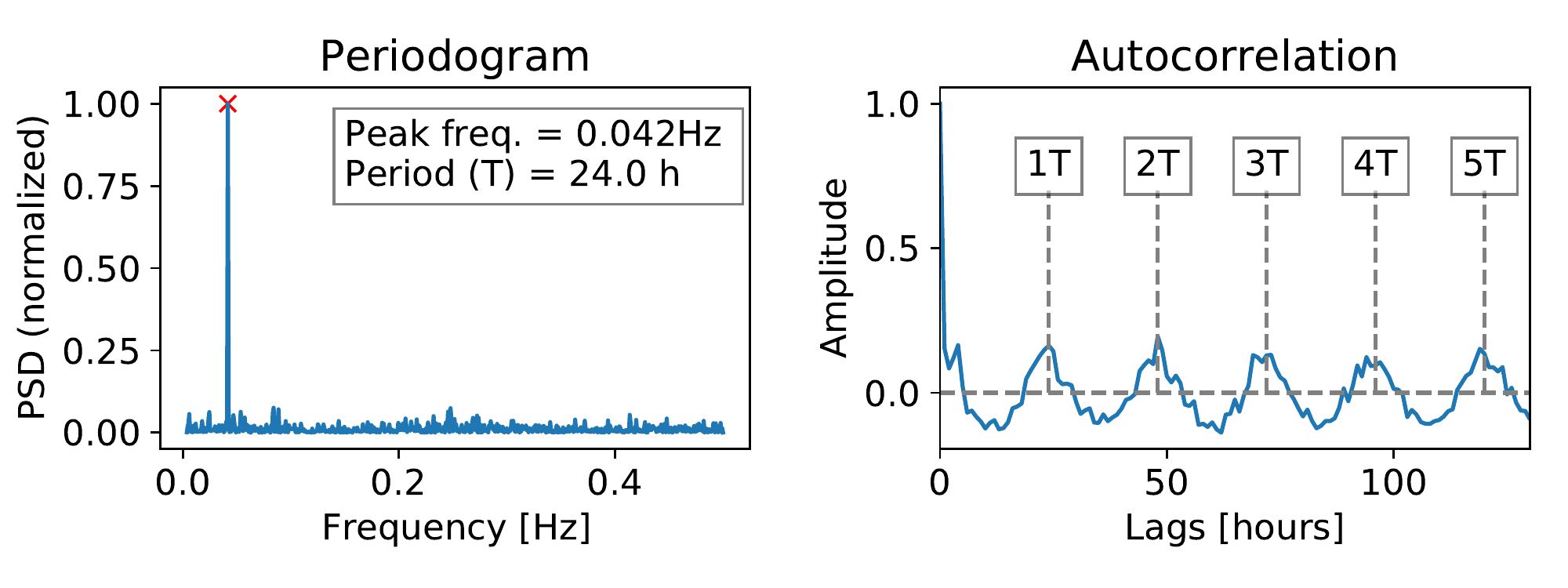}%
}

\subfloat[User $11$, condition $1$]{%
  \includegraphics[clip,width=\columnwidth]{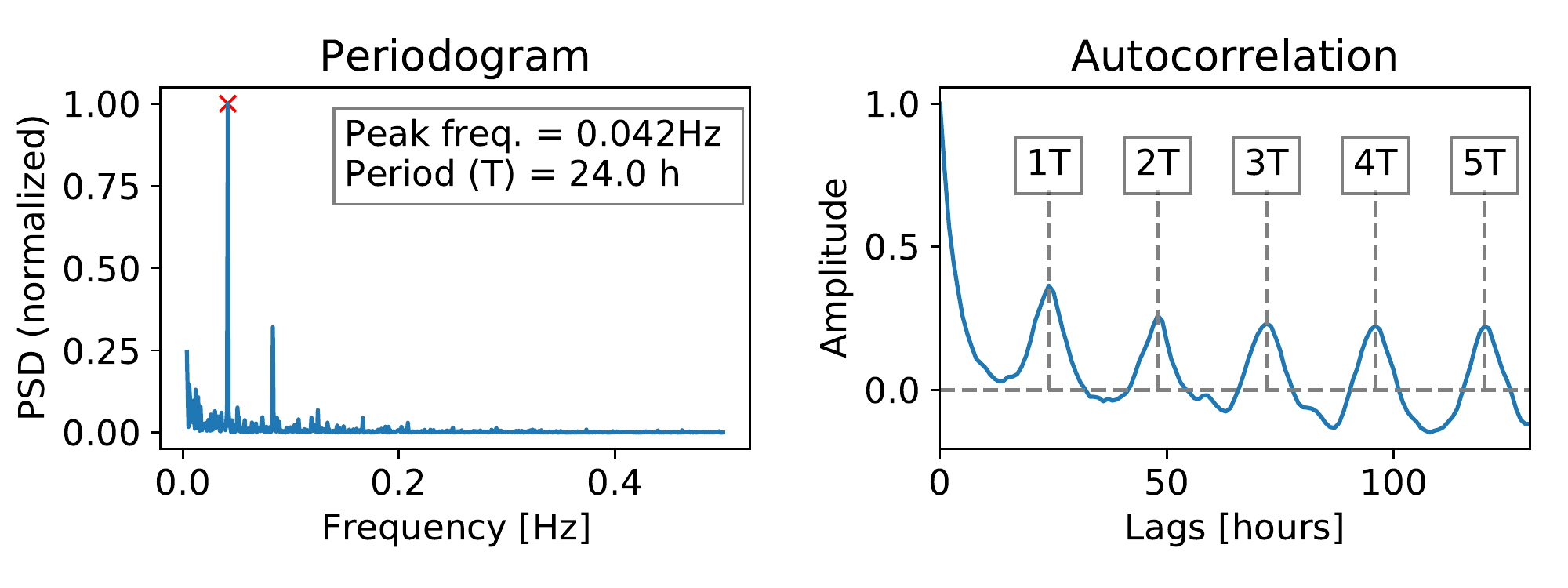}%
}
~
\subfloat[User $11$, condition $2$]{%
  \includegraphics[clip,width=\columnwidth]{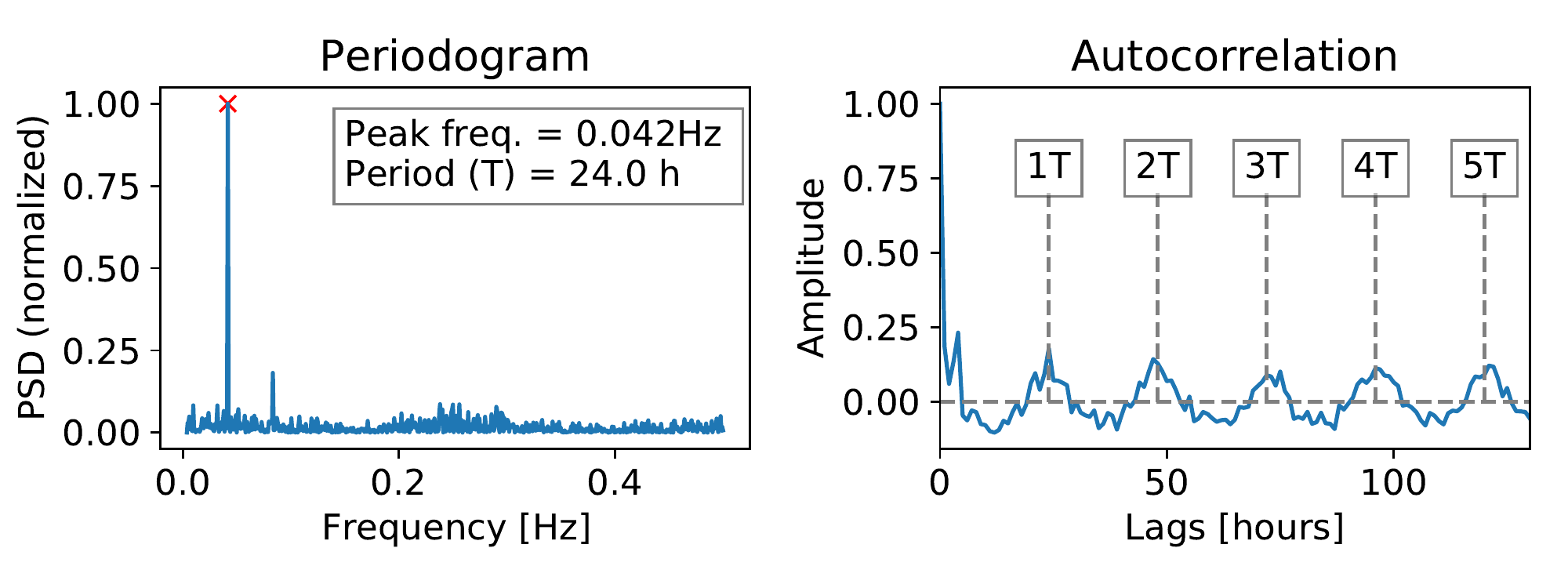}%
}

\subfloat[User $12$, condition $1$]{%
  \includegraphics[clip,width=\columnwidth]{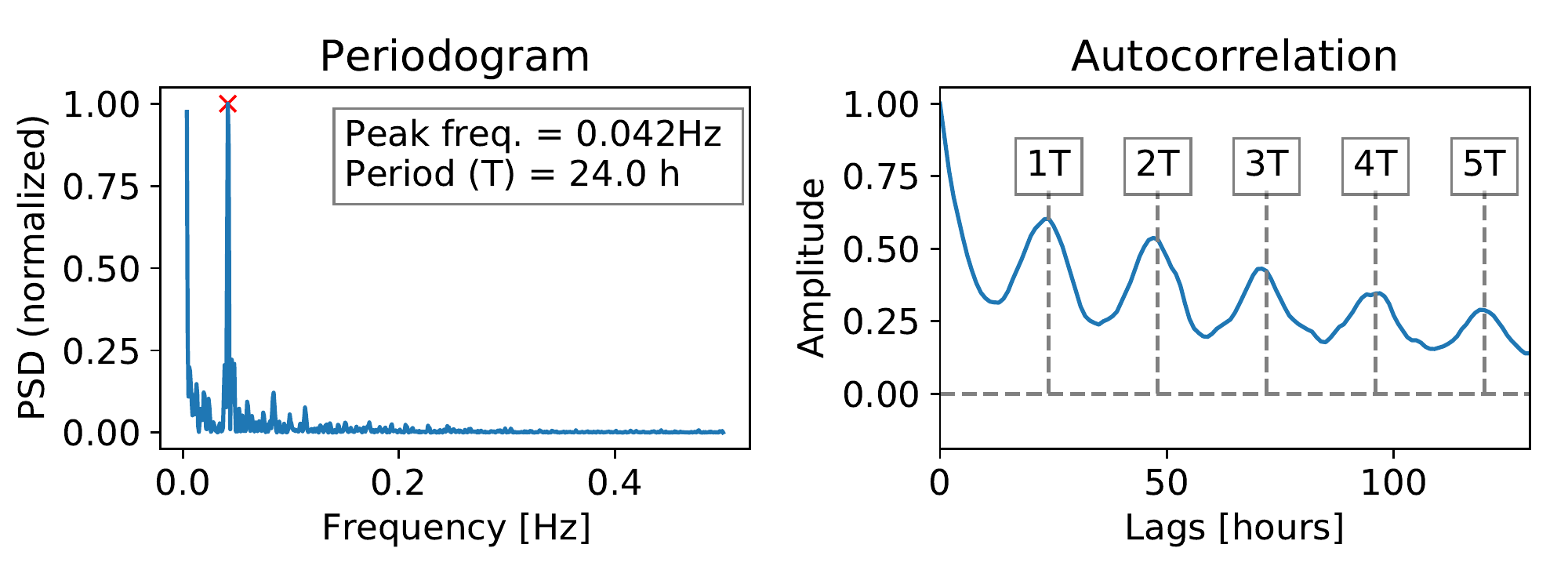}%
}
~
\subfloat[User $12$, condition $2$]{%
  \includegraphics[clip,width=\columnwidth]{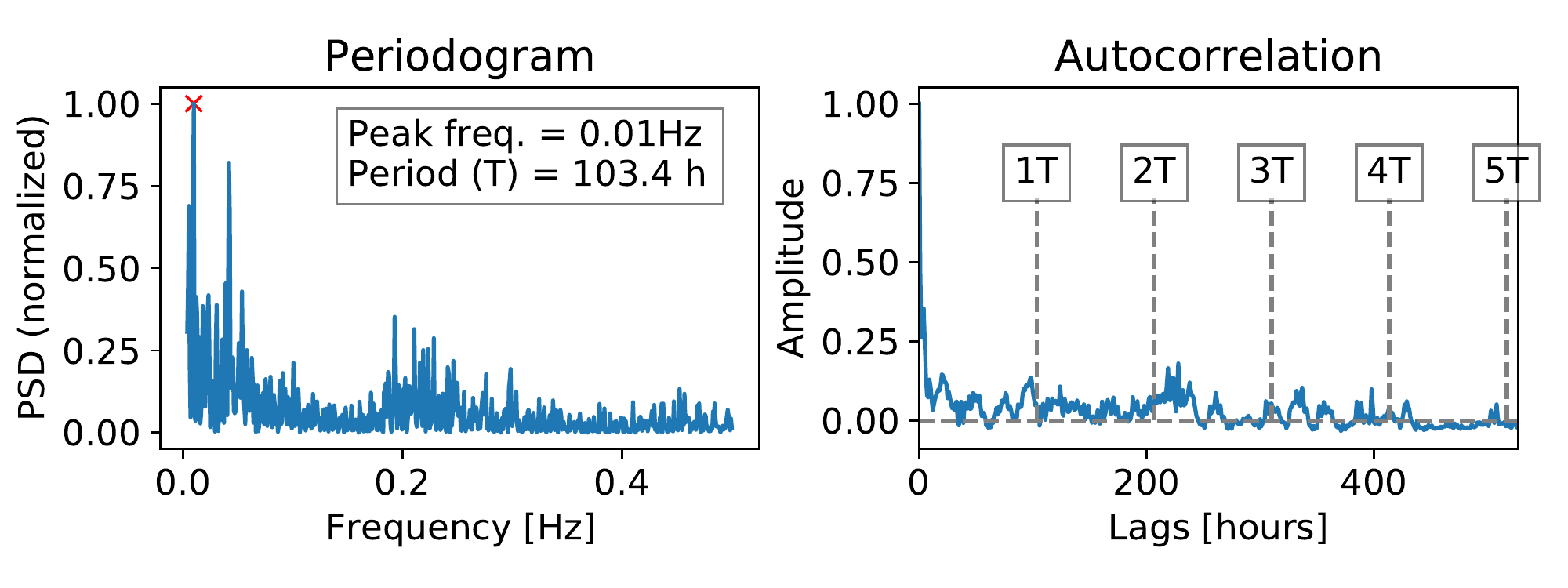}%
}

\end{figure*}

\begin{figure*}[htp]
\ContinuedFloat\centering

\subfloat[User $13$, condition $1$]{%
  \includegraphics[clip,width=\columnwidth]{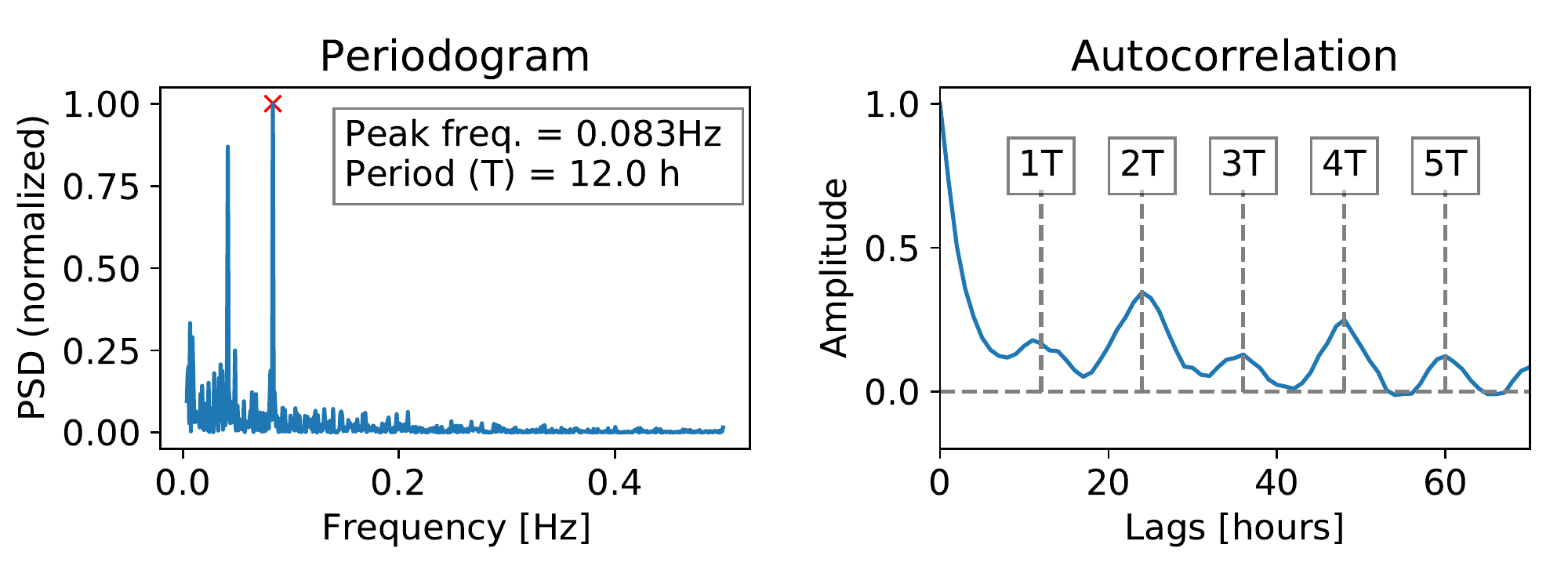}%
}
~
\subfloat[User $13$, condition $2$]{%
  \includegraphics[clip,width=\columnwidth]{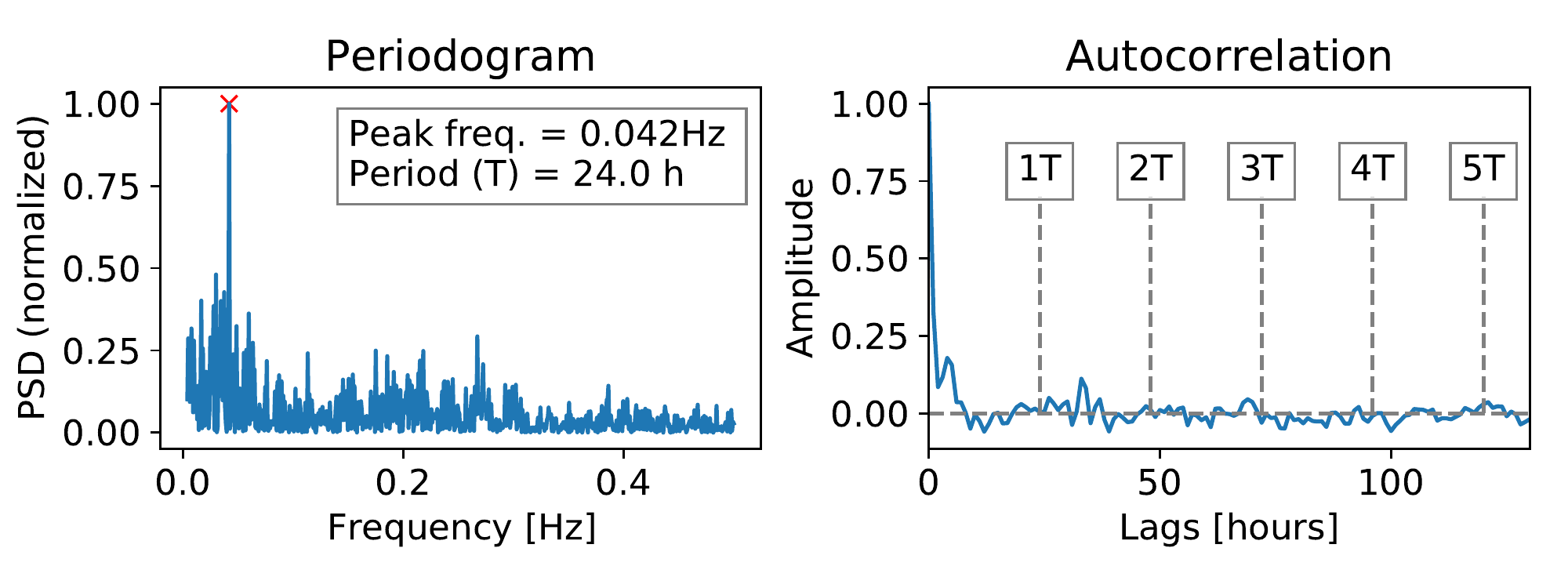}%
}

\subfloat[User $14$, condition $1$]{%
  \includegraphics[clip,width=\columnwidth]{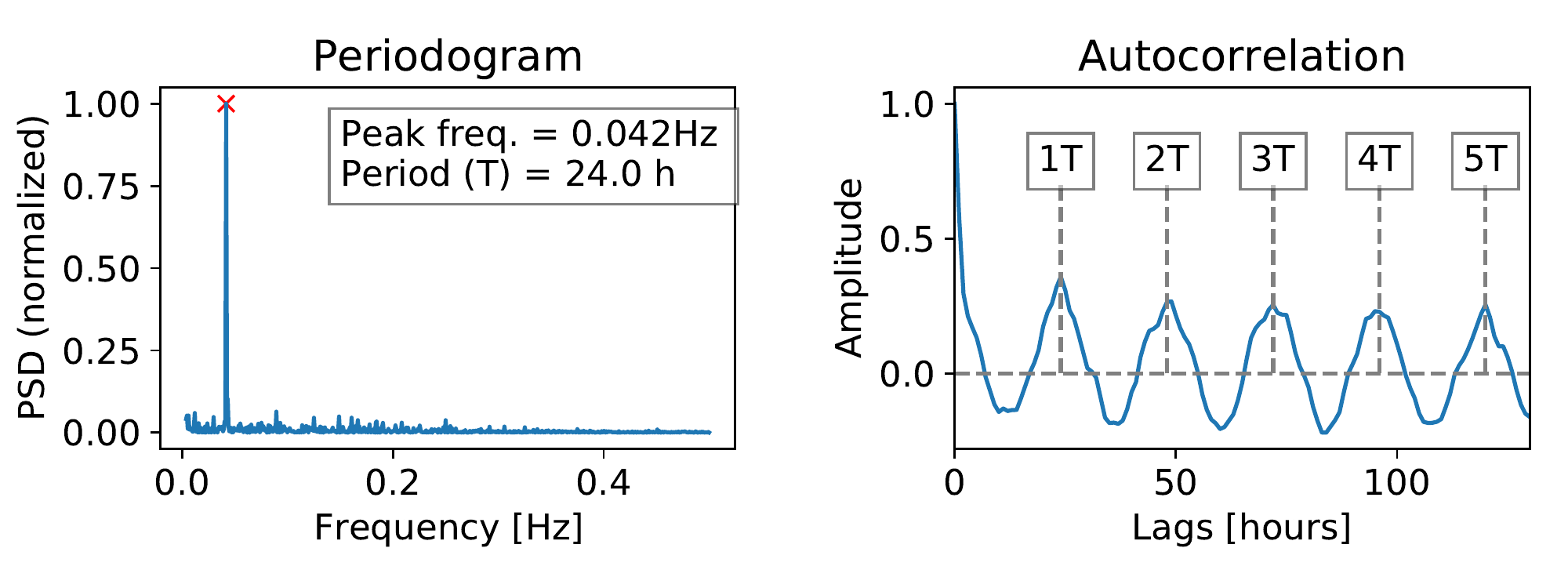}%
}
~
\subfloat[User $14$, condition $2$]{%
  \includegraphics[clip,width=\columnwidth]{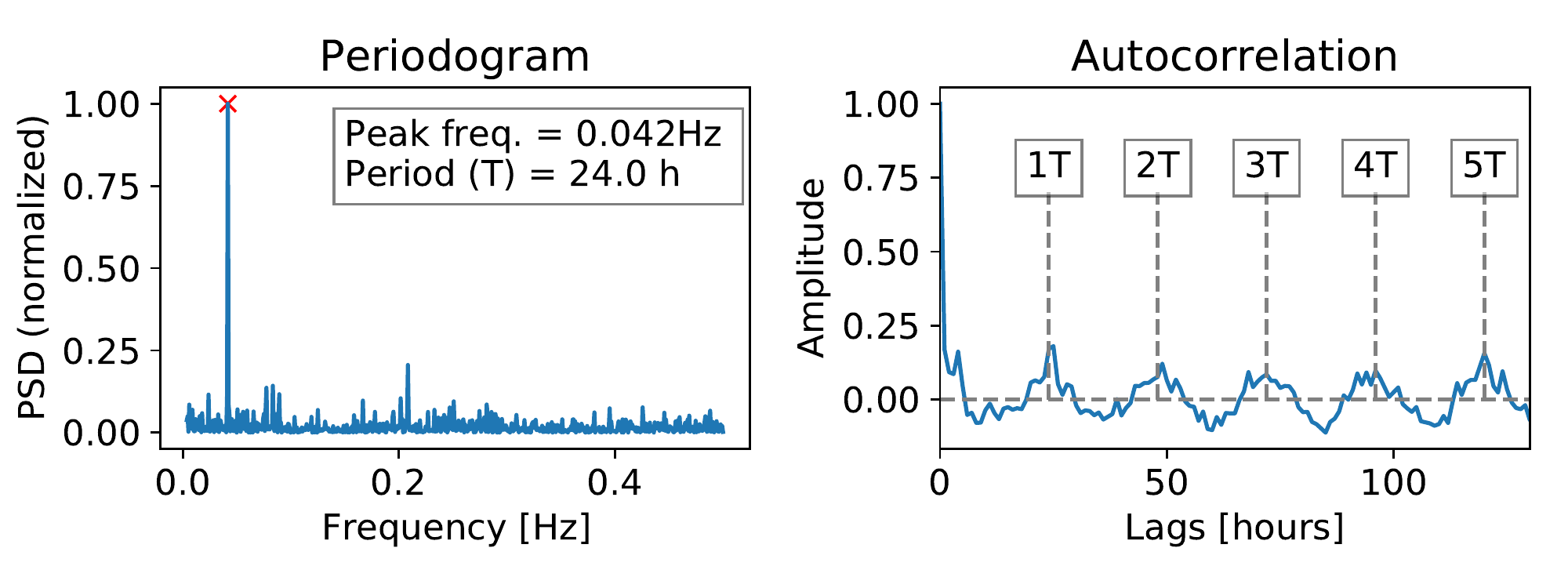}%
}

\subfloat[User $15$, condition $1$]{%
  \includegraphics[clip,width=\columnwidth]{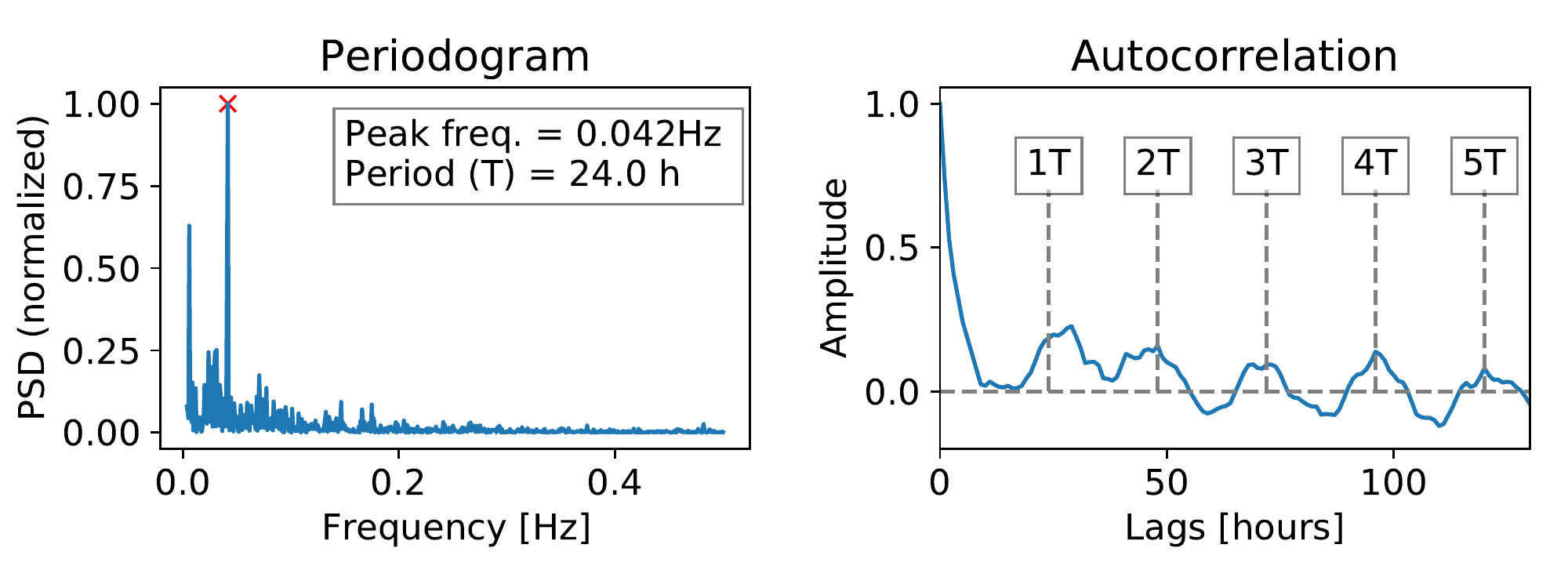}%
}
~
\subfloat[User $15$, condition $2$]{%
  \includegraphics[clip,width=\columnwidth]{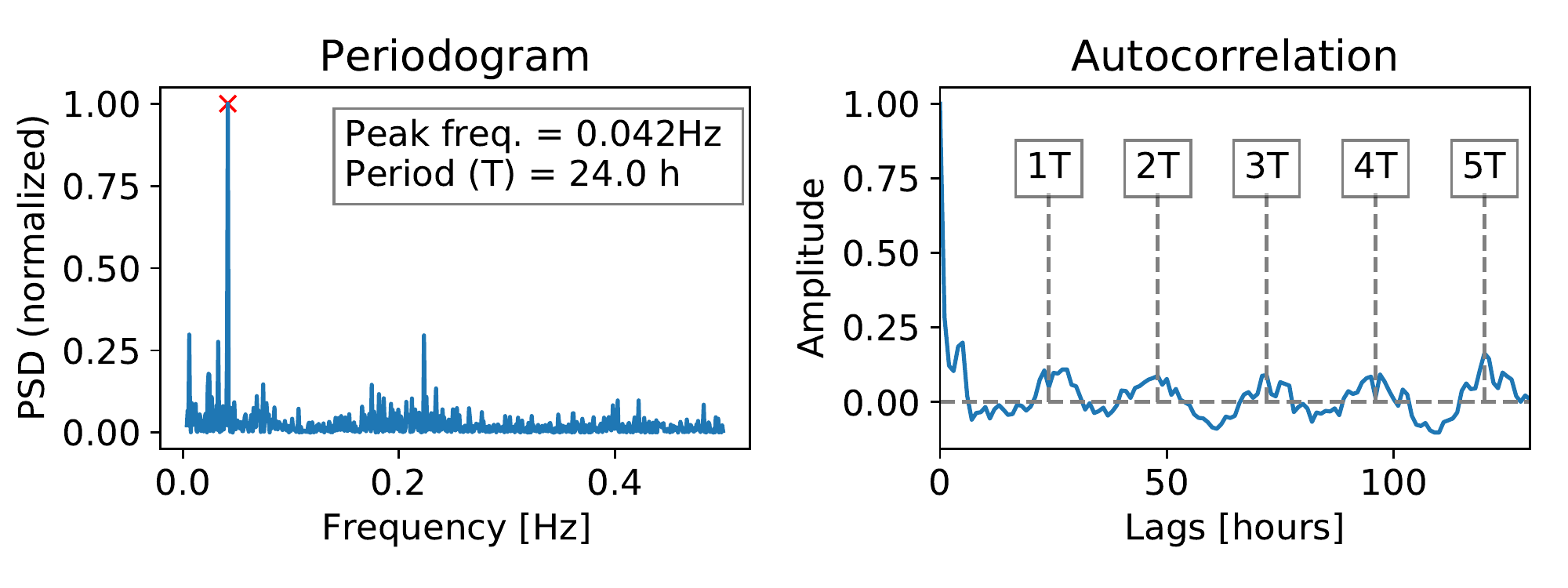}%
}

\subfloat[User $16$, condition $1$]{%
  \includegraphics[clip,width=\columnwidth]{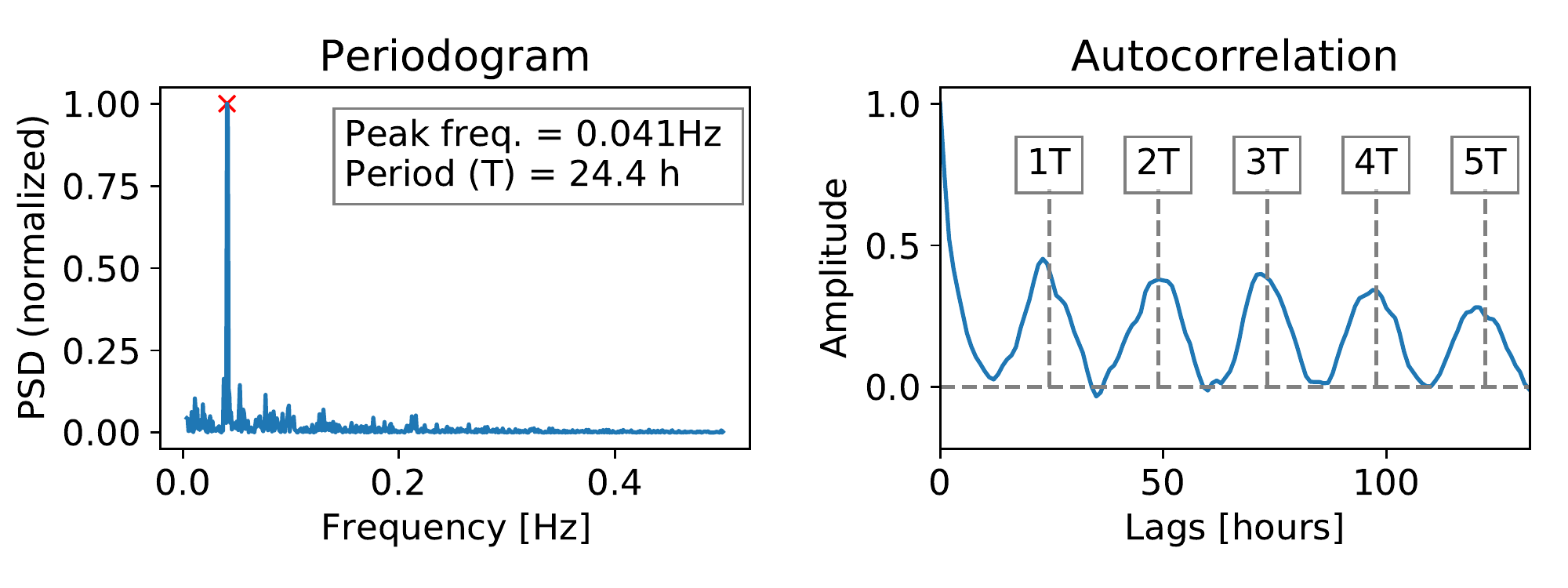}%
}
~
\subfloat[User $16$, condition $2$]{%
  \includegraphics[clip,width=\columnwidth]{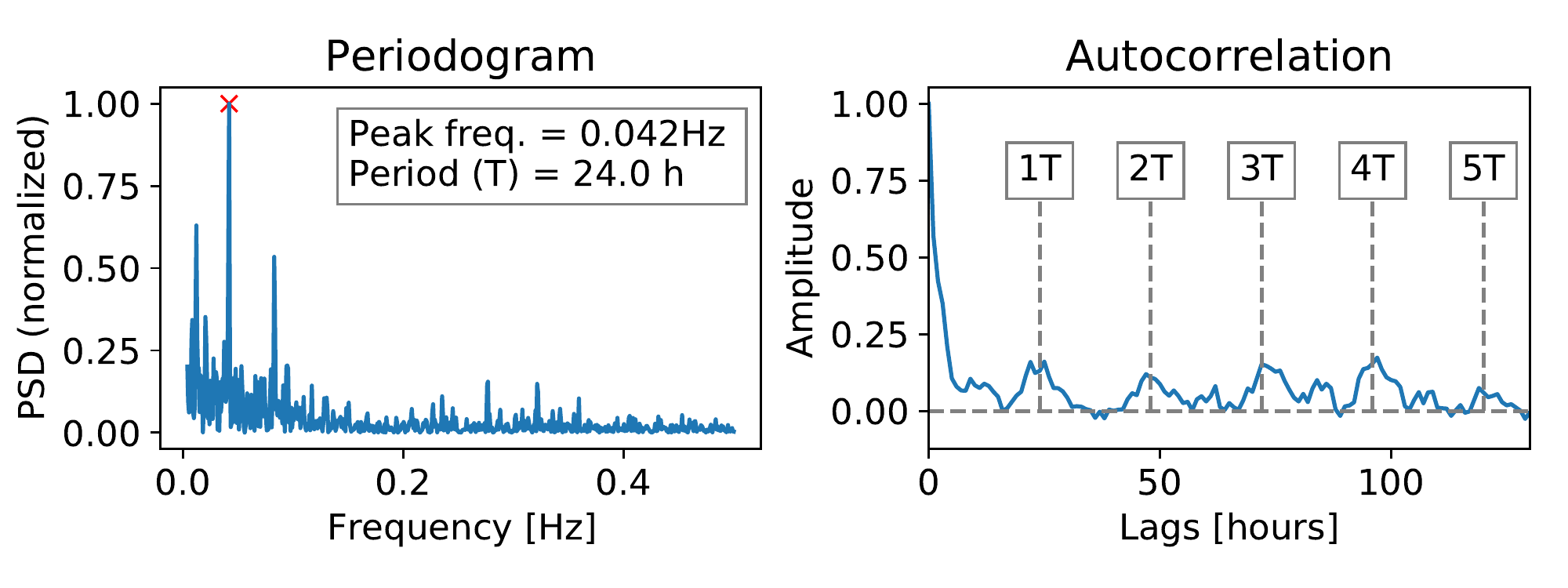}%
}

\subfloat[User $17$, condition $1$]{%
  \includegraphics[clip,width=\columnwidth]{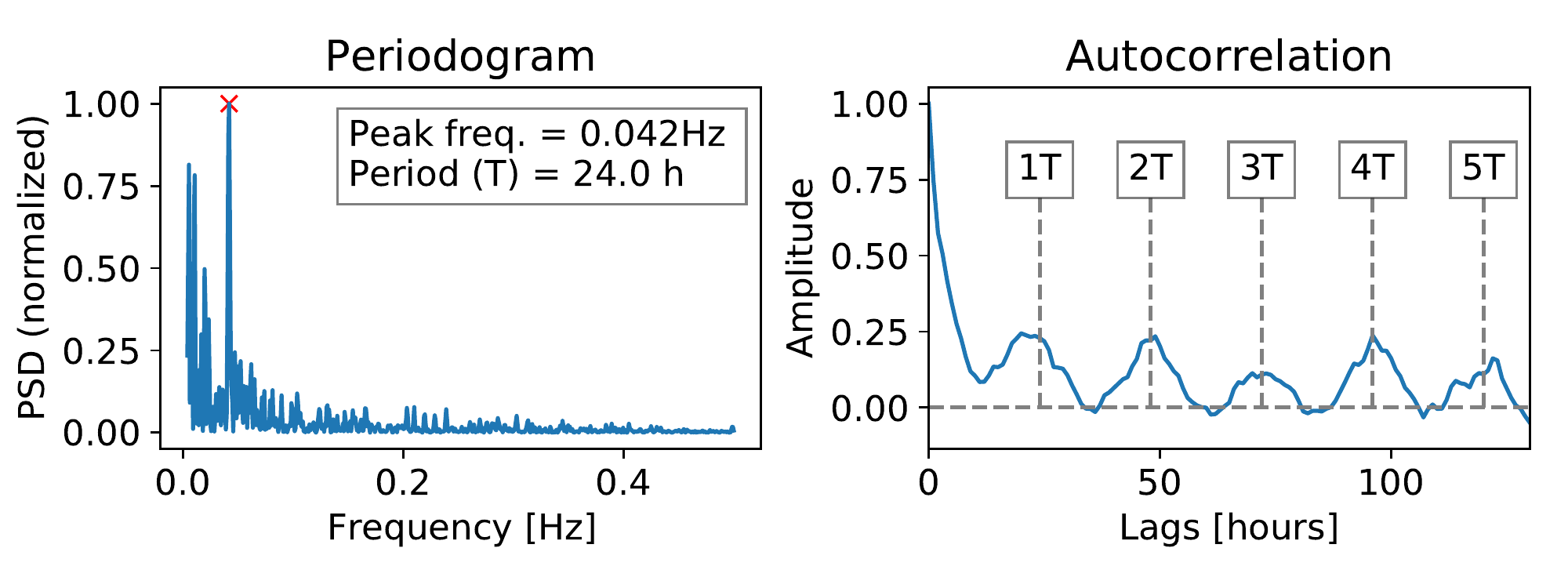}%
}
~
\subfloat[User $17$, condition $2$]{%
  \includegraphics[clip,width=\columnwidth]{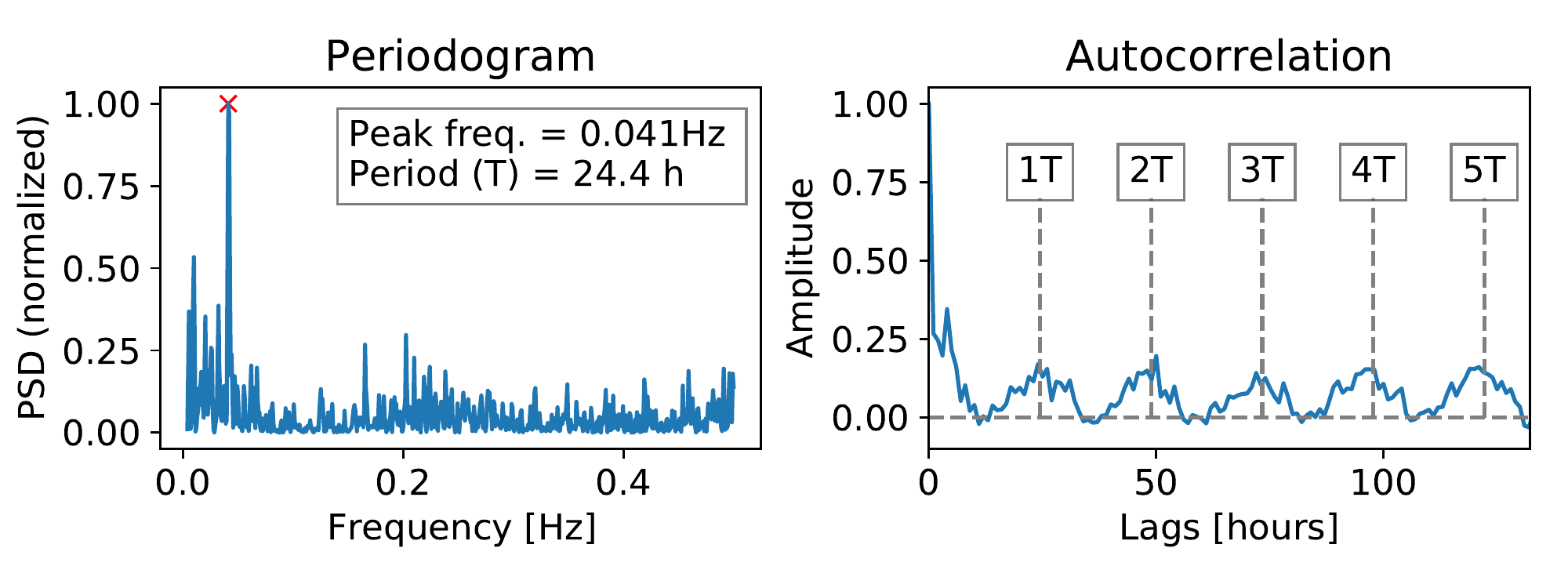}%
}

\subfloat[User $18$, condition $1$]{%
  \includegraphics[clip,width=\columnwidth]{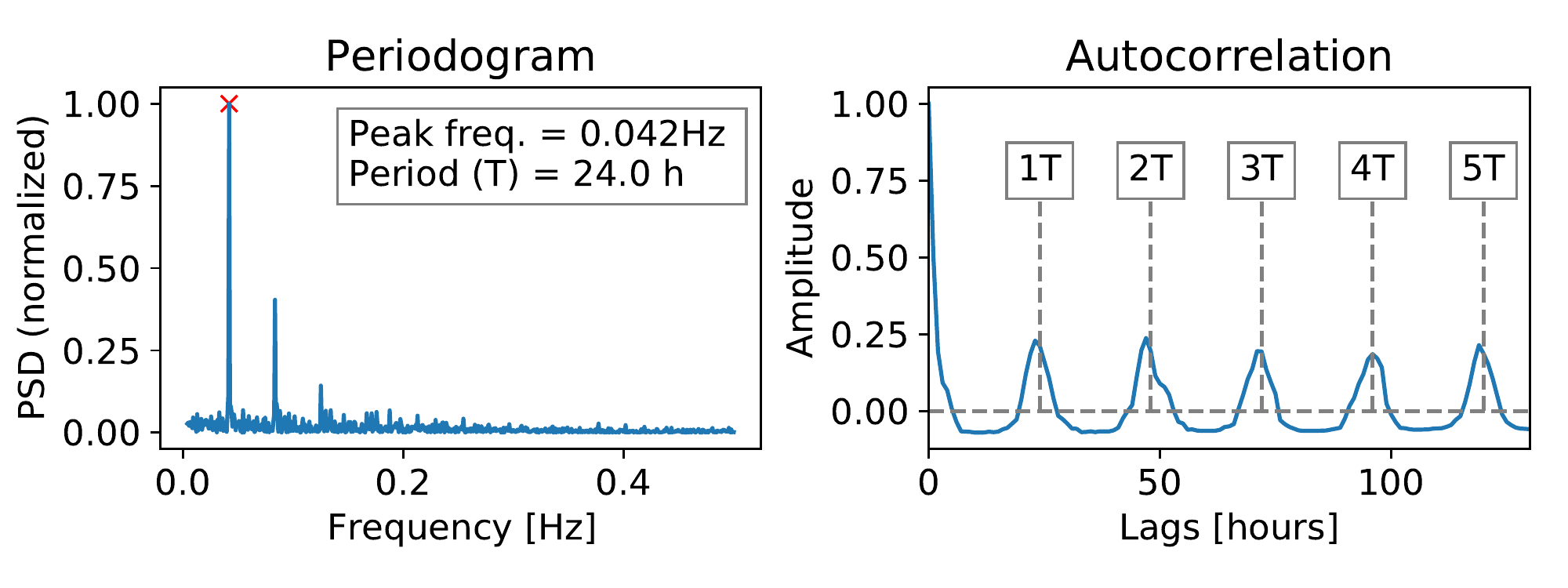}%
}
~
\subfloat[User $18$, condition $2$]{%
  \includegraphics[clip,width=\columnwidth]{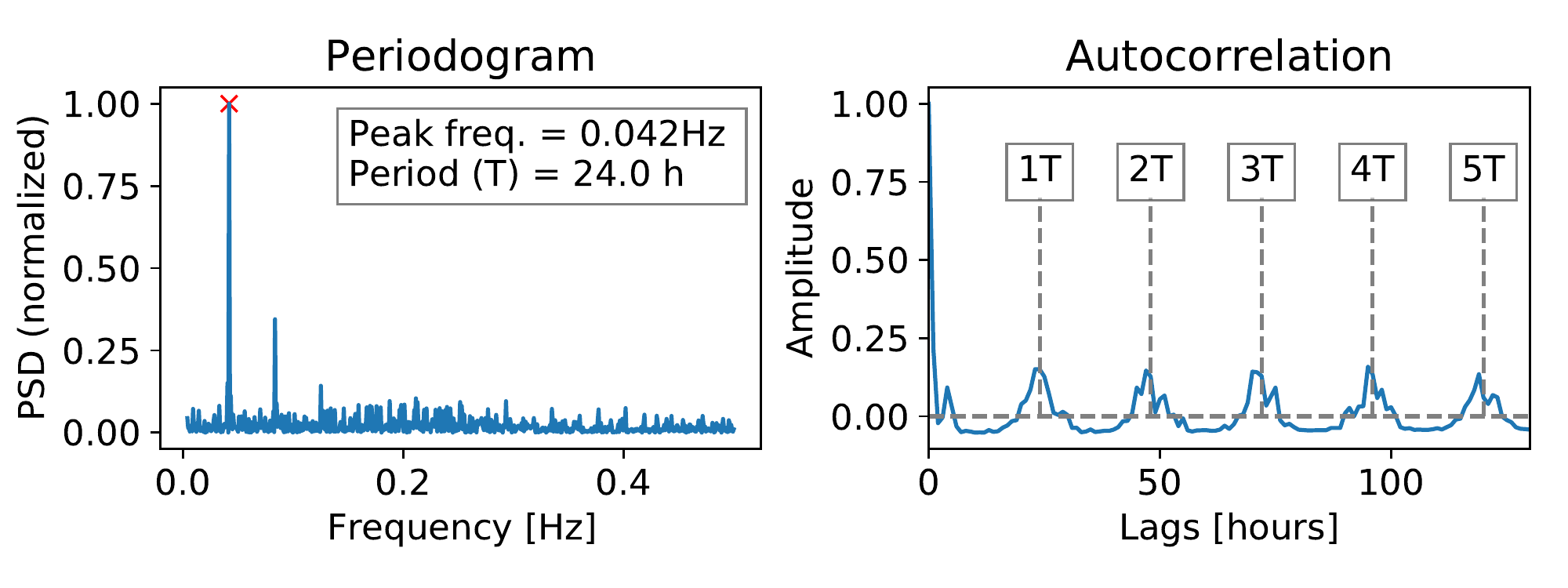}%
}

\end{figure*}

\begin{figure*}[htp]
\ContinuedFloat\centering

\subfloat[User $19$, condition $1$]{%
  \includegraphics[clip,width=\columnwidth]{figs/2021/Time series/periodogram_autocorr_user_19_condition_1.pdf}%
}
~
\subfloat[User $19$, condition $2$]{%
  \includegraphics[clip,width=\columnwidth]{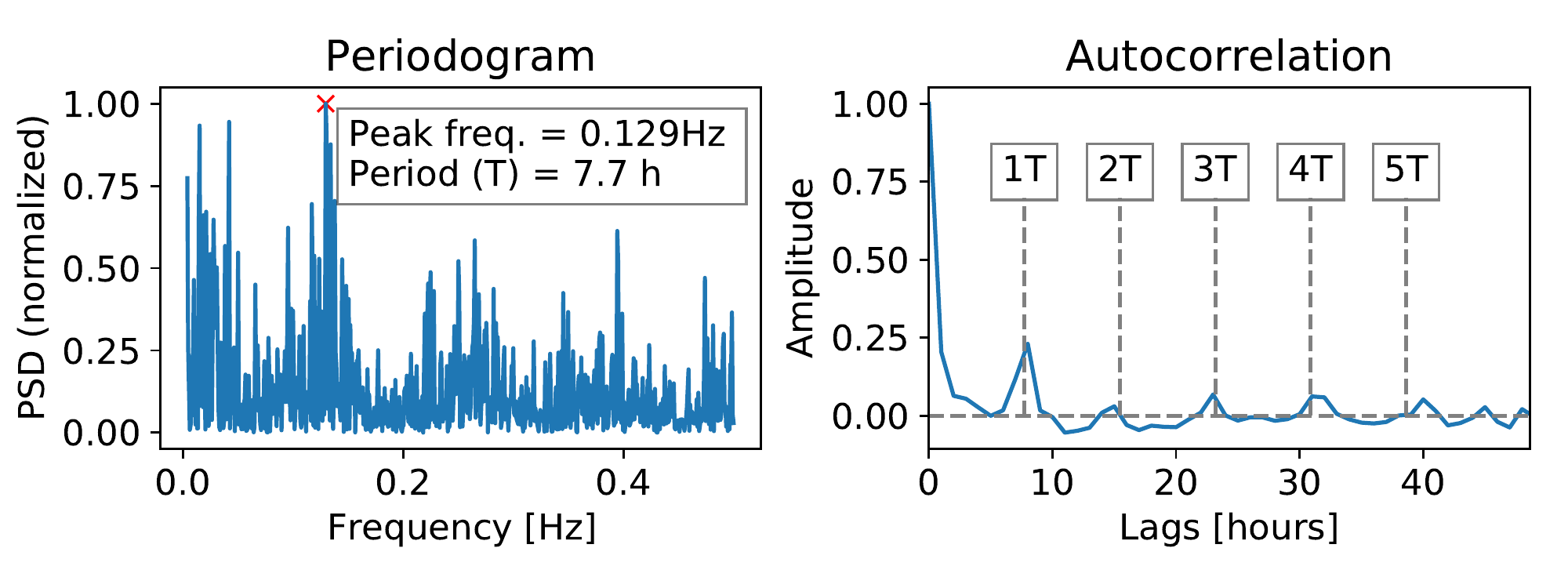}%
}

\subfloat[User $20$, condition $1$]{%
  \includegraphics[clip,width=\columnwidth]{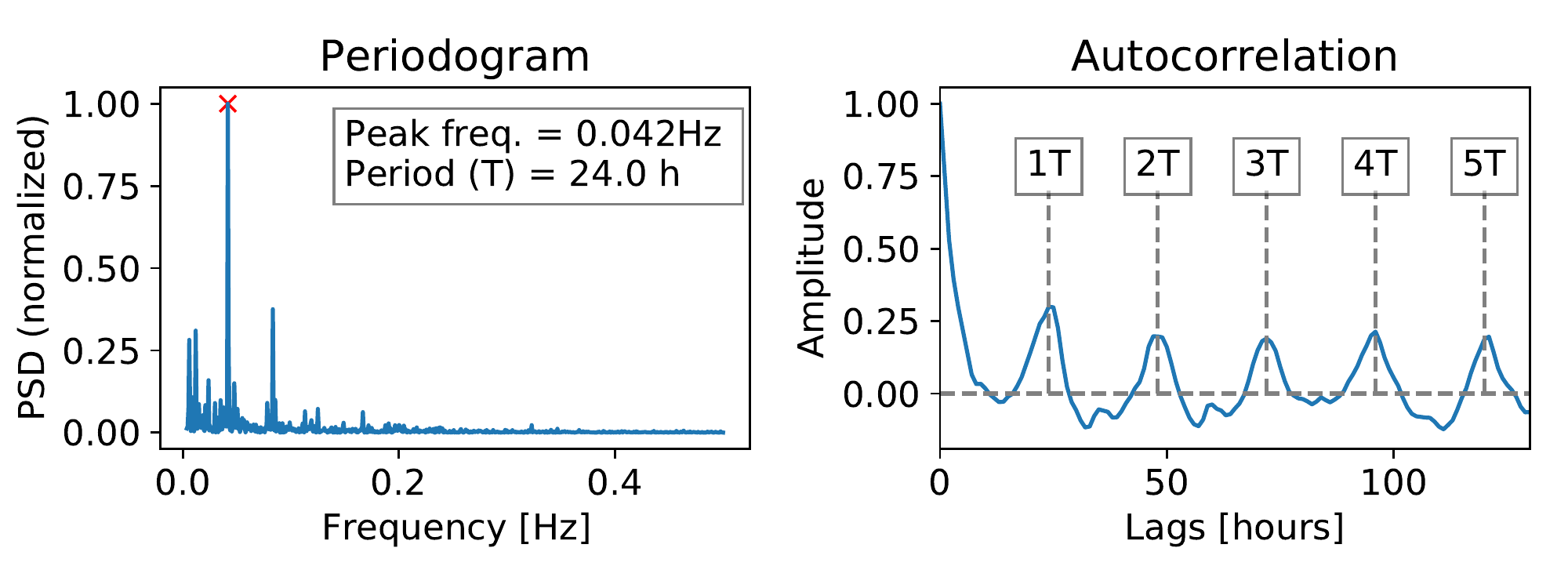}%
}
~
\subfloat[User $20$, condition $2$]{%
  \includegraphics[clip,width=\columnwidth]{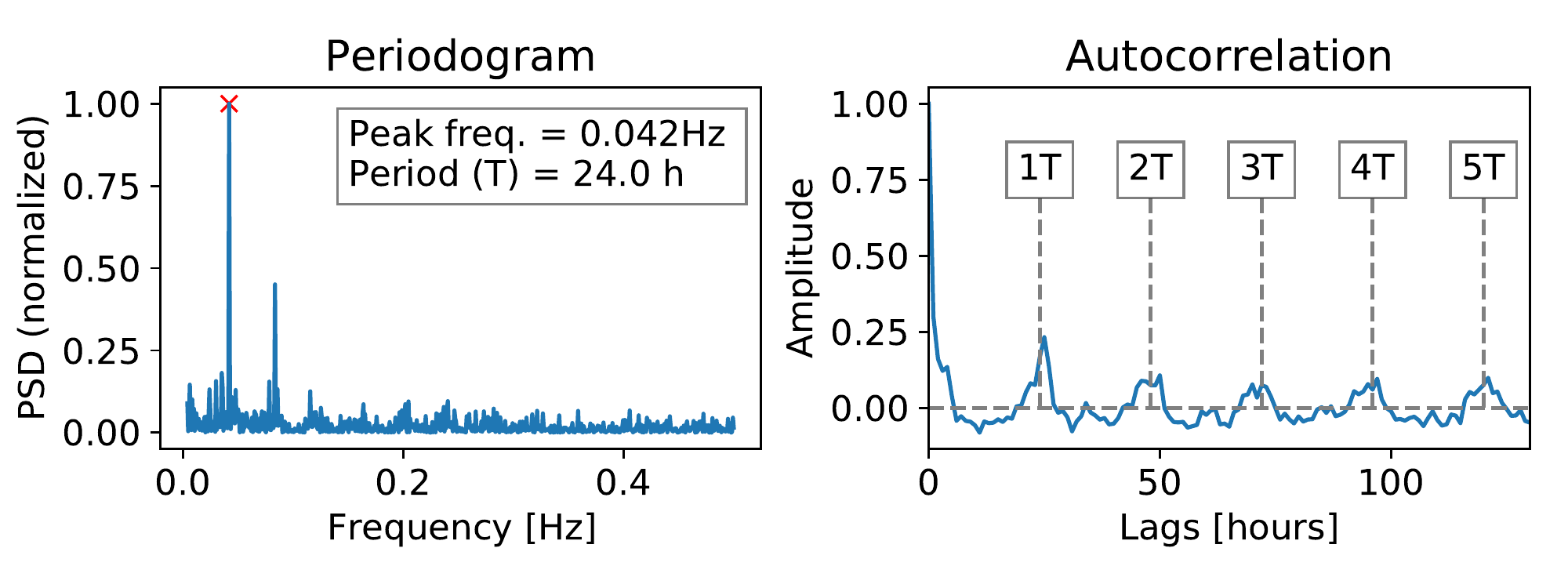}%
}

\subfloat[User $21$, condition $1$]{%
  \includegraphics[clip,width=\columnwidth]{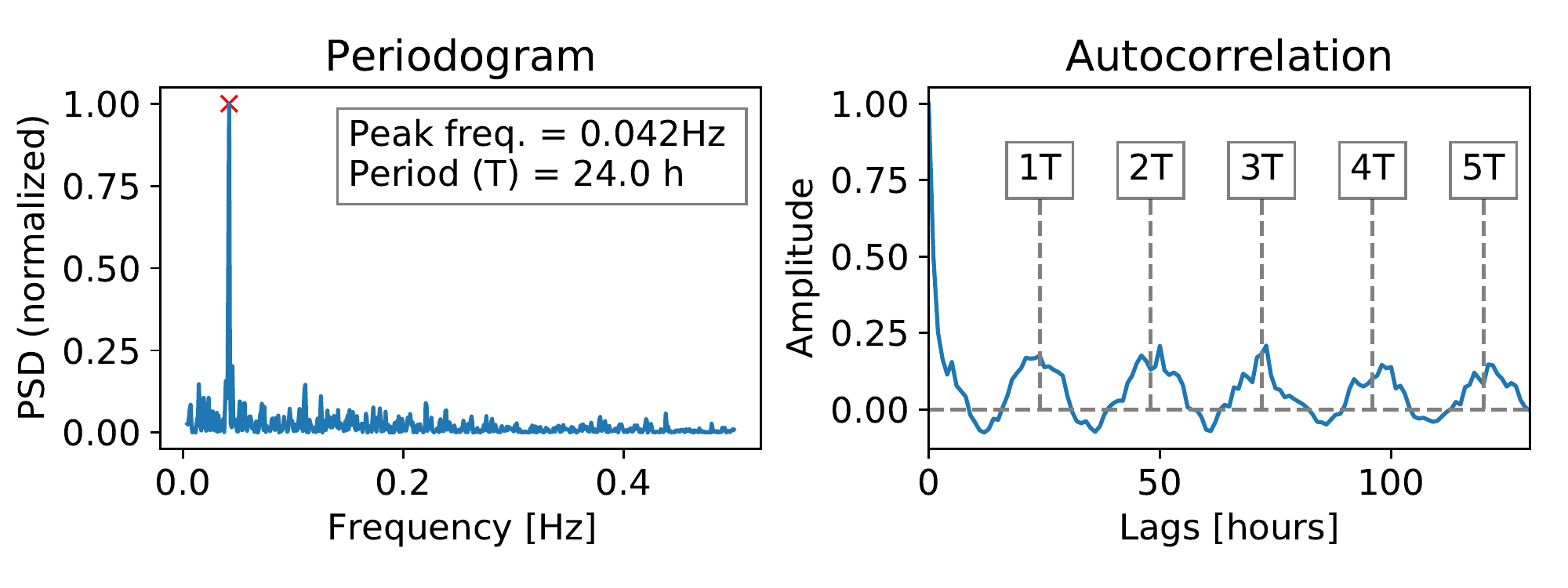}%
}
~
\subfloat[User $21$, condition $2$]{%
  \includegraphics[clip,width=\columnwidth]{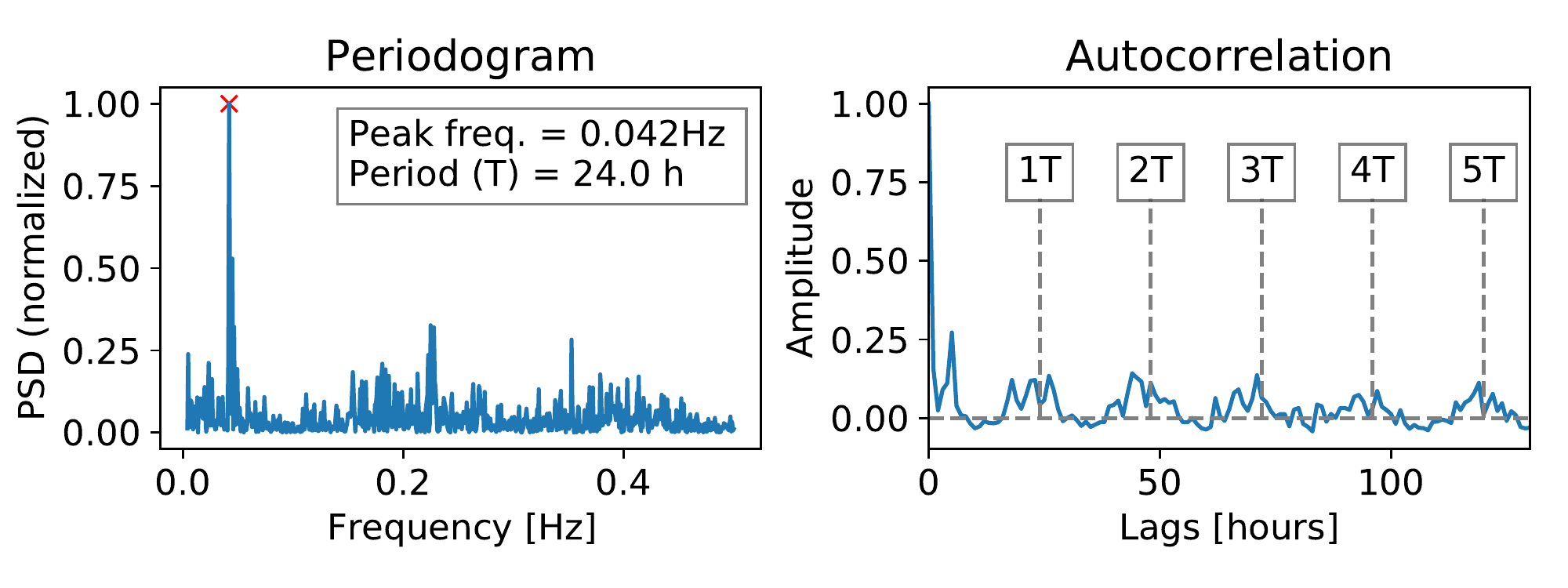}%
}

\subfloat[User $22$, condition $1$]{%
  \includegraphics[clip,width=\columnwidth]{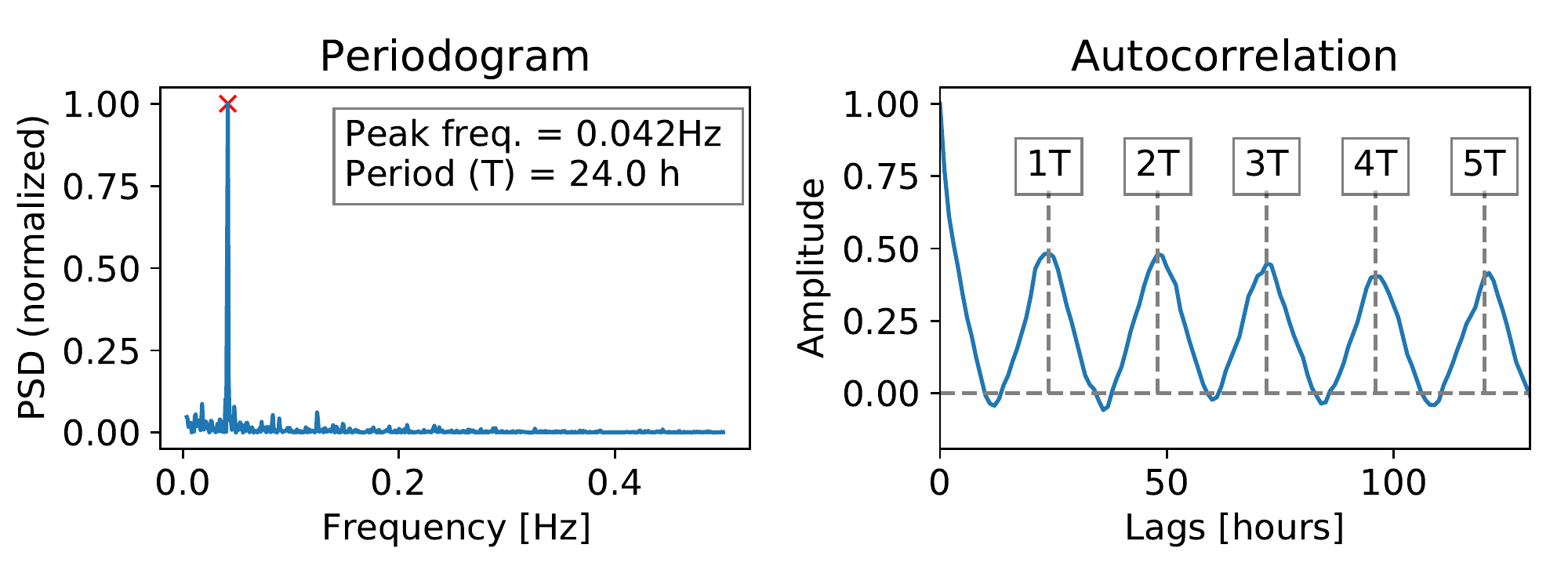}%
}
~
\subfloat[User $22$, condition $2$]{%
  \includegraphics[clip,width=\columnwidth]{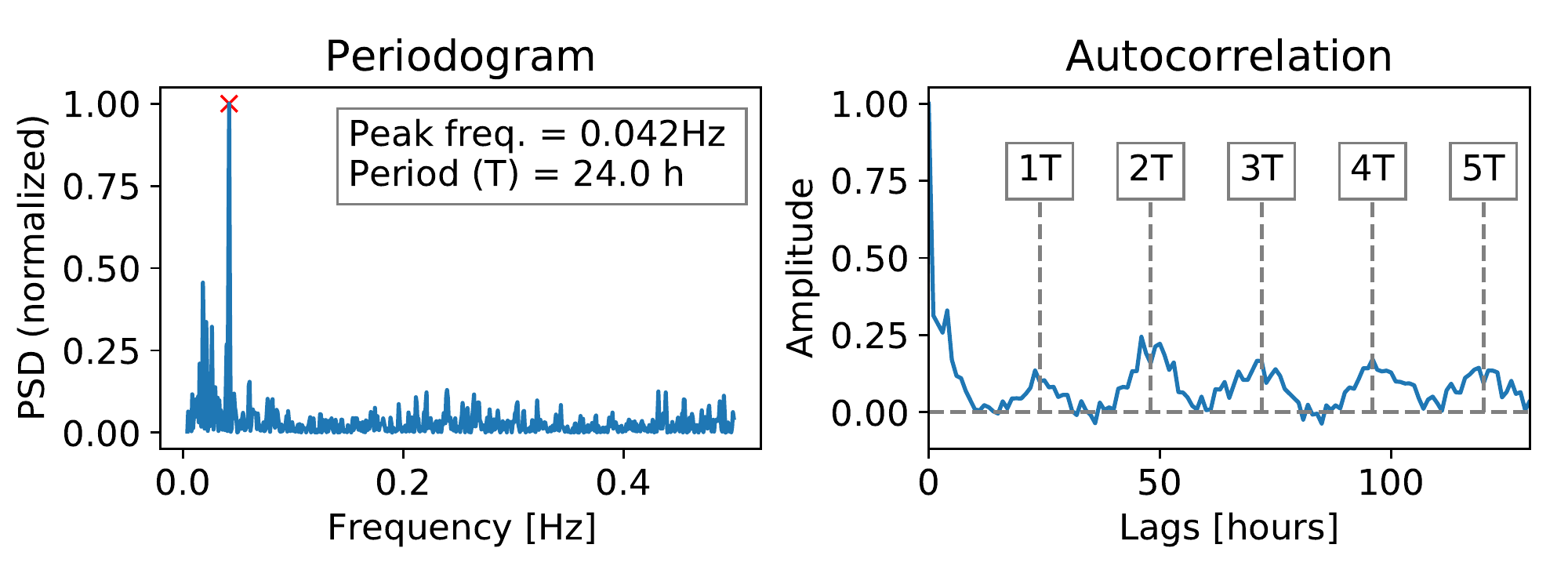}%
}

\subfloat[User $23$, condition $1$]{%
  \includegraphics[clip,width=\columnwidth]{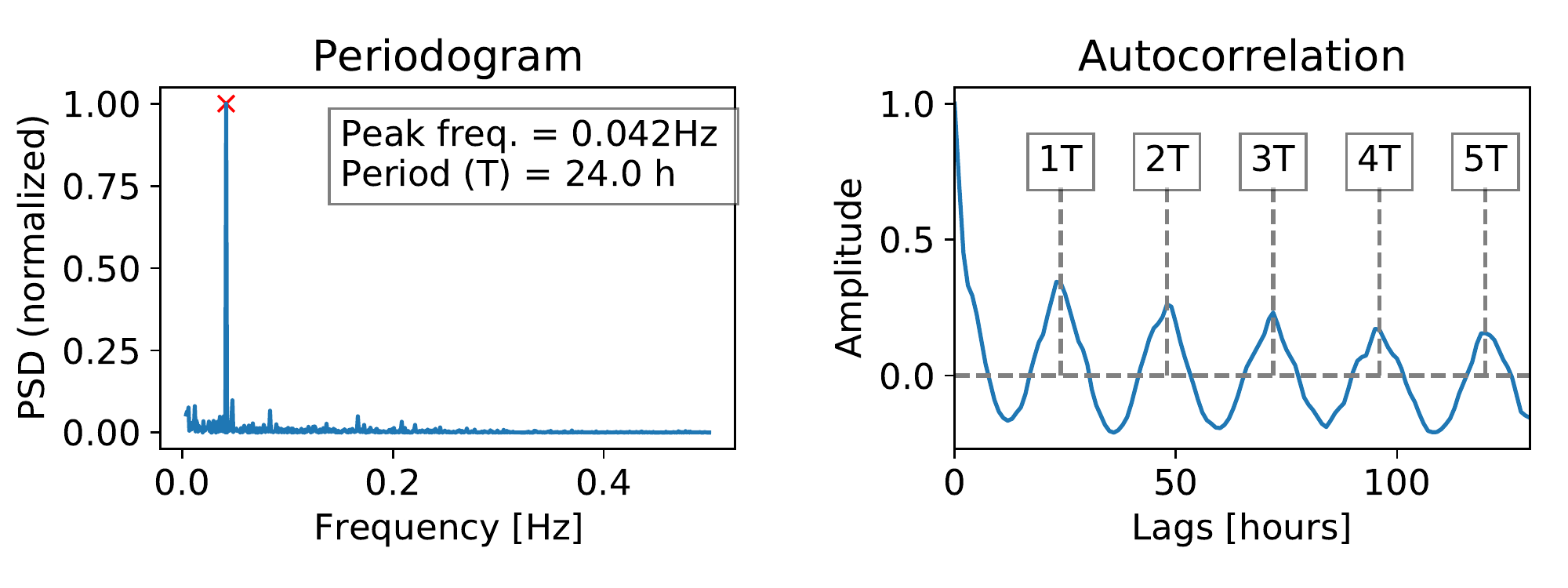}%
}
~
\subfloat[User $23$, condition $2$]{%
  \includegraphics[clip,width=\columnwidth]{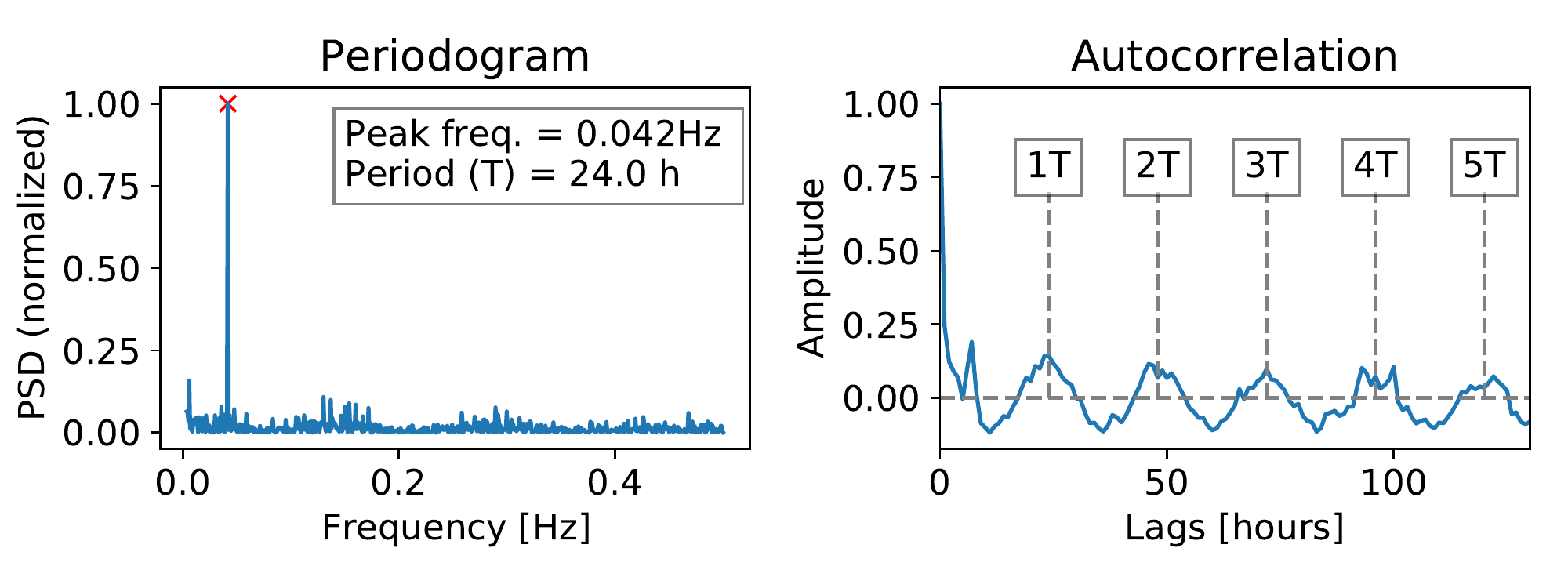}%
}

\subfloat[User $24$, condition $1$]{%
  \includegraphics[clip,width=\columnwidth]{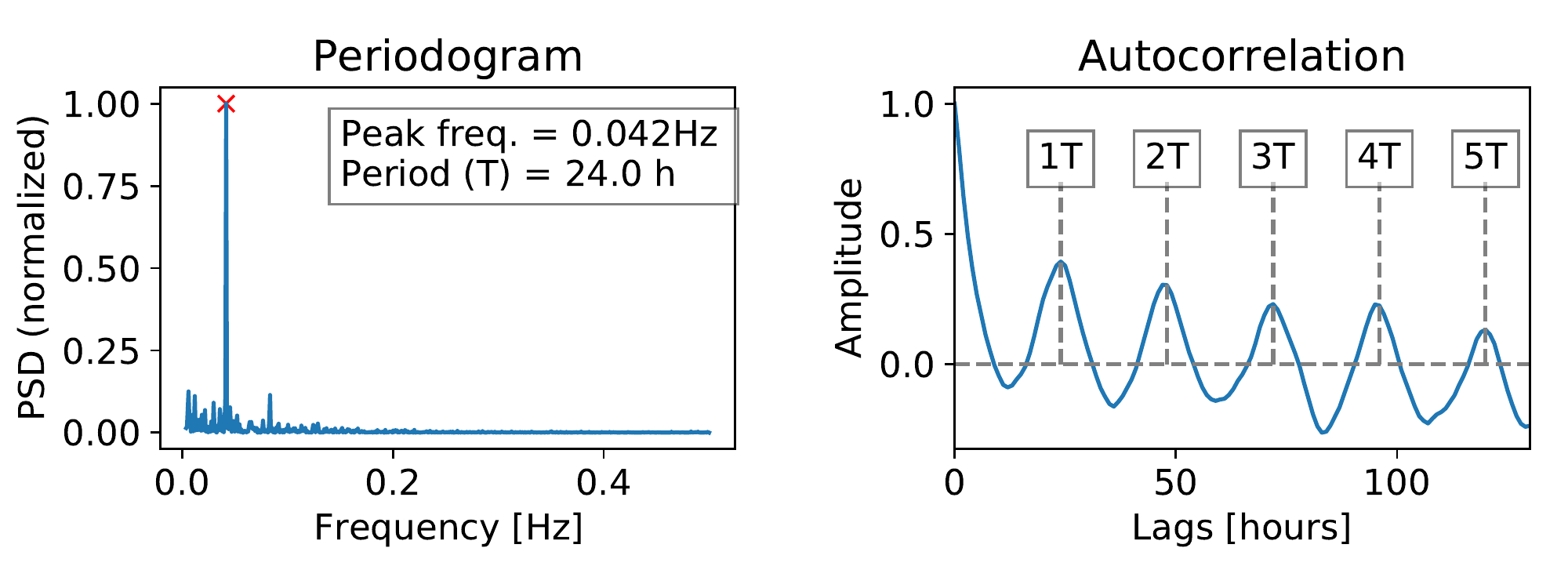}%
}
~
\subfloat[User $24$, condition $2$]{%
  \includegraphics[clip,width=\columnwidth]{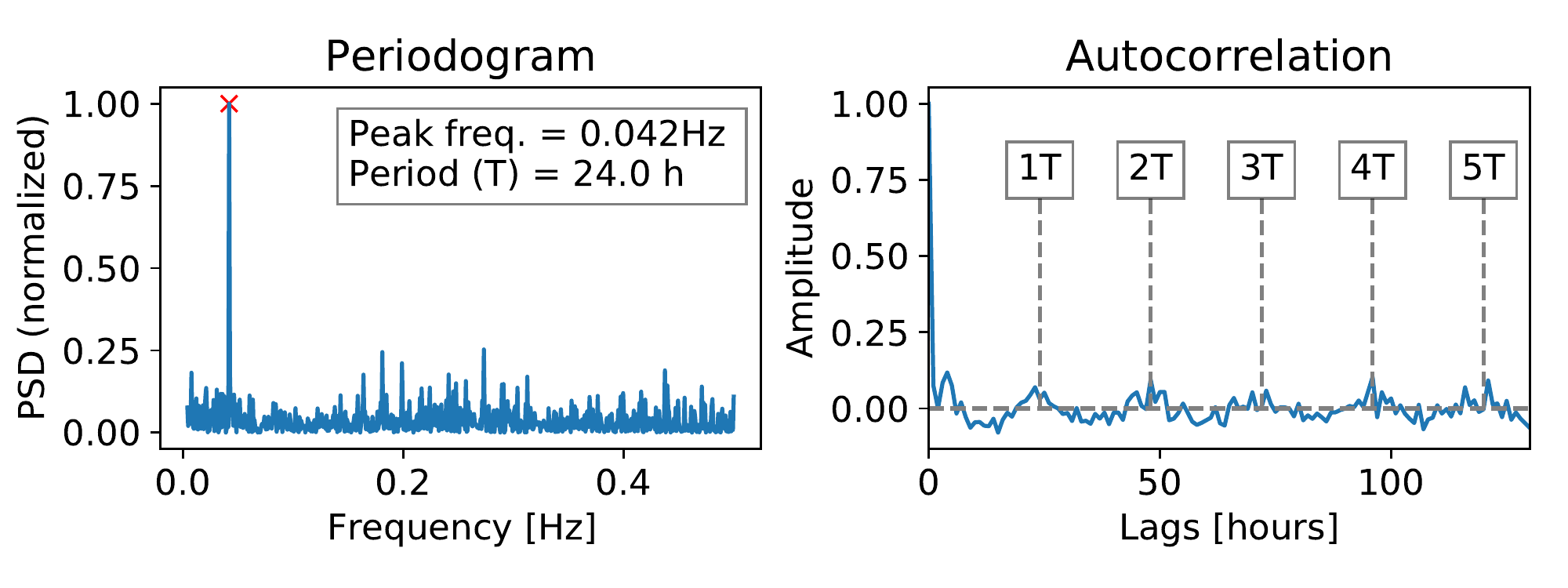}%
}

\end{figure*}

\begin{figure*}[htp]
\ContinuedFloat\centering

\subfloat[User $25$, condition $1$]{%
  \includegraphics[clip,width=\columnwidth]{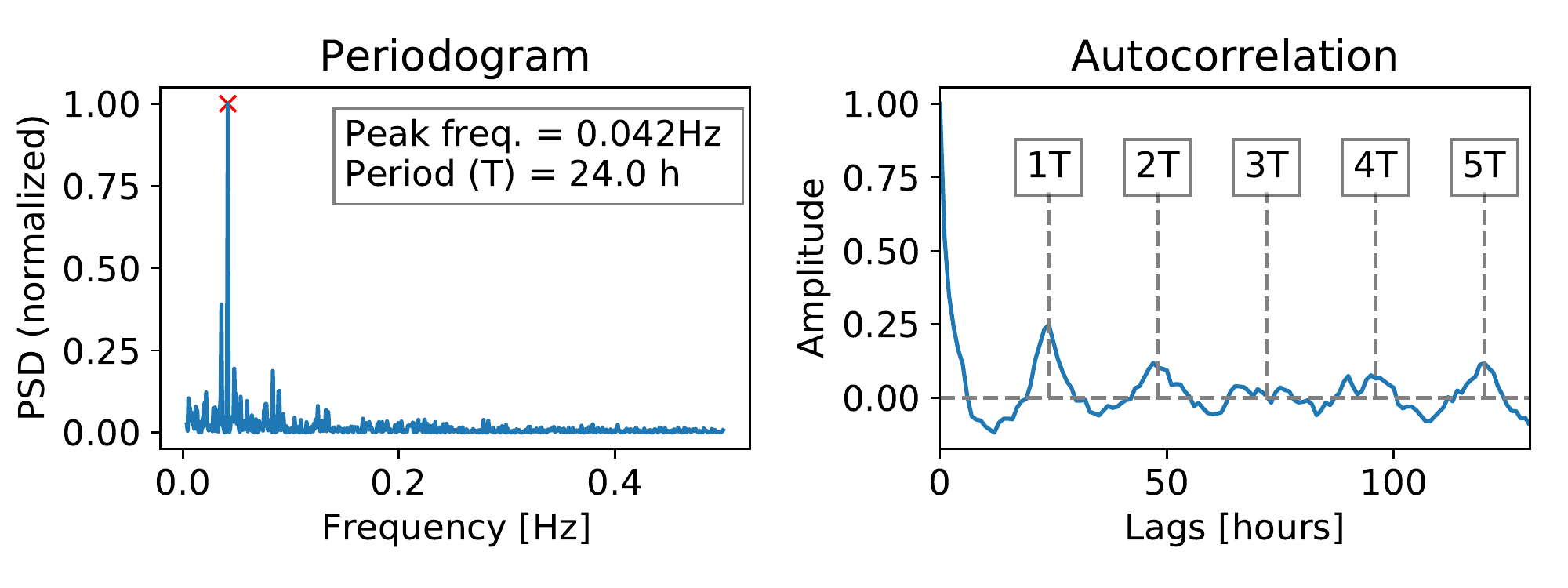}%
}
~
\subfloat[User $25$, condition $2$]{%
  \includegraphics[clip,width=\columnwidth]{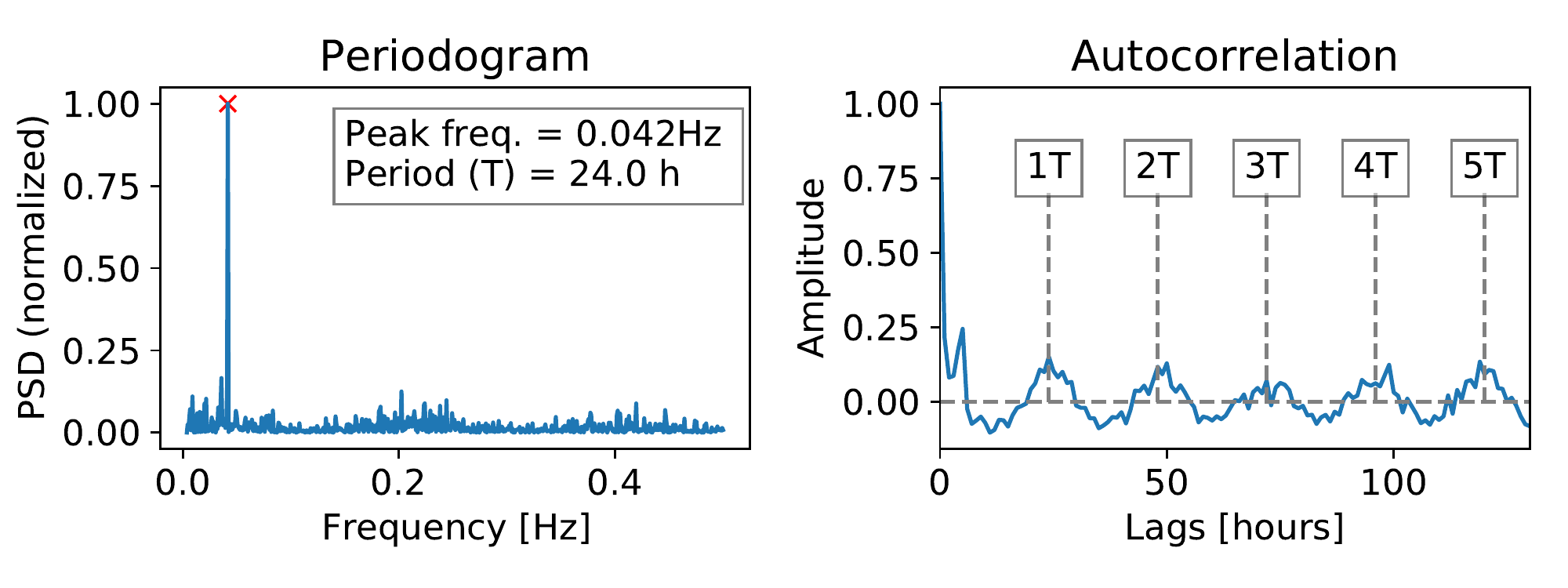}%
}

\subfloat[User $26$, condition $1$]{%
  \includegraphics[clip,width=\columnwidth]{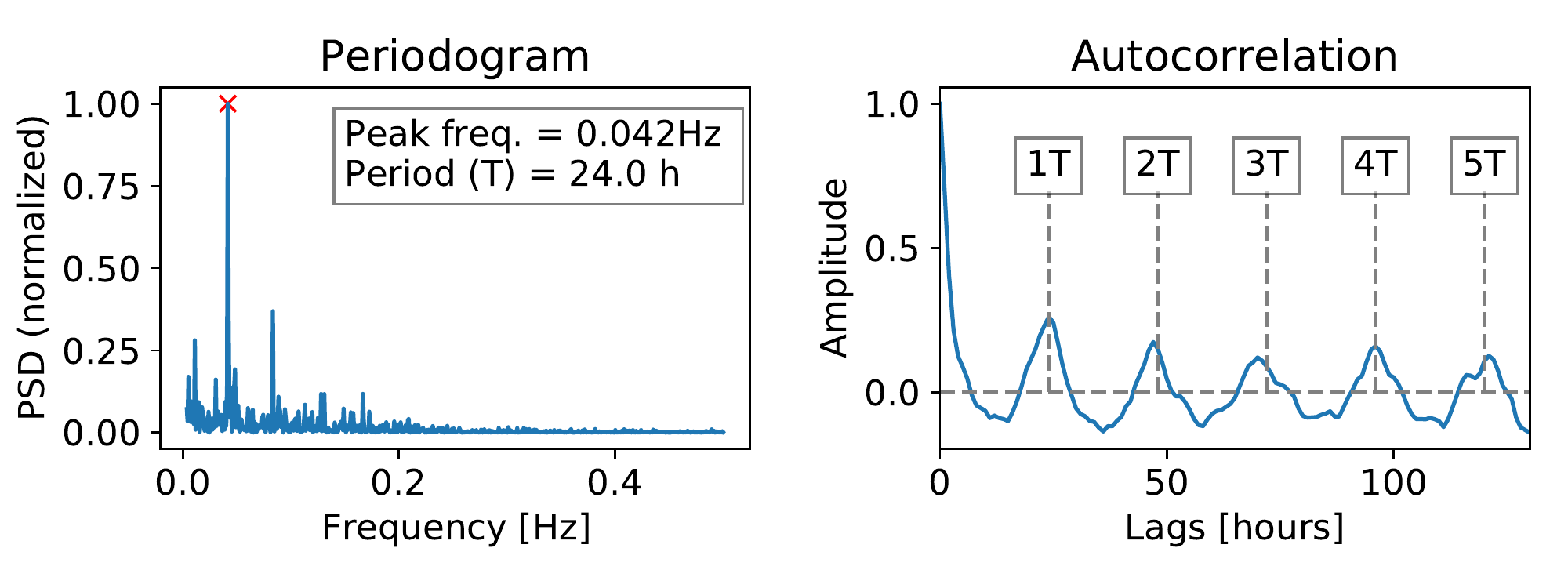}%
}
~
\subfloat[User $26$, condition $2$]{%
  \includegraphics[clip,width=\columnwidth]{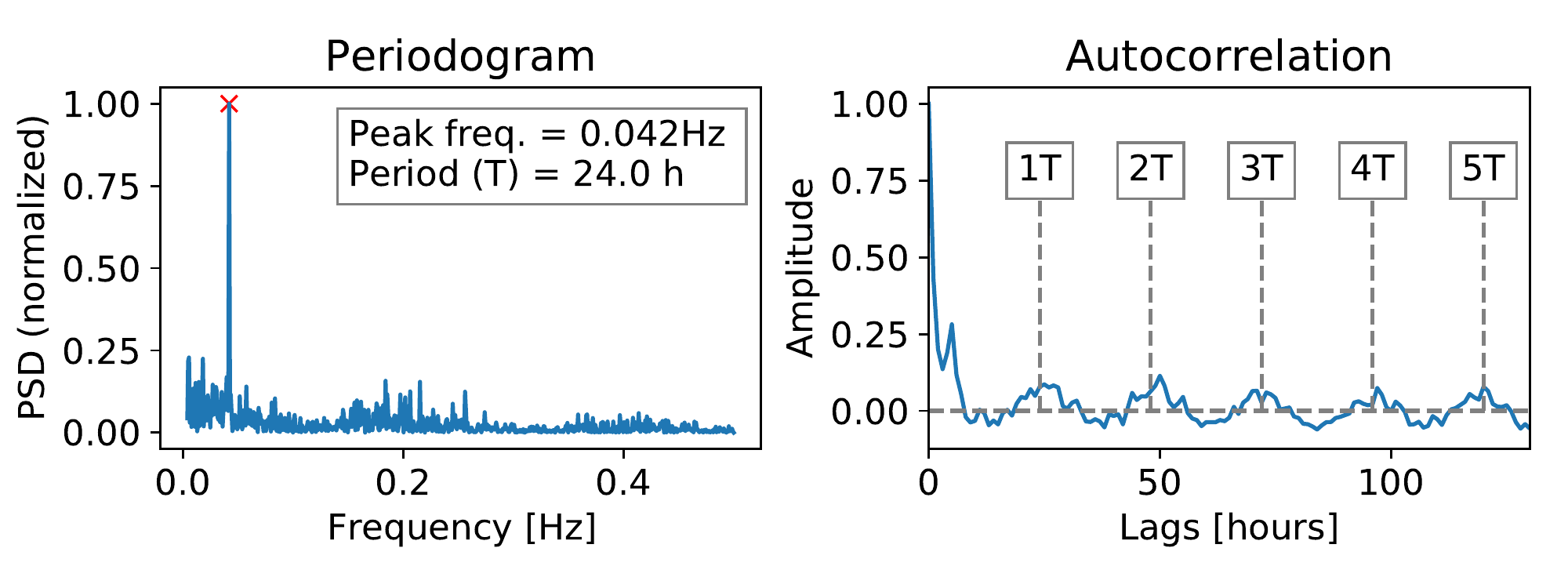}%
}

\subfloat[User $27$, condition $1$]{%
  \includegraphics[clip,width=\columnwidth]{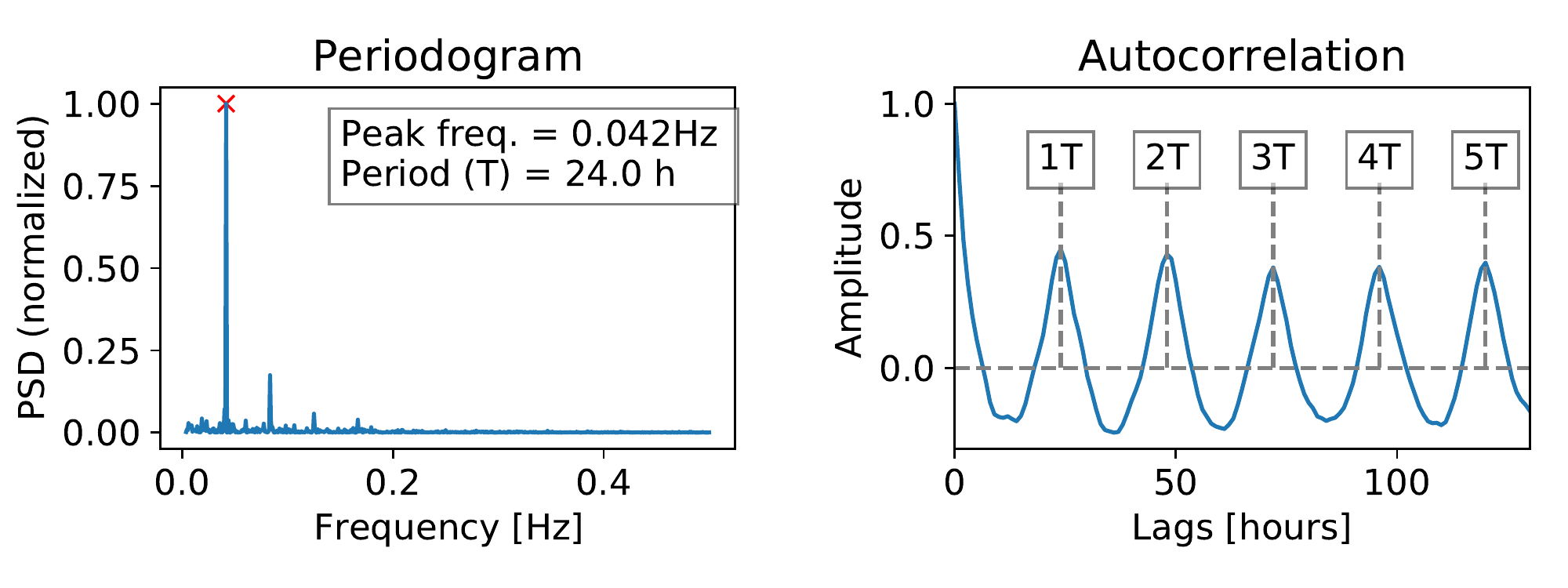}%
}
~
\subfloat[User $27$, condition $2$]{%
  \includegraphics[clip,width=\columnwidth]{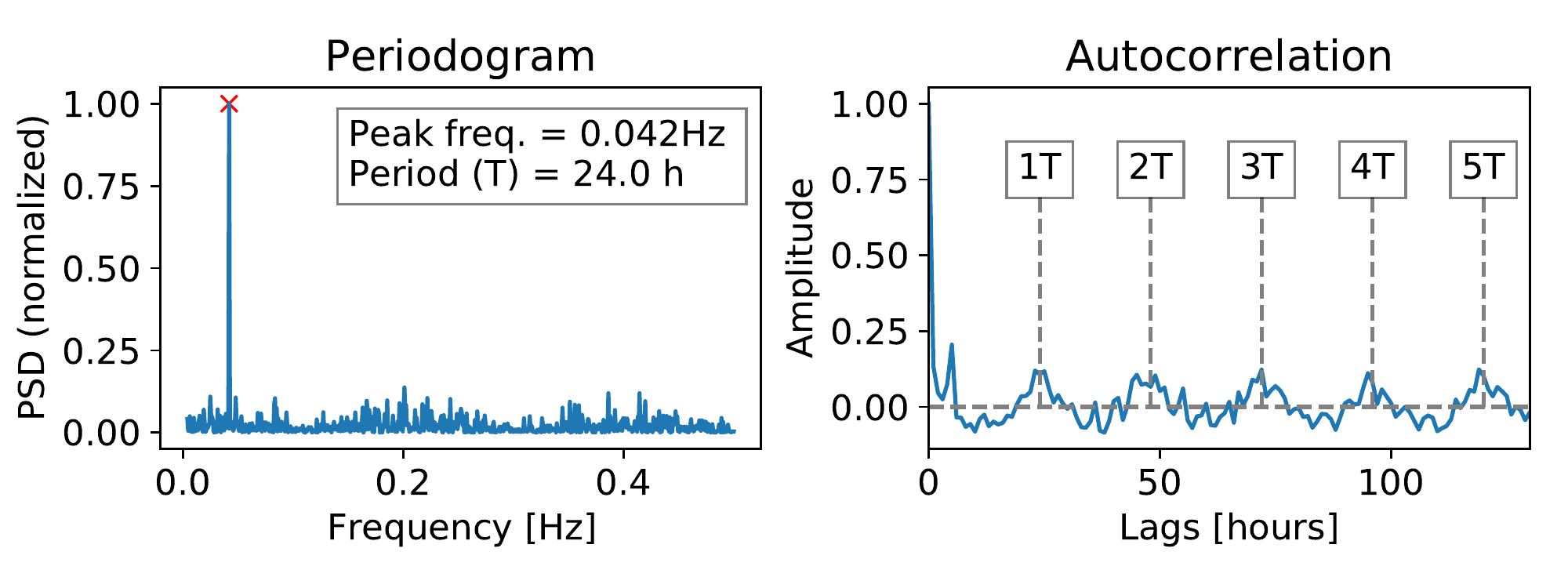}%
}

\subfloat[User $28$, condition $1$]{%
  \includegraphics[clip,width=\columnwidth]{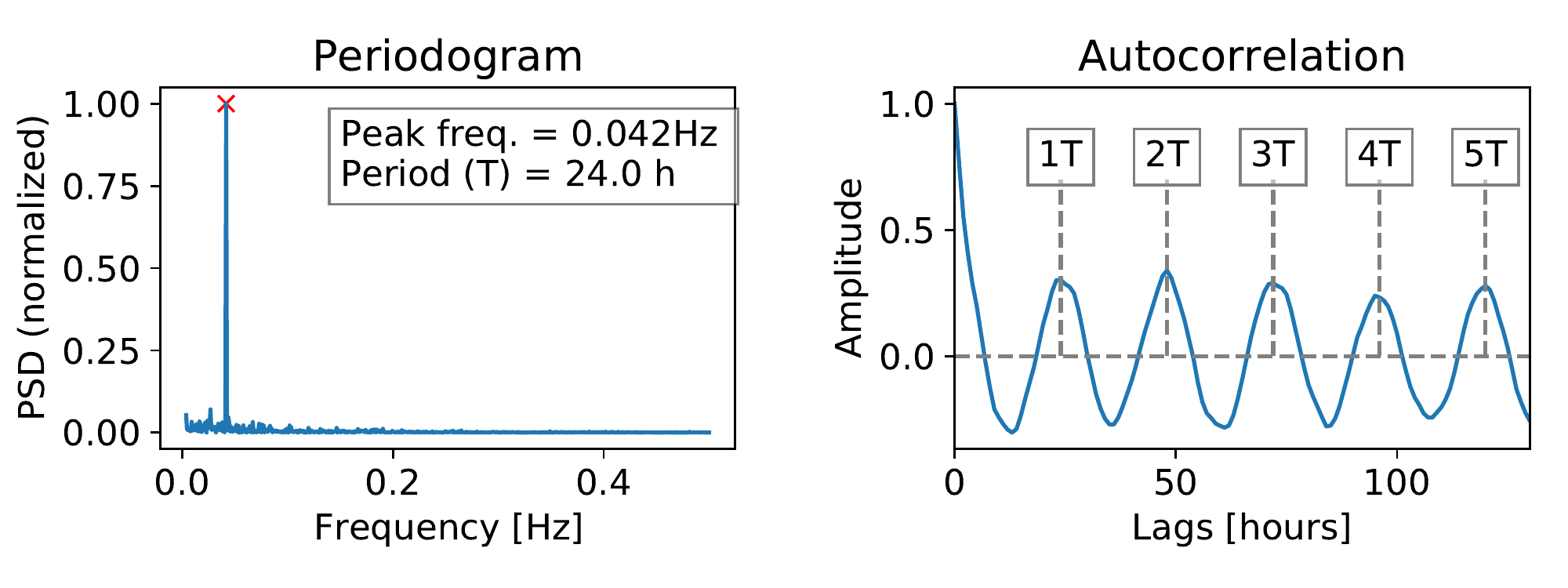}%
}
~
\subfloat[User $28$, condition $2$]{%
  \includegraphics[clip,width=\columnwidth]{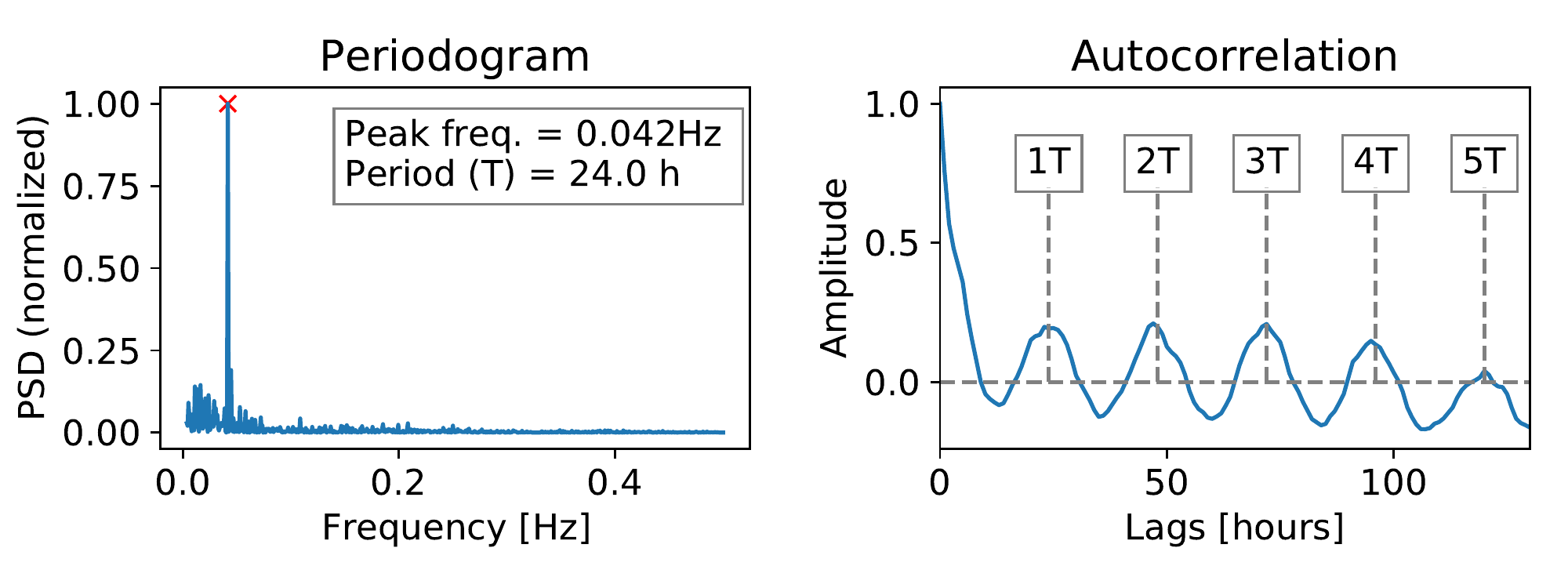}%
}

\subfloat[User $29$, condition $1$]{%
  \includegraphics[clip,width=\columnwidth]{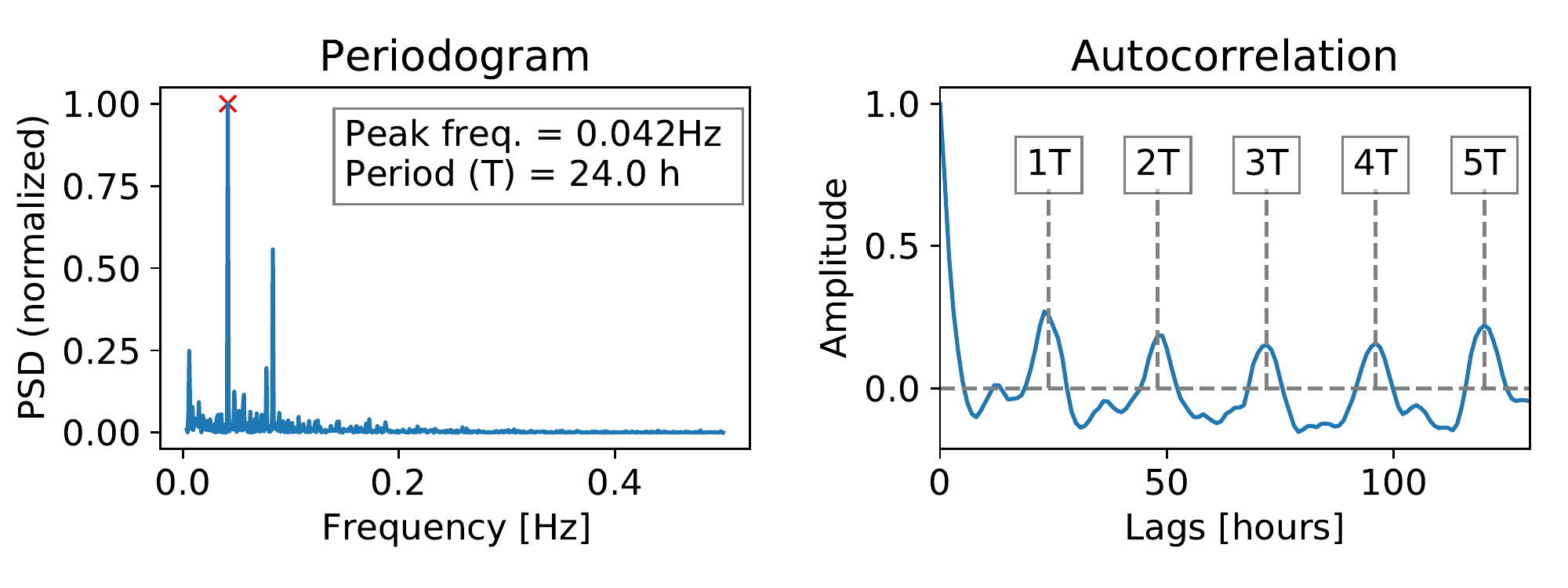}%
}
~
\subfloat[User $29$, condition $2$]{%
  \includegraphics[clip,width=\columnwidth]{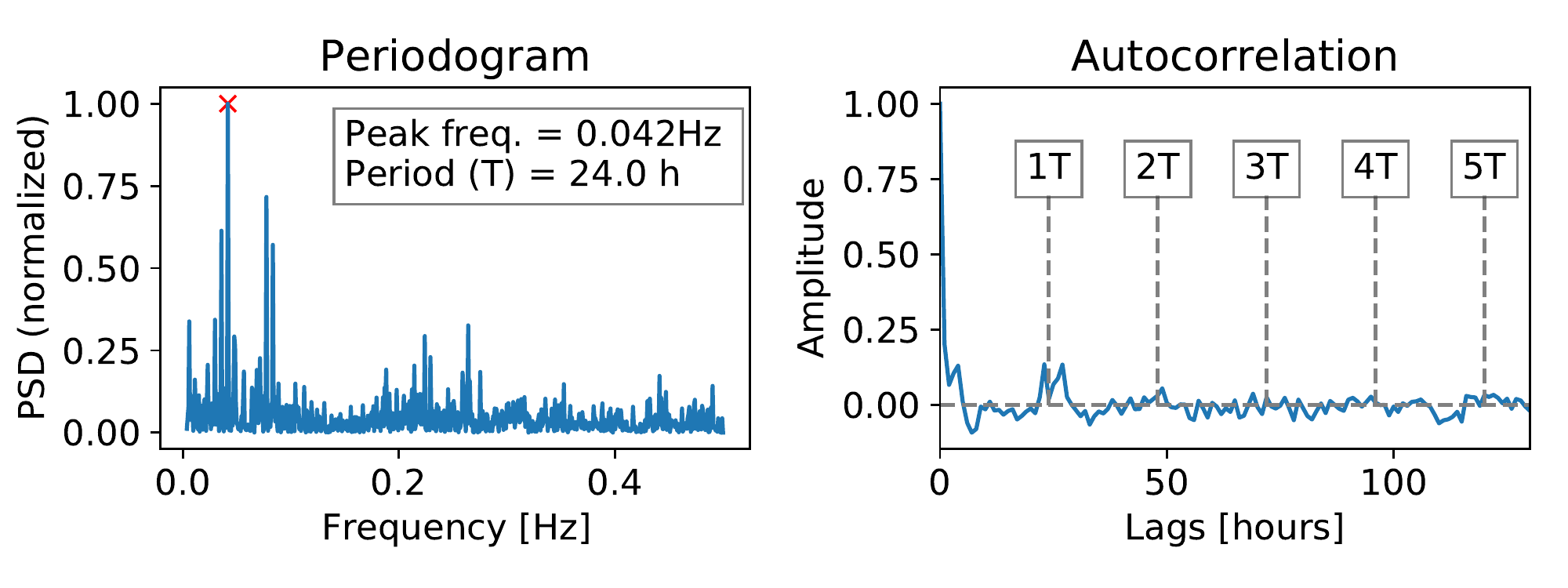}%
}

\end{figure*}

\begin{figure*}[htp]
\ContinuedFloat\centering

\subfloat[User $30$, condition $1$]{%
  \includegraphics[clip,width=\columnwidth]{figs/2021/Time series/periodogram_autocorr_user_30_condition_1.pdf}%
}
~
\subfloat[User $30$, condition $2$]{%
  \includegraphics[clip,width=\columnwidth]{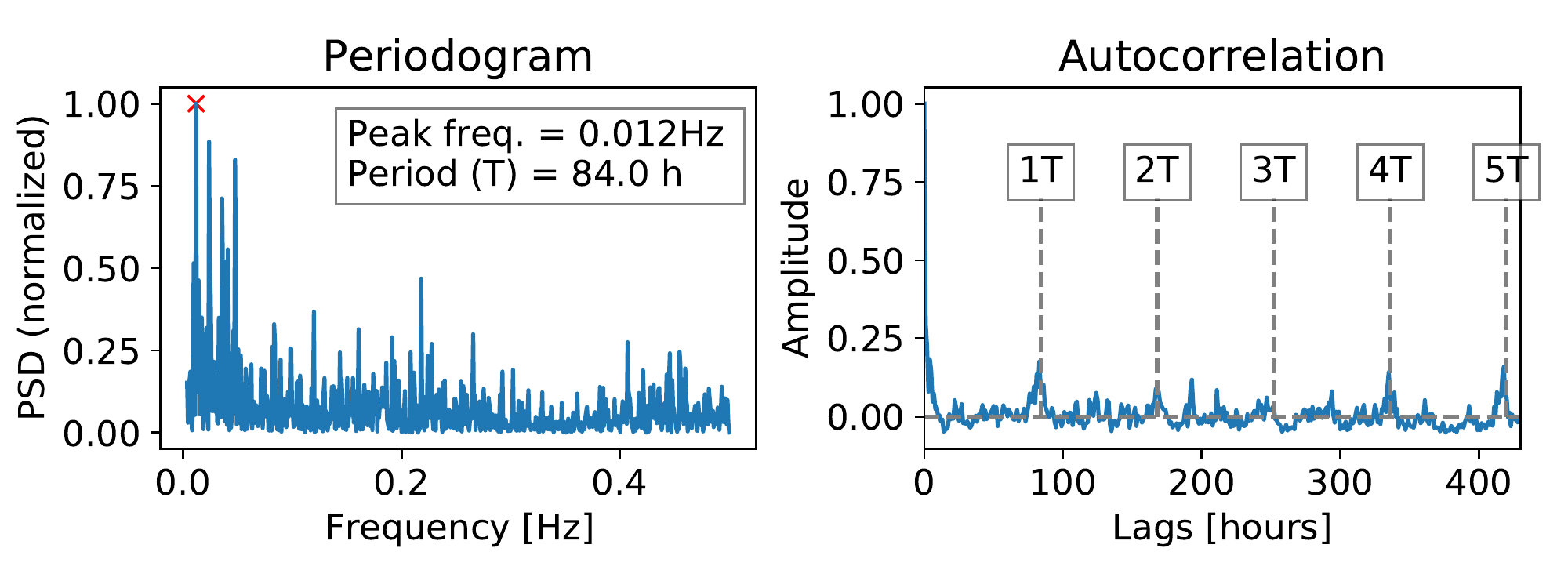}%
}

\subfloat[User $31$, condition $1$]{%
  \includegraphics[clip,width=\columnwidth]{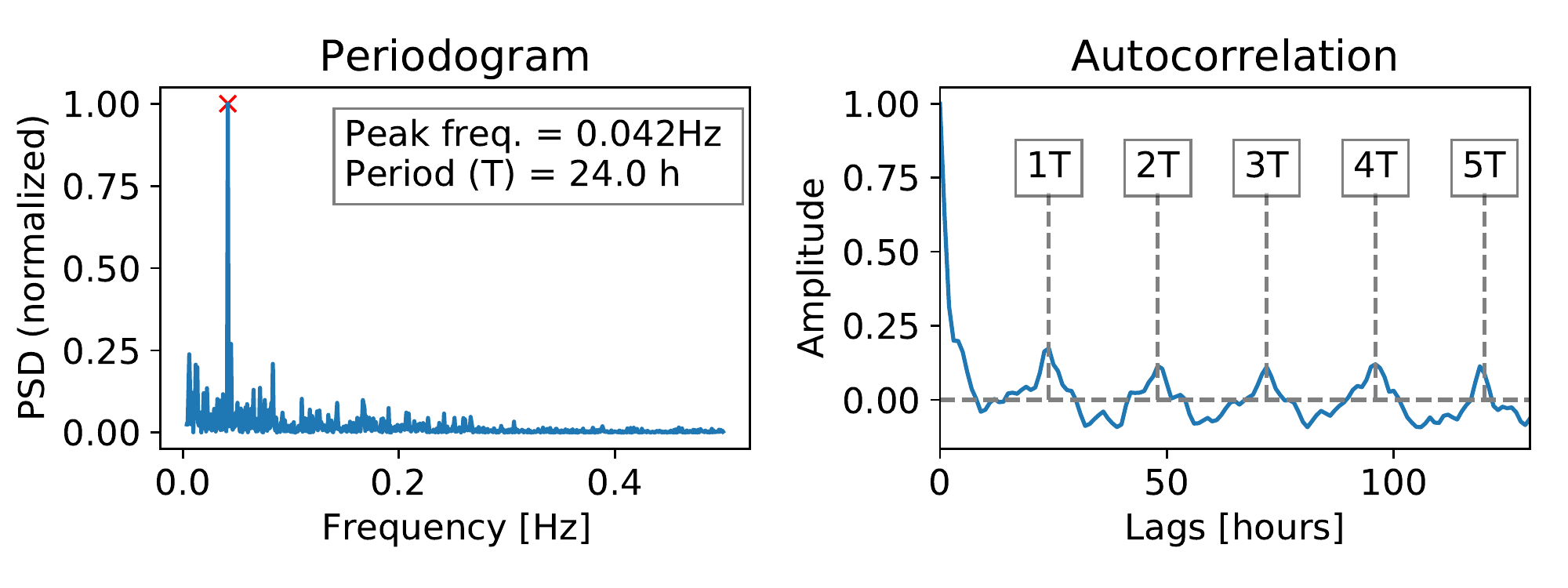}%
}
~
\subfloat[User $31$, condition $2$]{%
  \includegraphics[clip,width=\columnwidth]{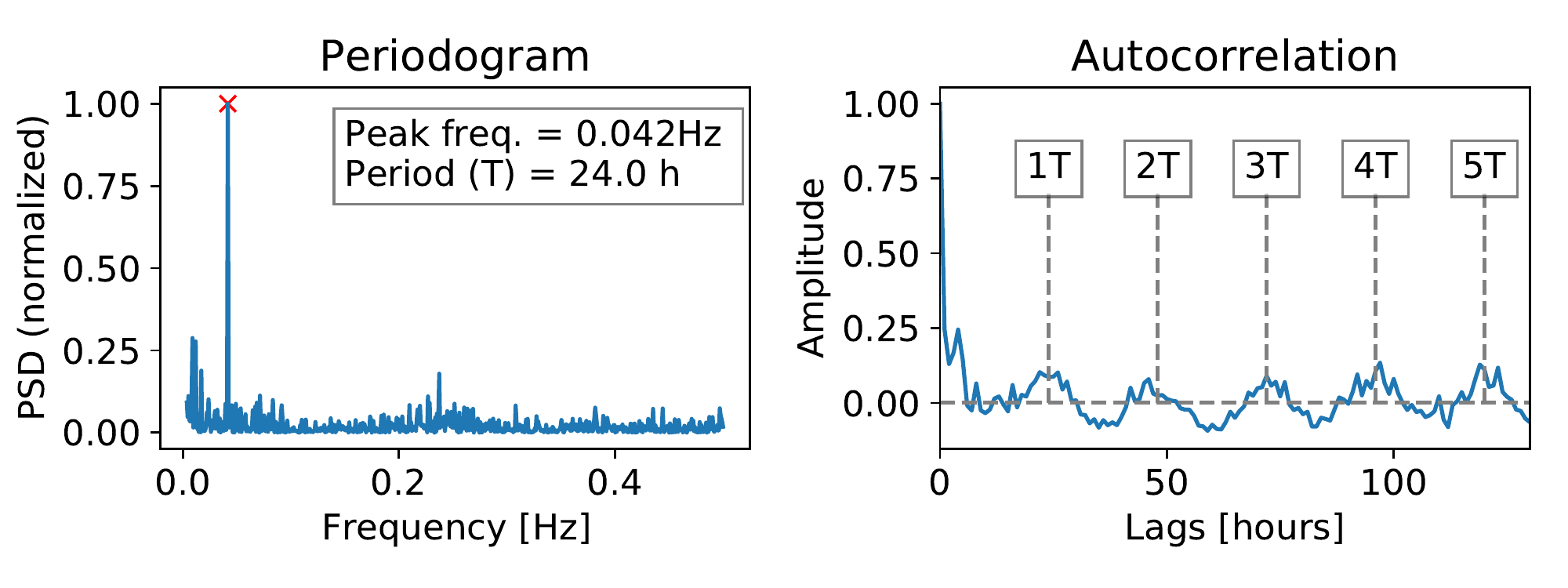}%
}

\caption{\changed{Periodogram and autocorrelation results for all $\samplesize$ participants in condition 1 (with background process activity) and condition 2 (without background process activity).}}

\label{fig:periodogram_autocorr_all_users}

\end{figure*}

\section{Complete Classification Results}
\label{appendix:classification_results}

\changed{Tables~\ref{tab:results_binary_offline},~\ref{tab:results_one_class_offline}, and~\ref{tab:results_online} exhibit the classification results for all window sizes obtained with the offline binary models, offine one-class models, and online binary models, respectively.}

\begin{table*}[]
\centering
\caption{\changed{Offline Binary Classification Results.}}
\label{tab:results_binary_offline}
\begin{tabular}{@{}cccccccc@{}}
\toprule
\multirow{2}{*}{\textbf{Classifier}} & \multirow{2}{*}{\textbf{\begin{tabular}[c]{@{}c@{}}Window\\ Size\end{tabular}}} & \multicolumn{2}{c}{\textbf{F-Score}} & \multicolumn{2}{c}{\textbf{Recall}} & \multicolumn{2}{c}{\textbf{Precision}} \\
                                     &                                                                                 & \textbf{Mean}    & \textbf{95\% CI}   & \textbf{Mean}    & \textbf{95\% CI}  & \textbf{Mean}     & \textbf{95\% CI}    \\ \midrule
\multirow{6}{*}{Random Forest}       & 1                                                                               & 75.94\%           & [74.82\%, 77.05\%]  & 75.10\%           & [73.96\%, 76.24\%] & 79.58\%            & [78.35\%, 80.82\%]   \\
                                     & 2                                                                               & 83.08\%           & [81.98\%, 84.18\%]  & 83.68\%           & [82.6\%, 84.75\%]  & 84.94\%            & [83.73\%, 86.16\%]   \\
                                     & 5                                                                               & 90.66\%           & [89.73\%, 91.6\%]   & 90.12\%           & [89.05\%, 91.18\%] & 93.11\%            & [92.16\%, 94.07\%]   \\
                                     & 10                                                                              & 92.34\%           & [91.24\%, 93.45\%]  & 92.00\%           & [90.81\%, 93.2\%]  & 94.50\%            & [93.44\%, 95.57\%]   \\
                                     & 30                                                                              & 94.37\%           & [93.31\%, 95.42\%]  & 93.51\%           & [92.35\%, 94.66\%] & 97.35\%            & [96.61\%, 98.08\%]   \\
                                     & 60                                                                              & \textbf{95.00\%}  & [94\%, 96\%]        & \textbf{94.03\%}  & [92.9\%, 95.16\%]  & \textbf{98.36\%}   & [97.87\%, 98.85\%]   \\ \midrule
\multirow{6}{*}{SGD}                 & 1                                                                               & 76.03\%           & [74.95\%, 77.11\%]  & 73.45\%           & [72.32\%, 74.59\%] & 82.71\%            & [81.5\%, 83.92\%]    \\
                                     & 2                                                                               & 83.85\%           & [82.71\%, 84.99\%]  & 83.37\%           & [82.27\%, 84.47\%] & 86.44\%            & [85.14\%, 87.74\%]   \\
                                     & 5                                                                               & 89.88\%           & [88.7\%, 91.05\%]   & 90.54\%           & [89.43\%, 91.64\%] & \textbf{90.66\%}   & [89.37\%, 91.95\%]   \\
                                     & 10                                                                              & \textbf{90.38\%}  & [89.17\%, 91.59\%]  & 92.33\%           & [91.21\%, 93.45\%] & 90.10\%            & [88.77\%, 91.43\%]   \\
                                     & 30                                                                              & 89.58\%           & [88.33\%, 90.84\%]  & 93.32\%           & [92.23\%, 94.41\%] & 88.04\%            & [86.66\%, 89.41\%]   \\
                                     & 60                                                                              & 88.13\%           & [86.88\%, 89.38\%]  & \textbf{93.52\%}  & [92.46\%, 94.58\%] & 85.75\%            & [84.35\%, 87.15\%]   \\ \midrule
\multirow{6}{*}{MLP}                 & 1                                                                               & 77.07\%           & [76.01\%, 78.12\%]  & 78.04\%           & [77.02\%, 79.06\%] & 77.60\%            & [76.43\%, 78.77\%]   \\
                                     & 2                                                                               & 83.00\%           & [81.92\%, 84.07\%]  & 86.06\%           & [85.08\%, 87.03\%] & 81.81\%            & [80.61\%, 83\%]      \\
                                     & 5                                                                               & 89.76\%           & [88.73\%, 90.8\%]   & 91.72\%           & [90.91\%, 92.54\%] & 89.29\%            & [88.12\%, 90.45\%]   \\
                                     & 10                                                                              & \textbf{91.94\%}  & [90.89\%, 92.99\%]  & \textbf{92.88\%}  & [92.02\%, 93.73\%] & 92.37\%            & [91.2\%, 93.54\%]    \\
                                     & 30                                                                              & 91.85\%           & [90.65\%, 93.04\%]  & 92.53\%           & [91.39\%, 93.68\%] & 92.49\%            & [91.24\%, 93.74\%]   \\
                                     & 60                                                                              & 91.76\%           & [90.62\%, 92.91\%]  & 91.70\%           & [90.46\%, 92.93\%] & \textbf{93.62\%}   & [92.59\%, 94.65\%]   \\ \midrule
\multirow{6}{*}{LinearSVC}           & 1                                                                               & 77.19\%           & [76.15\%, 78.23\%]  & 76.67\%           & [75.61\%, 77.73\%] & 79.47\%            & [78.31\%, 80.63\%]   \\
                                     & 2                                                                               & 84.52\%           & [83.4\%, 85.63\%]   & 85.56\%           & [84.52\%, 86.6\%]  & 84.93\%            & [83.69\%, 86.17\%]   \\
                                     & 5                                                                               & 90.10\%           & [88.97\%, 91.22\%]  & 91.19\%           & [90.17\%, 92.21\%] & 90.40\%            & [89.17\%, 91.63\%]   \\
                                     & 10                                                                              & \textbf{91.92\%}  & [90.86\%, 92.98\%]  & \textbf{93.30\%}  & [92.43\%, 94.17\%] & \textbf{92.00\%}   & [90.82\%, 93.19\%]   \\
                                     & 30                                                                              & 90.66\%           & [89.42\%, 91.9\%]   & 93.18\%           & [92.04\%, 94.33\%] & 90.01\%            & [88.68\%, 91.33\%]   \\
                                     & 60                                                                              & 90.79\%           & [89.56\%, 92.02\%]  & 93.24\%           & [92.07\%, 94.42\%] & 90.09\%            & [88.8\%, 91.39\%]    \\ \bottomrule
\end{tabular}
\end{table*}

\begin{table*}[]
    \centering
    \caption{\changed{Offline One-Class Classification Results.}}
    \label{tab:results_one_class_offline}
    \begin{tabular}{@{}cccccccc@{}}
    \toprule
    \multirow{2}{*}{\textbf{Classifier}}             & \multirow{2}{*}{\textbf{\begin{tabular}[c]{@{}c@{}}Window\\ Size\end{tabular}}} & \multicolumn{2}{c}{\textbf{F-Score}} & \multicolumn{2}{c}{\textbf{Recall}} & \multicolumn{2}{c}{\textbf{Precision}} \\
                                                     &                                                                                 & \textbf{Mean}    & \textbf{95\% CI}   & \textbf{Mean}    & \textbf{95\% CI}  & \textbf{Mean}     & \textbf{95\% CI}    \\ \midrule
    \multirow{6}{*}{Isolation Forest}                & 1                                                                               & 3.08\%            & [2.84\%, 3.32\%]    & \textbf{49.99\%}  & [49.99\%, 49.99\%] & 4.44\%             & [3.31\%, 5.56\%]     \\
                                                     & 2                                                                               & 3.08\%            & [2.84\%, 3.32\%]    & 49.98\%           & [49.97\%, 49.98\%] & 11.15\%            & [9.29\%, 13.01\%]    \\
                                                     & 5                                                                               & 3.08\%            & [2.28\%, 3.88\%]    & 49.92\%           & [49.87\%, 49.97\%] & 23.51\%            & [15.72\%, 31.3\%]    \\
                                                     & 10                                                                              & 3.13\%            & [2.34\%, 3.92\%]    & 49.83\%           & [49.72\%, 49.95\%] & 23.79\%            & [16.43\%, 31.15\%]   \\
                                                     & 30                                                                              & 3.93\%            & [3.68\%, 4.19\%]    & 48.96\%           & [48.59\%, 49.34\%] & 37.51\%            & [35.49\%, 39.52\%]   \\
                                                     & 60                                                                              & \textbf{7.80\%}   & [6.8\%, 8.8\%]      & 48.62\%           & [48.02\%, 49.22\%] & \textbf{44.10\%}   & [42.3\%, 45.9\%]     \\ \midrule
    \multirow{6}{*}{\begin{tabular}[c]{@{}c@{}}One-Class SVM \\ (Kernel = RBF)\end{tabular}}    & 1                                                                               & 44.97\%           & [43.62\%, 46.31\%]  & 55.09\%           & [54.03\%, 56.15\%] & 52.70\%            & [52.25\%, 53.15\%]   \\
                                                     & 2                                                                               & 45.36\%           & [40.15\%, 50.56\%]  & 53.69\%           & [50.31\%, 57.07\%] & 53.82\%            & [50.89\%, 56.76\%]   \\
                                                     & 5                                                                               & 46.20\%           & [45.04\%, 47.36\%]  & 56.88\%           & [55.81\%, 57.95\%] & 53.44\%            & [52.77\%, 54.1\%]    \\
                                                     & 10                                                                              & 51.64\%           & [47.56\%, 55.72\%]  & 58.07\%           & [55.18\%, 60.96\%] & 57.37\%            & [52.68\%, 62.06\%]   \\
                                                     & 30                                                                              & 60.20\%           & [56.14\%, 64.25\%]  & \textbf{60.37\%}  & [57.32\%, 63.42\%] & 74.30\%            & [66.69\%, 81.92\%]   \\
                                                     & 60                                                                              & \textbf{62.17\%}  & [58.23\%, 66.12\%]  & 59.64\%           & [56.7\%, 62.59\%]  & \textbf{84.10\%}   & [76.64\%, 91.56\%]   \\ \midrule
    \multirow{6}{*}{\begin{tabular}[c]{@{}c@{}}One-Class SVM \\ (Kernel = Linear)\end{tabular}} & 1                                                                               & 60.48\%           & [59.49\%, 61.48\%]  & 64.11\%           & [63.26\%, 64.97\%] & 62.38\%            & [61.33\%, 63.42\%]   \\
                                                     & 2                                                                               & 60.92\%           & [59.79\%, 62.04\%]  & 63.55\%           & [62.63\%, 64.48\%] & 65.07\%            & [63.7\%, 66.44\%]    \\
                                                     & 5                                                                               & 63.74\%           & [62.44\%, 65.05\%]  & 63.70\%           & [62.69\%, 64.71\%] & 71.84\%            & [70.32\%, 73.36\%]   \\
                                                     & 10                                                                              & 67.48\%           & [63.86\%, 71.1\%]   & \textbf{64.88\%}  & [62.05\%, 67.71\%] & 80.29\%            & [75.04\%, 85.54\%]   \\
                                                     & 30                                                                              & \textbf{68.00\%}  & [64.25\%, 71.75\%]  & 64.64\%           & [61.62\%, 67.66\%] & \textbf{83.31\%}   & [77.41\%, 89.2\%]    \\
                                                     & 60                                                                              & 65.95\%           & [62.1\%, 69.8\%]    & 63.73\%           & [60.63\%, 66.83\%] & 82.09\%            & [75.32\%, 88.87\%]   \\ \bottomrule
    \end{tabular}
\end{table*}

\begin{table*}[]
    \centering
    \caption{\changed{Online Binary Classification Results.}}
    \label{tab:results_online}
    \begin{tabular}{@{}cccccccc@{}}
    \toprule
    \multirow{2}{*}{\textbf{Classifier}}                         & \multirow{2}{*}{\textbf{\begin{tabular}[c]{@{}c@{}}Window\\ Size\end{tabular}}} & \multicolumn{2}{c}{\textbf{F-Score}} & \multicolumn{2}{c}{\textbf{Recall}} & \multicolumn{2}{c}{\textbf{Precision}} \\
                                                                 &                                                                                 & \textbf{Mean}    & \textbf{95\% CI}   & \textbf{Mean}    & \textbf{95\% CI}  & \textbf{Mean}     & \textbf{95\% CI}    \\ \midrule
    \multirow{6}{*}{\begin{tabular}[c]{@{}c@{}}Adaptive\\ Random Forest \\ (No Drift Detection)\end{tabular}} & 1                                                                               & 73.39\%           & [66.46\%, 80.33\%]  & 63.23\%           & [54.63\%, 71.83\%] & 94.56\%            & [93.08\%, 96.05\%]   \\
                                                                 & 2                                                                               & 81.66\%           & [75.24\%, 88.07\%]  & 73.11\%           & [64.97\%, 81.26\%] & 97.96\%            & [97.26\%, 98.65\%]   \\
                                                                 & 5                                                                               & 93.16\%           & [90.11\%, 96.2\%]   & 88.36\%           & [83.71\%, 93.01\%] & 99.60\%            & [99.26\%, 99.95\%]   \\
                                                                 & 10                                                                              & 97.07\%           & [95.4\%, 98.73\%]   & 94.70\%           & [91.82\%, 97.58\%] & 99.92\%            & [99.88\%, 99.96\%]   \\
                                                                 & 30                                                                              & 99.60\%           & [99.37\%, 99.84\%]  & 99.25\%           & [98.8\%, 99.7\%]   & 99.97\%            & [99.95\%, 99.99\%]   \\
                                                                 & 60                                                                              & \textbf{99.90\%}  & [99.81\%, 99.98\%]  & \textbf{99.81\%}  & [99.65\%, 99.97\%] & \textbf{99.99\%}   & [99.97\%, 100\%]     \\ \midrule
    \multirow{6}{*}{SGD}                                         & 1                                                                               & 85.90\%           & [84.26\%, 87.54\%]  & 80.76\%           & [78.86\%, 82.65\%] & 93.64\%            & [92.92\%, 94.35\%]   \\
                                                                 & 2                                                                               & 93.98\%           & [92.9\%, 95.07\%]   & 91.45\%           & [90.11\%, 92.79\%] & 97.29\%            & [96.77\%, 97.82\%]   \\
                                                                 & 5                                                                               & 98.45\%           & [98.07\%, 98.83\%]  & 97.80\%           & [97.36\%, 98.25\%] & 99.13\%            & [98.81\%, 99.44\%]   \\
                                                                 & 10                                                                              & 99.28\%           & [99.15\%, 99.42\%]  & \textbf{98.95\%}  & [98.79\%, 99.11\%] & 99.63\%            & [99.51\%, 99.74\%]   \\
                                                                 & 30                                                                              & 99.22\%           & [99.02\%, 99.43\%]  & 98.69\%           & [98.35\%, 99.04\%] & 99.81\%            & [99.77\%, 99.84\%]   \\
                                                                 & 60                                                                              & \textbf{99.29\%}  & [98.99\%, 99.59\%]  & 98.82\%           & [98.33\%, 99.32\%] & \textbf{99.87\%}   & [99.84\%, 99.9\%]    \\ \midrule
    \multirow{6}{*}{Perceptron}                                  & 1                                                                               & 87.83\%           & [86.92\%, 88.74\%]  & 87.43\%           & [86.5\%, 88.36\%]  & 88.23\%            & [87.34\%, 89.13\%]   \\
                                                                 & 2                                                                               & 94.50\%           & [93.92\%, 95.08\%]  & 94.53\%           & [93.95\%, 95.11\%] & 94.48\%            & [93.9\%, 95.06\%]    \\
                                                                 & 5                                                                               & 98.12\%           & [97.84\%, 98.4\%]   & 98.16\%           & [97.88\%, 98.44\%] & 98.08\%            & [97.8\%, 98.37\%]    \\
                                                                 & 10                                                                              & 99.17\%           & [99.07\%, 99.27\%]  & 99.22\%           & [99.12\%, 99.31\%] & 99.12\%            & [99.01\%, 99.23\%]   \\
                                                                 & 30                                                                              & 99.69\%           & [99.65\%, 99.74\%]  & 99.72\%           & [99.68\%, 99.77\%] & 99.66\%            & [99.62\%, 99.7\%]    \\
                                                                 & 60                                                                              & \textbf{99.80\%}  & [99.77\%, 99.84\%]  & \textbf{99.85\%}  & [99.81\%, 99.88\%] & \textbf{99.76\%}   & [99.73\%, 99.8\%]    \\ \bottomrule
    \end{tabular}
\end{table*}




\end{document}